 \documentclass[final,3p,times,twocolumn]{elsarticle}
 
 \makeatletter
\def\ps@pprintTitle{%
    \let\@oddhead\@empty
    \let\@evenhead\@empty
    \def\@oddfoot{\footnotesize\itshape
         {~} \hfill\today}%
    \let\@evenfoot\@oddfoot
    }
\makeatother

\usepackage[figuresleft]{rotating}
\usepackage{adjustbox}
\usepackage{setspace}
\usepackage{tocloft}
\usepackage{mathtools}
\usepackage{eurosym}
\usepackage{fancyhdr}
\usepackage{algpseudocode} 
\usepackage{textcomp}
\usepackage{tikz}
\usepackage{subcaption}
\usepackage{booktabs}
\usepackage{placeins}
\usepackage{graphicx}
\usepackage{color}
\usepackage{array}
\usepackage{multirow}
\usepackage{url} 
\usepackage{needspace}
\usepackage[printonlyused]{acronym}
\usepackage{float}
\usepackage{rotating}
\usepackage[titletoc,toc,page]{appendix}
\usepackage{lipsum}

\newcommand\MyBox[2]{
  \fbox{\lower0.75cm
    \vbox to 1cm{\vfil
      \hbox to 1cm{\hfil\parbox{1.4cm}{#1\\#2}\hfil}
      \vfil}%
  }%
}
\usepackage{amssymb}

\let\today\relax

\begin{document}

\begin{frontmatter}





\title{Indoor Environment Data Time-Series Reconstruction Using Autoencoder Neural Networks}


 \author[label1]{Antonio Liguori*$^{,}$} \author[label2]{Romana Markovic} 
 \author[label1]{Thi Thu Ha Dam}
 \author[label1]{J\'{e}r\^{o}me Frisch} \author[label1]{Christoph van Treeck} \author[label3]{Francesco Causone}   


\address[label1]{E3D - Institute of Energy Efficiency and Sustainable Building, RWTH Aachen University, Mathieustr. 30, 52074 Aachen, Germany}

\address[label2]{Building Science Group, Karlsruhe Institute of Technology, Englerstr. 7, 76131 Karlsruhe, Germany}

\address[label3]{Department of Energy, Politecnico di Milano, Via Lambruschini 4, 20156 Milano, Italy}

\cortext[cor1]{Corresponding author. Tel.: +49 241 80 25541; \\E-Mail: liguori@e3d.rwth-aachen.de }

\begin{abstract}

As the number of installed meters in buildings increases, there is a growing number of data time-series that could be used to develop data-driven models to support and optimize building operation. However, building data sets are often characterized by errors and missing values, which are considered, by the recent research, among the main limiting factors on the performance of the proposed models. Motivated by the need to address the problem of missing data in building operation, this work presents a data-driven approach to fill these gaps. In this study, three different autoencoder neural networks are trained to reconstruct missing short-term indoor environment data time-series in a data set collected in an office building in Aachen, Germany. This consisted of a four year-long monitoring campaign in and between the years 2014 and 2017, of 84 different rooms. The models are applicable for different time-series obtained from room automation, such as indoor air temperature, relative humidity and $CO_{2}$ data streams. The results prove that the proposed methods outperform classic numerical approaches and they result in reconstructing the corresponding variables with average RMSEs of 0.42 \textdegree C, 1.30~\% and 78.41 ppm, respectively.
\end{abstract}

\begin{keyword}
indoor environment data time-series  \sep machine learning \sep data analytics \sep autoencoder \sep neural networks. 
\end{keyword}
\end{frontmatter}
\onecolumn

\FloatBarrier

\section{Introduction}






In the European Union buildings account for more than 40~\% of the total final energy consumption and approximately 36~\% of $CO_{2}$ emissions \cite{EUDirective}. As a consequence, reliable estimation of building consumption data could foster energy efficiency strategies, such as the analyses of retrofit options \cite{fumo} or the development of fault detection and diagnosis (FDD) schemes \cite{khan}. In the related research, two approaches are generally followed to achieve this goal \cite{Fundamentals}: forward modeling and data-driven modeling. While the former is based on solid engineering principles, the latter relies on data collected during normal or predetermined system operation and it can usually capture more accurate as-built system's performance with a limited number of known parameters \cite{Fundamentals}. Additionally, data-driven approaches can be successfully applied to represent energy-related human actions in buildings (e.g. window openings), being the result of a number of stochastic driving forces \cite{RoM,causone}. 
\par
By definition, data-driven modeling explicitly requires the availability of useful data \cite{markovic2020}. Therefore, missing values present the major limitation on this approach \cite{chong}. As stated by multiple studies \cite{markovic2020, chong}, data gaps are a common problem in building automation systems (BAS) and they may be caused by a number of reasons such as power outages, sensors defects, communication problems or network issues. As a result, the presence of these  anomalies could significantly reduce the size of the available data set and hinder further energy analysis \cite{star,zapata}. So far, existing studies have often handled missing data either using simplified methods \cite{chong} or excluding them from further analytics due to the lack of ground truth values \cite{markovic2020}. In summary, both latter approaches have usually led to limited inserting accuracy and lower resulting model performance \cite{chong,markovic2020}.
\par
Even though the related research has already identified some promising techniques to address missing values and anomalies in monitoring building data sets \cite{luis,rahman,benitez,wang,peppanen,ma,zhang}, some significant research questions remain still unaddressed. In particular, the existing studies have often either focused on reconstructing a single type of signal or made use of a number of specific information from additional sensors. Furthermore, the advance of deep learning in the building related research \cite{marjan,zuo,luo,somu,ruan,iruela,tran,bui,tian,qian,ji,meng,mo,ke,hengda,cao,han,das,RoM1,RoM3} opens to more interesting possibilities that have not yet been fully explored, due to the small amount of available data and computational power experienced in most of the existing studies.
\par
Based on the previous considerations, the aim of this paper is to propose a deep learning-based method for reconstructing missing sequences of various indoor environment streams obtained from room control sensors. Due to the frequency of short-term gaps (i.e. $<$ 24 h) in building data sets \cite{clar,balt}, the focus is on the inserting of small intervals of values. For that purpose, based on data availability to the authors, a data set collected in an office building located in Aachen, Germany, was analyzed and preprocessed. This consisted of a four year-long monitoring campaign of 84 different rooms and it could, therefore, provide significant generalization capability to the implemented approach. Models for handling missing data points were implemented, trained and evaluated on indoor air temperature ($T$), relative humidity ($RH$) and $CO_2$ concentration data. In particular, due to the increasing use of autoencoder neural networks in the FDD-related research for buildings' control \cite{jiayuan,Yu,araya,legrand,benitez1}, three promising autoencoder architectures were investigated: feed-forward denoising autoencoder, convolutional denoising autoencoder and long short-term memory (LSTM) denoising autoencoder. The optimal problem hypothesis was identified based on the results obtained from 7,000 core hours of GPU and CPU computations. Eventually, the performance of the proposed methods were compared to analytical methods based on polynomial interpolations. 

\par 
The scientific contribution of the presented work consists of the following:
\begin{itemize}
    \item To explore the full potential of autoencoder neural networks for reconstructing indoor environment data time-series.
    \item To analyze the performance variability of deep learning models when applied to different kinds of monitoring building data.
    \item To present a generalized gap-filling method to address the problem of missing values in building data sets.
    \item To propose a solution to address the issue of artificial neural networks' (ANNs) saturation for energy systems applications.
\end{itemize}

The rest of this paper is organized as follows: Section \ref{sec:background} presents the motivation that led to the development of a missing data inserting model based on autoencoder neural networks. Section \ref{sec:methodology} provides the reader with further information on the used data set and on the models' theory and implementation. Section \ref{sec:results} presents results on developing a suitable  tool  for  indoor  environment  data time-series reconstruction. Finally, the results and novel findings are discussed and summarized in Sections \ref{sec:discussion} and \ref{sec:conclusions}.

\begin{table}[H]

\centering

   \caption{List of abbreviations. }
     \begin{tabular}{|ll|}
     			
\hline
ANNs	&	artificial neural networks \\
BAS	&	building automation systems	\\
BIT & 	bi-directional imputation and transfer learning \\
BN & bayesian network \\
CR	&	corruption rate	\\
DBN &   deep belief network \\
ELM &   extreme learning machine \\
EM & expectation–maximization \\
FDD	&	fault detection and diagnosis	\\
FFT	&	fast fourier transform	\\
GANs &   generative adversarial networks \\
IAQ	&	indoor air quality	\\
IQR	&	interquartile range	\\
LSTM	&	long short-term memory	\\
MAE &   mean absolute error \\
MELs	&	miscellaneous electric loads	\\
MSE	&	mean squared error	\\
NRMSE	& 	normalized root mean squared error \\
OB	&	occupant behavior	\\
OWA &  optimally weighted average \\
RBMs &   restricted boltzmann machines \\
RF	&	random forest \\
RH	&	indoor relative humidity	\\
RMSE	&	root mean squared error	\\
RNN	&	recurrent neural network	\\
SAT	&	saturation performance metric	\\
SGD	&	stochastic gradient descendent	\\
T	&	indoor air temperature \\
\hline				
\end{tabular}

\label{tab:abbr}
\end{table}

\section{Background}
\label{sec:background}

\subsection{Missing data in buildings' control}
\label{sec:missing1}
The importance of sufficient large data sets on time-series modeling was empirically explored in the scope of the Texas LoanSTAR program \cite{star}, whose objective was to measure savings from energy conservation retrofits. By increasing the length of building data sets from one to five months, the average cooling prediction error decreased from 7.3~\% to 3.0~\% and the annual heating prediction error decreased from 27.5~\% to 12.9~\%. Zapata et al. \cite{zapata} discovered also that a fast fourier transform (FFT), as applied to assess the frequency content of time-series wind speed data, gave unacceptable results when information loss was at least 2.5~\% of the data set. Therefore, the presence of missing values in building data sets may affect the performance of the adopted method and hinder further energy analysis. In particular, the analysis of an energy use data set, collected in over 200 buildings in Texas \cite{star}, revealed that about 60~\% of the observed missing data were in the range between 1 and 6 hours \cite{clar,balt}.
\par 
As pointed out by Chong et al. \cite{chong}, in 2016 there was still little relevant research about handling missing values in building data sets. As a consequence, existing studies often relied on simplified methods such as complete case analysis, mean inserting and zeros inserting \cite{chong}. However, these methods often resulted in poor reconstruction of missing data and they often led to limited performance of later applied data-driven models. In recent years, the topic of missing data inserting for building monitoring data has gained increasing importance and different studies have been carried out. Candanedo et al. \cite{luis} reconstructed the average indoor air temperatures of a passive house, achieving accurate results with a random forest (RF) model. Rahman et al. \cite{rahman} presented a recurrent neural network (RNN) for the prediction of electricity consumption in commercial and residential buildings. The same model was applied for the missing values inserting in the event of large data gaps (i.e. bigger than 5 hours). However, the models proposed in the previous studies used as input features multiple time-series (e.g. external weather data, total electrical energy use, indoor variables) that could not be always known for a data reconstruction problem. In this regard, a further improvement was noted in Benitez et al. \cite{benitez} and  Wang et al. \cite{wang}. Benitez et al. \cite{benitez} coupled multivariate variational autoencoders with convolutional layers to estimate missing indoor air quality (IAQ) subway data. The model's inputs consisted of 9 different IAQ variables and the missing values were reconstructed based on the captured correlations between these variables. A similar approach was followed in Wang et al. \cite{wang}. Here, a FDD method based on expectation–maximization (EM) algorithm and Bayesian network (BN) was developed to insert missing values from a chiller's data set. In this case, the model's inputs consisted of 9 different variables that indicated the healthy status of the system. Hence, in the last two approaches, weather information were not exploited but measurements from multiple sensors were required. This may represent a particular complexity for the models, since the required meters' data might be not available for diverse reasons. Differently from the previous studies, the models proposed in  Peppanen et al. \cite{peppanen} and Ma et al. \cite{ma} had the advantage to need only the historical measurements of the same type of data (i.e. power), without additional specific information. Peppanen et al. \cite{peppanen} implemented an optimally weighted average (OWA) load power imputation method for smart meter measurements. Ma et al. \cite{ma} applied an univariate LSTM with bi-directional imputation and transfer learning (BIT) for electric consumption data inserting of a campus lab building, achieving a reconstruction error approximately 30~\% less than linear interpolation models. However, since these studies focused only on the reconstruction of a single signal, it is not clear how the proposed models would behave when applied to other types of building monitoring data. \par
In summary, there is a gap in the related research that does not address the missing data reconstruction performance variability of a single model when applied to different single streams of building monitoring data. Few works are found to analyze this case and they either focus on simplified approaches (i.e. forward propagation) \cite{zhang} or on advanced methods without further optimization \cite{zhang}. This paper tries to fill this gap, by exploiting the full potential of deep learning on diverse indoor environment streams.

\subsection{Deep learning methods for buildings' control}
Deep learning models are neural networks with learned feature representation over multiple hidden layers \cite{Goodfellow}. In the related building research, these methods have been extensively used for energy consumption and occupant behavior (OB) modeling applications \cite{marjan,zuo,luo,somu,ruan,iruela,tran,bui,tian,qian,ji,meng,mo,ke,hengda,cao,han,das,RoM1,RoM3}. Qian et al. \cite{qian} explored the potential of ANNs for HVAC load forecasting, when applied to small amount of data. The results proved that the fitting degree of the proposed models was over 85~\%. They pointed out the importance of the sufficiently large training data set. In particular, the use of a smaller training set that consisted of one month and one week of monitoring data led to accuracy decrease for 6~\% and 20~\%, respectively. Zhang et al. \cite{zuo} trained a deep belief network (DBN) and extreme learning machine (ELM) based framework to predict half-hourly building energy consumption data. Here, DBNs consisted of a stack of restricted boltzmann machines (RBMs), where RBMs had fully connected visible and hidden layers \cite{zuo}. The correspondent mean absolute error (MAE) was 10~\% higher than the results obtained by support vector regression. Yan et al. \cite{mo} addressed the issue of the imbalanced properties of training data sets for automatic FDD of chillers. They used generative adversarial networks (GANs) to generate faulty training samples. The results proved that without the implemented model, classification accuracy could hardly reach 90~\%. Conclusively, Markovic et al. \cite{RoM1} developed a LSTM neural network for day-ahead prediction of miscellaneous electric loads (MELs). The proposed implementation outperformed benchmark approaches based on Weibull distribution and Gaussian mixture methods when MELs and occupancy information data were used as input parameters to the model.

\subsection{Autoencoder neural networks for buildings' control}
Autoencoders are ANNs which learn to reconstruct original inputs from a noisy version, making missing data reconstruction one of their reasonable applications \cite{moore}. In the field of building energy systems, these models have been increasingly applied for FDD purposes \cite{jiayuan,Yu,araya,legrand,benitez1}. Fan et al. \cite{jiayuan} explored the potential of different types of autoencoder neural networks in the anomaly detection of building operational data. The results showed that a 1D convolutional architecture could effectively capture the intrinsic characteristics in building energy data, while preserving the temporal data information. Liu et al. \cite{Yu} applied different machine learning-based anomaly detection methods to vertical plant wall systems. The results confirmed that the autoencoder configuration outperformed other models for both contextual and point anomaly detection of temperature and $CO_2$ data. Araya et al. \cite{araya} used autoencoders to capture HVAC consumption patterns and they used the gained knowledge to identify abnormal consumption behaviours in the same system. Legrand et al. \cite{legrand} applied different autoencoder architectures for the anomaly detection of 19 sensors in multiple houses. The results showed that RNN autoencoders had better generalization capabilities than the other used configurations. Benitez et al. \cite{benitez1} developed an autoencoder-based sensor validation method for the FDD of a subway system. This approach outperformed conventional methods for detecting and reconstructing faulty measurements.

\section{Methodology}
\label{sec:methodology}
\begin{figure}[H]
    \centering
    \includegraphics[height=0.50\textheight]{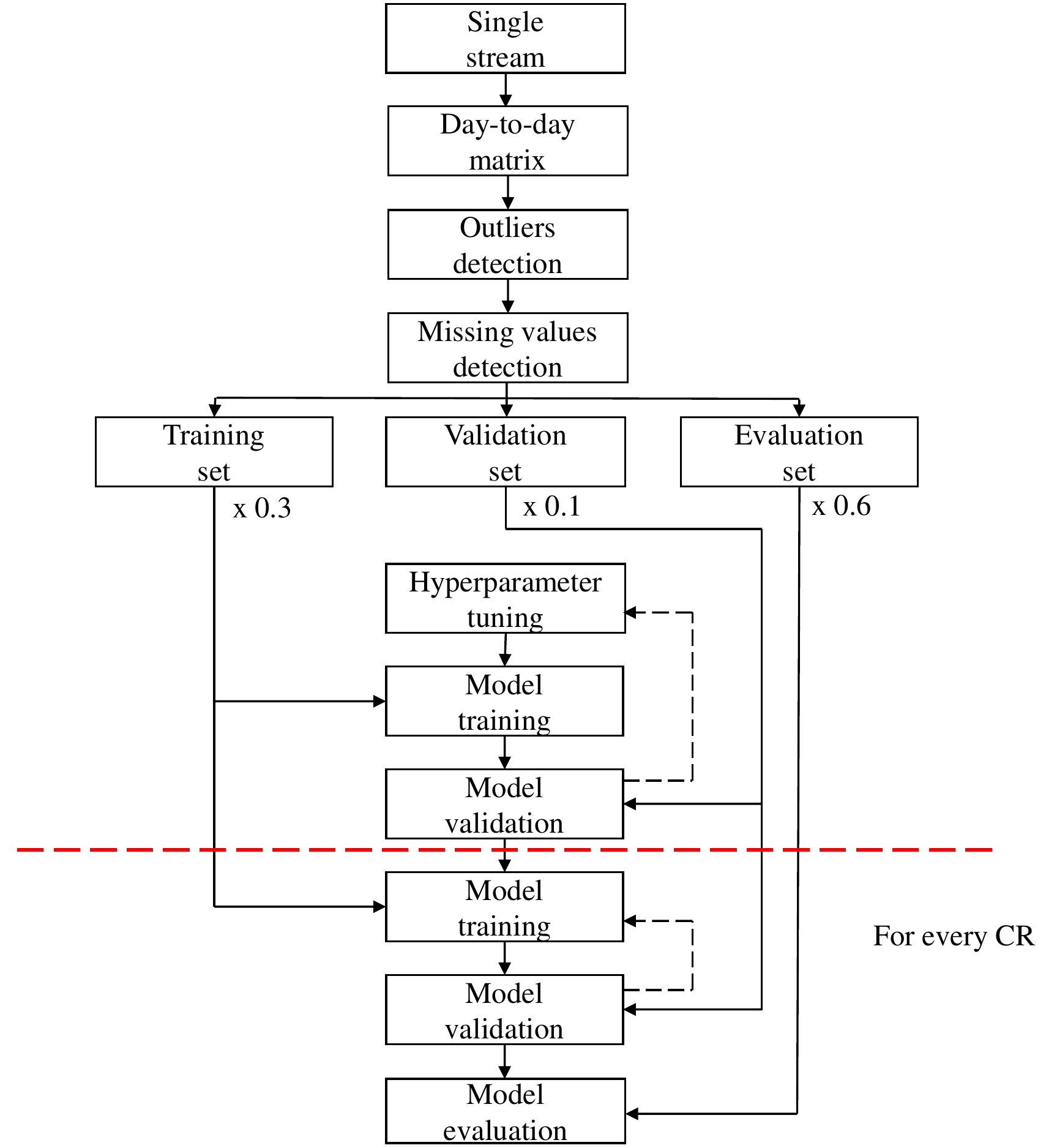}
    \caption{Modeling flowchart. CR is the applied masking noise.}
    \label{fig:workflow}
\end{figure}

The aim of this paper is to develop an approach for filling short-term indoor environment data gaps. For that purpose, three different autoencoder neural network architectures were implemented in order to identify the optimal model hypothesis. As described in Fan et al. \cite{jiayuan}, before further model development, it is common to rearrange the time-series into a matrix of sub-sequences. The length of each sub-sequence is selected based on the most dominant period identified in the data. In this work, indoor air temperature, relative humidity and $CO_2$ concentration data were found to have clear daily seasonality. Therefore, the indoor environment streams were rearranged into day-to-day matrices. Missing data in each stream were then simulated by setting to zero sub-daily sequences of random length. In particular, the reconstructed gaps ranged between few hours (10~\% of the daily values) and around 22 hours (90~\% of the daily values). A summary of the adopted modeling approach for each indoor environment stream is proposed in Figure \ref{fig:workflow}. 

\subsection{Denoising autoencoder neural networks}
\label{sec:autoencoder}

\begin{figure}[H]
\centering
\includegraphics[height=0.16\textheight]{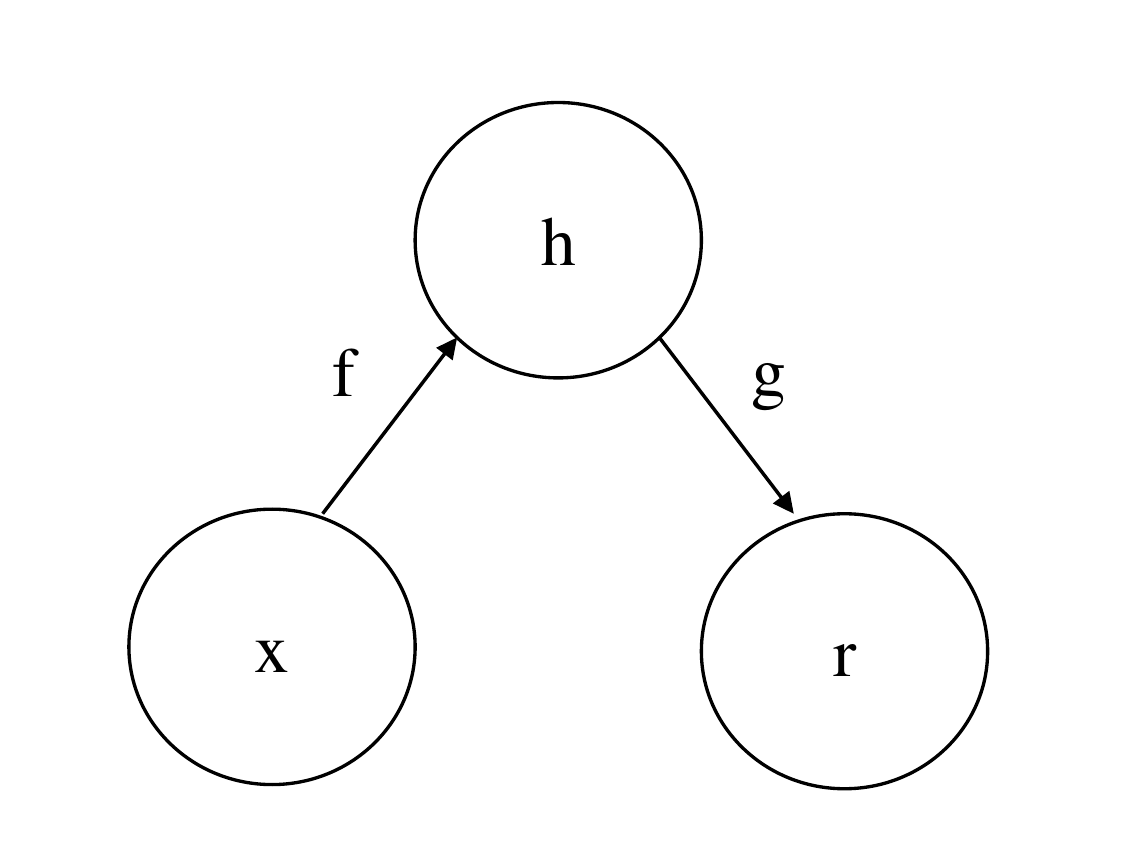}
\caption{Working principle of a general autoencoder. Figure reproduced based on Goodfellow \cite{Goodfellow}.}
\label{fig:autoencoder}
\end{figure}

The general structure of an autoencoder is presented in Figure \ref{fig:autoencoder}.
An autoencoder neural network is a representation learning approach which turns incoming data into different representations, through an encoder function, and reconstructs the original input through a decoder function \cite{Goodfellow}.

Input $x$ is mapped to an output reconstruction $r$ through an internal representation $h$, namely code. The autoencoder has two components: the encoder $f$ (mapping $x$ to $h$) and the decoder $g$ (mapping $h$ to $r$) \cite{Goodfellow}. \par
The encoder is defined as \cite{Goodfellow}:

\begin{equation}
 h = c(W x + b)~,
\label{eq:encoder}
\end{equation}

while the decoder is formulated as:

\begin{equation}
r = c(W' h + b')~,
\label{eq:decoder}
\end{equation}

where $c$ is a non-linear function, called activation function, $W$ and $W'$ are called weights, $b$ and $b'$ are called biases.
By implementing activation several times (training), the model could acquire useful knowledge about the systems' properties \cite{Goodfellow}. \par
\begin{figure}[H]
\centering
\includegraphics[height=0.16\textheight]{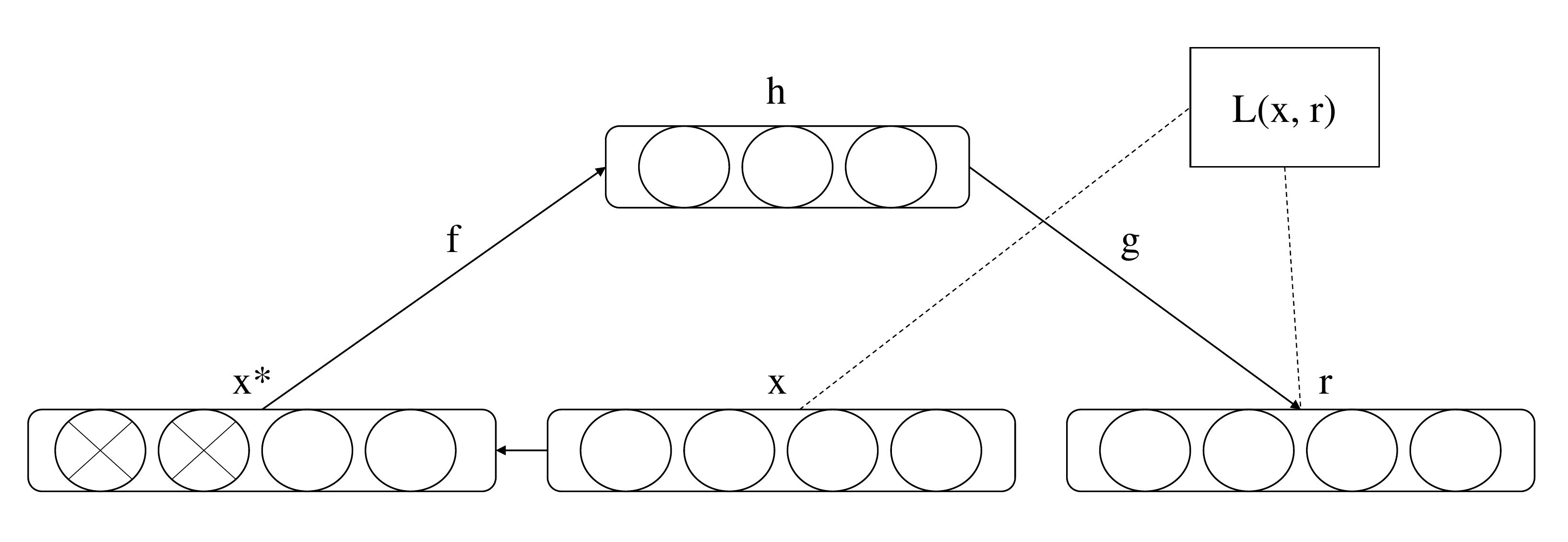}
\caption{Working principle of a denoising autoencoder. Figure reproduced based on Vincent et al. \cite{Vincent}.}
\label{fig:autocorr}
\end{figure}
\begin{figure}[H]
\centering
\includegraphics[height=0.23\textheight]{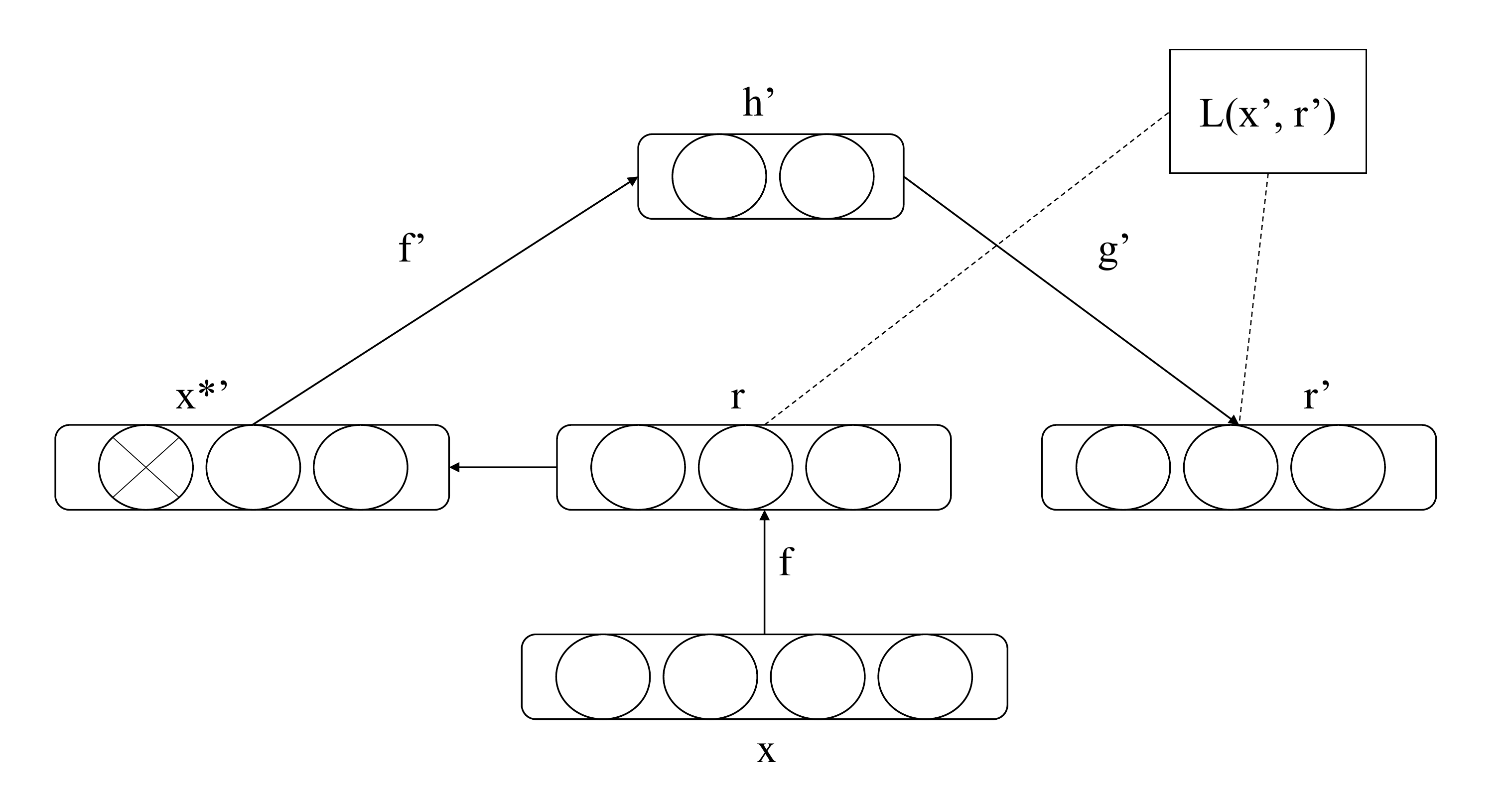}
\caption{Working principle of a stacked denoising autoencoder. Figure reproduced based on Vincent et al. \cite{Vincent}.}
\label{fig:autocorr1}
\end{figure}

The structure of a denoising autoencoder is presented in Figure \ref{fig:autocorr}, while its extension to the stacked denoising autoencoder is presented in Figure \ref{fig:autocorr1}. In contrast to a general autoencoder, a denoising autoencoder receives a corrupted input $x^{*}$ and is trained to reconstruct the original uncorrupted data $x $ \cite{Goodfellow}.

The training process consists of minimizing a loss function $L(x, r)$ which quantifies the difference between the original input and output at each step.
In this study, corruption of input data is performed by setting to zero an interval of sequential values of random length. This approach is used to simulate how missing data are distributed. 

After the first level of denoising autoencoder has been trained (Figure \ref{fig:autocorr}), a second level of denoising autoencoder is trained  using the previously optimized encoding function, $f$. Corruption takes place on the output of the previous optimized layer $r$ \cite{Vincent}. \par

\subsection{Data set}
\label{sec:database}
The used data were collected in the E.ON ERC main building, located in Aachen, Germany. The building under investigation is a mechanically ventilated building with a usable area of 7,500 $m^2$ over four storeys. The same building has U-values of 1.22 $\frac{W}{m^2K}$ for windows and 0.19 $\frac{W}{m^2K}$ for the outer walls, a yearly primary energy demand of 105.1 $\frac{kWh}{m^2a}$ and a net energy demand of 68 $\frac{kWh}{m^2a}$ \cite{eon, eon1, eon2, eon3}. \par
Temperature, relative humidity and $CO_2$ concentration data were recorded in every room of the building, namely offices, seminar and conference rooms, staff facilities, laboratories and common area. Here, each sensor was integrated within a BACnet network. Room temperature was measured by means of thermocouples, which were calibrated by measuring and comparing the temperature offsets with the actual sensors' accuracy. Due to data sharing agreement regarding the hardware installation, no additional information about the $RH$ and $CO_2$ sensors' architecture and calibration was provided to the authors in the energy report of the building. \par
Based on the logging frequency and monitoring duration, it could be expected that around 181 million sets of observations were collected for each variable from 2014 to 2017. The monitoring data were grouped in two subsets, namely data set "A" (2014-2015) and "B" (2016-2017).
Data set "A" contained measurements for 73 rooms from 2014 and 2015, stored into "HDF5" data containers  \cite{hdf} on a monthly basis. 
Monitoring data for the years 2016 and 2017 were collected in 84 offices and they were stored in "pickle" files \cite{Python}. Here, each file contained data for a single office over the whole observed biannual period.

\subsection{Data preprocessing}
\label{sec:processing}

Before further analysis and modeling, data were cleaned and preprocessed. This step involved the detection of frequently encountered anomalies, such as missing values and outliers \cite{Sang}. Missing values reduce the size of the available data set, hence compromising the reliability of the model's outcome \cite{Sang}. Outliers are noisy data points with values significantly different from the majority of other data points \cite{Sang}. For this reason, they could lead to underestimated or overestimated results \cite{Sang}. For the detailed explanation of the adopted data cleaning procedure the reader is referred to the Appendix \ref{sec:cl}.\par
According to the existing literature \cite{Ratz,T,causone1}, additional preprocessing, such as resampling resolution and data normalization was performed in order to increase performance and computational efficiency of the proposed models. Here, resampling resolution is the process of changing the frequency of time-series data \cite{Ratz}, while data normalization is applied to prepare raw data for a better network use \cite{T}. In particular, data were downsampled to 30 minutes frequency and normalized using the Statistical or Z-Score Normalization function \cite{T}:

\begin{equation}
    z = (x-u)/s~,
    \label{eq:norm}
\end{equation}

where, $x$ are data to normalize, $u$ is the mean of the training samples and $s$ is the standard deviation of the training samples. Before normalizing, data were split into training (30~\%), validation (10~\%) and evaluation set (60~\%) for every variable. In order to favour models’ generalization, limits were defined based on training set and eventually adopted for each of the three parts identically. \par 

\subsection{Model development}
\label{sec:models}

\begin{figure}[H]
\centering
\includegraphics[height=0.35\textheight]{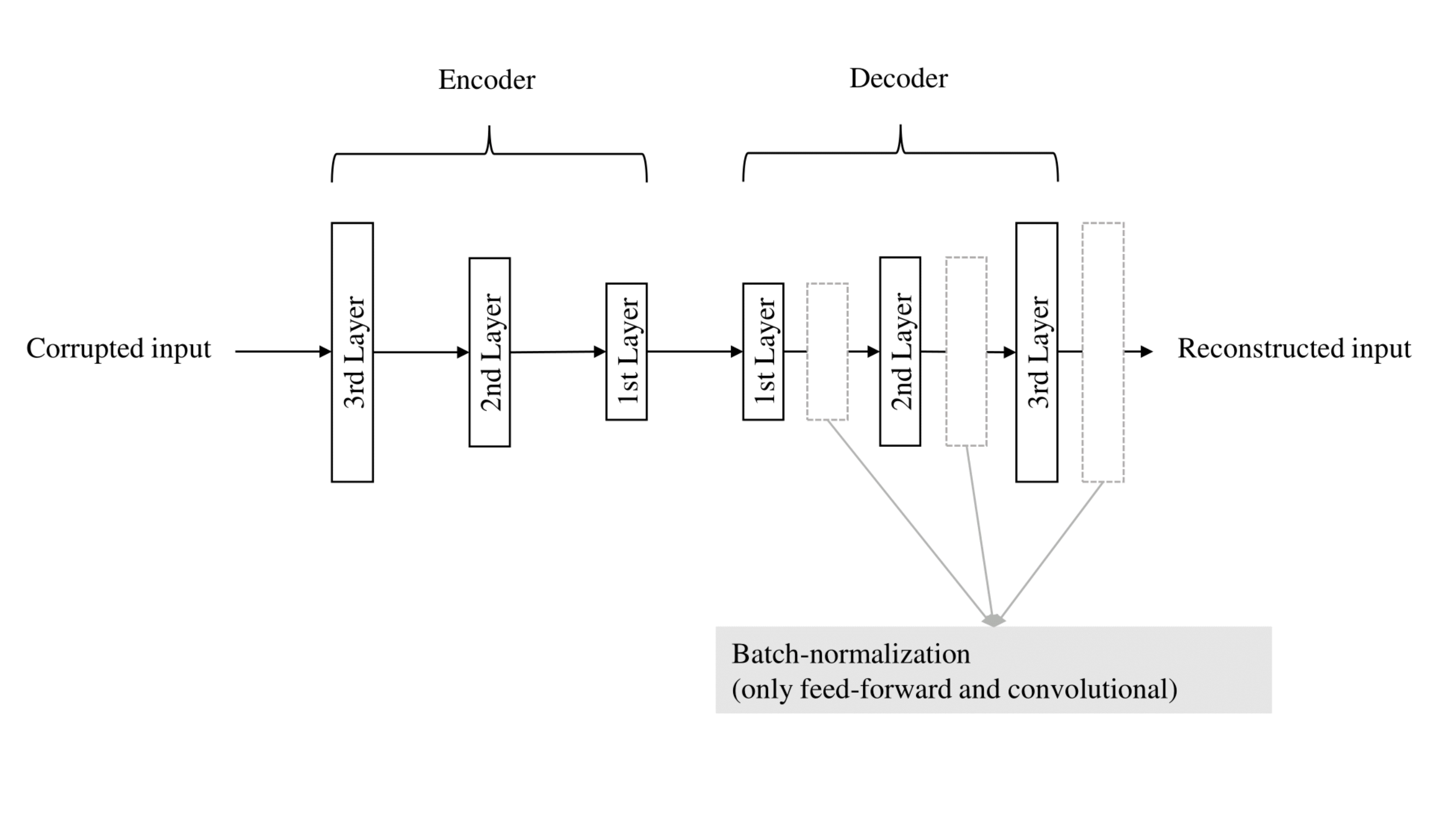}
\caption{General autoencoder architecture with 3 layers per side.}
\label{fig: general_autoencoder}
\end{figure}

The models were developed using the Python programming language and open source libraries Tensorflow \cite{tensorflow} and Keras \cite{keras}. Figure \ref{fig: general_autoencoder} shows the autoencoder structure as implemented in the scope of the performed experiments. The explored models included the architectures with one and up to three hidden layers per each encoder and decoder. Batch-normalization was included after every layer in order to avoid the network saturation for both feed-forward and convolutional autoencoders \cite{ioffe}. 

Feed-forward denoising autoencoders were fed with unrolled half-hourly daily observations, which resulted in 48 features. Unrolling the 1-D temporal sequences into a single input layer is, indeed, a commonly adopted approach to address temporal dependencies for a feed-forward neural network \cite{RoM,RoM3}. However, feed-forward neural networks with unrolled sequences have separate parameters for each feature, which means that weights cannot be shared along the input series \cite{Goodfellow}. While convolutional neural networks apply the same kernel to every time-step, in the LSTM configuration each output node is a function of the previous one and parameters can be shared along very long sequences \cite{Goodfellow}. 
\par 
As common for the supervised learning, the overfitting can occur when the number of training iterations is set too high \cite{prechelt1998early}. In that case, the training error keeps decreasing, while the validation error increases due to lower model's generalization capabilities \cite{Goodfellow}. The similar phenomena could be observed in case of this study and occurance of overfitting is examplarly presented in Appendix \ref{sec:ov}. Therefore, the early stopping criteria was introduced in order to avoid overfitting \cite{Goodfellow}. Furthermore, the mean squared error (MSE) loss function was applied to the reconstructed and original input over all the training samples, as follows \cite{Goodfellow}:

\begin{equation}
MSE = \frac{\sum_{i}^{m}(Y' - Y)^{2}_i}{m}~,
\label{mse}
\end{equation}

where $m$ is the batch size. In order to find the optimal parameters for the models, the MSE was minimized using either a stochastic gradient descendent (SGD) optimizer with momentum \cite{ilya} or Adam optimizer \cite{adam}. The optimizer choice was handled as an additional hyperparameter. In this study, the hyperparameter tuning was conducted separately for each target variable (temperature, relative humidity, $CO_2$ concentration). Additionally, the whole hyperparameter range was explored for all neural units, such as feed-forward, LSTM and convolutional. The range of values in which the optimal model's configuration was investigated is summarized in Table \ref{tab:hyp_basic}.

\begin{table}[H]
\centering
\caption{Overview of tuned hyperparameter and explored values.}
\label{tab:hyp_basic}
\begin{tabular}{ |*{2}{p{3cm}|}  }
 \toprule
 \multicolumn{1}{c}{Hyperparameter} & \multicolumn{1}{c}{Values}\\
 \toprule
 \multicolumn{1}{l}{Hidden layers per side} & \multicolumn{1}{c}{1~-~3}\\
 \multicolumn{1}{l}{Hidden layer units/ filters (convolutional)} & \multicolumn{1}{c}{8, 16, 32, 64, 128}\\
  \multicolumn{1}{l}{Kernel size (convolutional)} & \multicolumn{1}{c}{2, 3, 5, 7, 11}\\
 \multicolumn{1}{l}{Batch size}& \multicolumn{1}{c}{128}\\
 \multicolumn{1}{l}{Learning rate}& \multicolumn{1}{c}{0.001, 0.01, 0.1}\\
 \multicolumn{1}{l}{Optimizer}& \multicolumn{1}{c}{SGD with 0.9 momentum, Adam}\\
 \toprule
\end{tabular}
\end{table}

In the scope of experimental design setting, both grid search and random search for hyperparameter tuning were considered. Eventually, it was opted for grid search, due to the relative small hyperparameter space. Namely, the searched hyperparameter space consisted of  approximately 1,890 model hypotheses, and no significant computational efficiency would be achieved by opting for an alternative hyperparameter tuning strategy. However, in case of the larger hyperparameter space, alternative tuning strategies should be considered, such as random or greedy search \cite{bengioo}. 

\par
For every possible combination of hyperparameter, the score given to the model was obtained by averaging the MSEs, computed on the normalized validation set, with a corruption rate (CR) ranging between 0.2 and 0.8. Given the computational cost of this process, multiple independent jobs with different hyperparameters were run in parallel using the computational resources at the RWTH Aachen University Compute Cluster. In order to address the stochastic initialization of models' weights, the tuned configurations were run again 10 times and evaluated on the same validation data as before. Models with the lowest MSE were eventually exported for further evaluation on the evaluation set. \par

\section{Results}
\label{sec:results}

\subsection{Performance evaluation metrics}
\label{sec: metrics}

The performance of missing data insertion was assessed using the root mean squared error (RMSE) method, since it is the established evaluation method by the existing research \cite{ma,luis,benitez}. In order to obtain objective evaluation results, the RMSE was applied to all the corrupted evaluation data. Additionally, the ability of the proposed models for capturing indoor environmental data patterns was estimated by computing the RMSE on each sequence \cite{jiayuan}. The RMSE equation is given as follows \cite{ma,luis,jiayuan}:

\begin{equation}
RMSE = \sqrt{\frac{\sum_{i=1}^{n}(X^{obs}_i - X^{inserted}_i)^2}{n}}~,
\label{eq:rmse}
\end{equation}

where $X^{obs}_i$ are the $i-th$ real values, $X^{inserted}_i$ are the $i-th$ reconstructed values and $n$ are the total number of data points on which RMSE is computed.\par
Comparison between different variables and studies was made by means of the normalized root mean squared error (NRMSE), obtained by normalizing the RMSE over the interquartile range (IQR), due to the possible presence of noisy data points:

\begin{equation}
NRMSE = \frac{RMSE}{IQR}~.
\label{eq:nrmse}
\end{equation}

\subsection{General observations during model development}
\label{sec: saturation}

Network saturation was identified to be a major modeling complexity in case of both feed-forward and convolutional denoising autoencoders. Formally, neural network saturation could be described as an impediment to gradient propagation \cite{glorot,RoM1}. Its main effect was to produce always the same output, no matter how different the input sequence was. \par
As presented earlier, neural networks are trained by minimizing a loss function between targets and true values. This can be accomplished by applying an iterative algorithm called gradient descendent \cite{groose}. The working principle of a gradient descendent algorithm is simple: weights and biases are initialized to some values and then they are continuously updated in the direction that decreases the loss function, namely opposite to the gradient \cite{groose}. Each unit in a neural network receives signals and weights from previous units and computes a value called pre-activation \cite{groose}:

\begin{equation}
z = \sum_{j}w_jx_j+b~,
\label{eq:pre_activation}
\end{equation}

where $z$ is the pre-activation value, $j$ is the number of input signals and weights, $w$ are the weights, $b$ is the bias which determines units' activation in case no inputs are present. The activation value is the pre-activation passed through an activation function $\phi$ \cite{groose}:

\begin{equation}
a = \phi(z)~.
\label{eq:activation}
\end{equation}

If, during training, the activation of a neural network unit is always near the boundaries of its dynamic range (the possible outputs of an activation function), then the gradient of the pre-activation is very small and weights are not updated \cite{groose}. These neural network units are called saturated units \cite{groose}.
Hence, saturated units can be identified looking at the histogram of the average activations and checking that they are not concentrated at the endpoints \cite{groose}.
Different approaches are followed in the literature to avoid network saturation. Glorot and Bengio \cite{glorot} proposed a novel normalized initialization for neural network weights:

\begin{equation}
W = U[-\frac{\sqrt{6}}{\sqrt{n_j+n_{j+1}}}, \frac{\sqrt{6}}{\sqrt{n_j+n_{j+1}}}]~,
\label{eq:weight}
\end{equation}

where $W$ are the neural network weights, $U$ is a uniform distribution function, $n$ is the size of the $j-th$ layer.
Maas et al. \cite{maas} used ReLU activation functions as alternative to sigmoid and $tanh$, causing in this way network saturation at exactly 0. Ioffe and Szegedy \cite{ioffe} proposed a new mechanism called batch-normalization, which ensures that distribution of nonlinearity inputs (i.e. pre-activations) remains more stable as the network trains.\par
In this study, since no other formal metrics were available at the current state of the research, network saturation was evaluated using the saturation performance metric (SAT) proposed by Markovic et al. \cite{RoM1}, calculated over the normalized reconstructed values. \par
\begin{figure}[H]
\centering
\includegraphics[height=0.30\textheight]{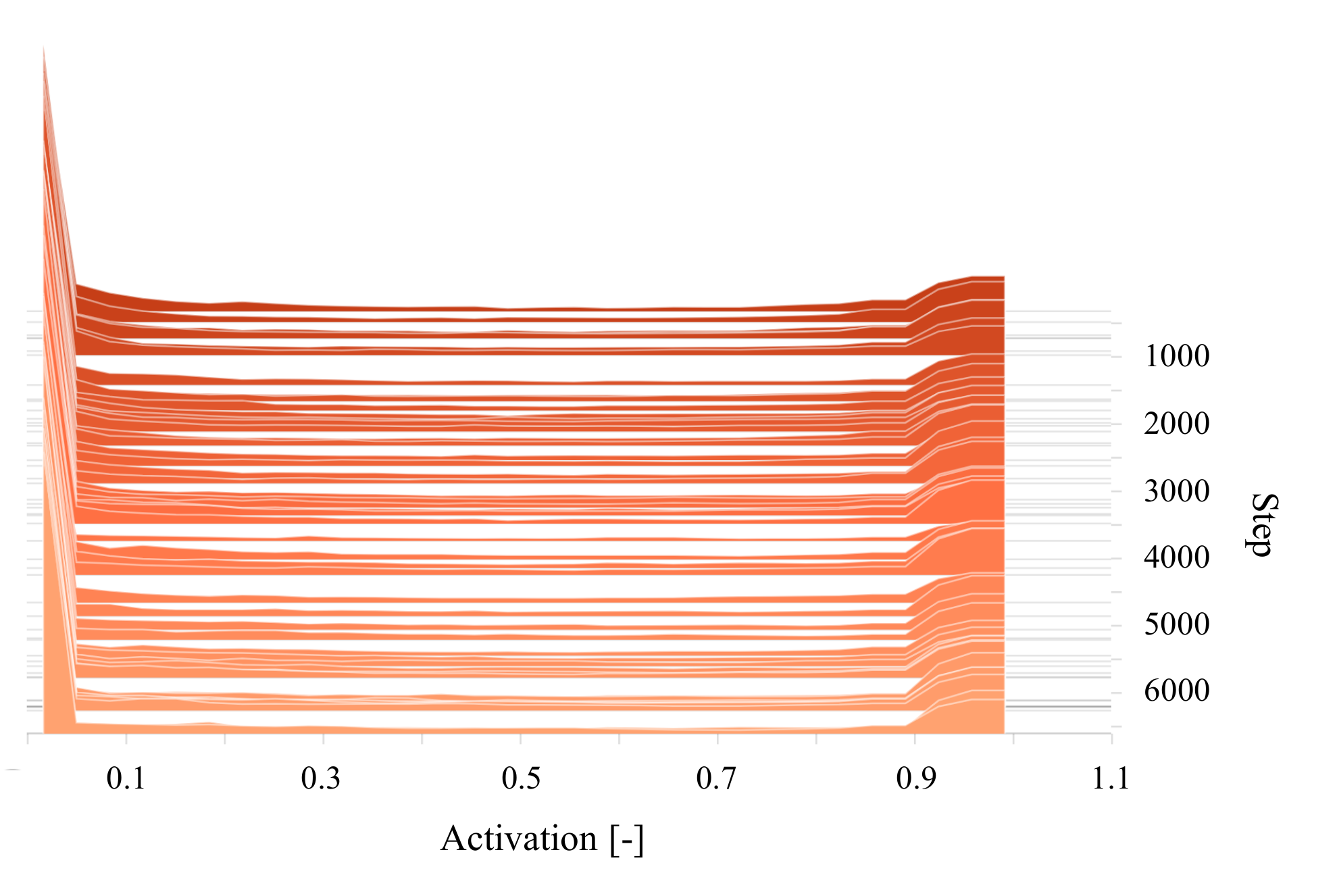}
\caption{3D Histogram of the average activations for the original feed-forward denoising autoencoder with 10~\% corruption rate.}
\label{fig: histo1}
\end{figure}

In order to guarantee an effective model training, all the previous approaches followed in the literature were implemented \cite{glorot,maas,ioffe}.
Figure \ref{fig: histo1} shows the histogram of the average activations for the original feed-forward denoising autoencoder configuration with classical uniform weights initialization, sigmoid activation functions and no batch-normalization. It is possible to observe the typical network saturation problem, being the average activations always concentrated at 0 and 1, namely the endpoints of a sigmoid activation function.

\begin{figure}[H]
\centering
\includegraphics[height=0.30\textheight]{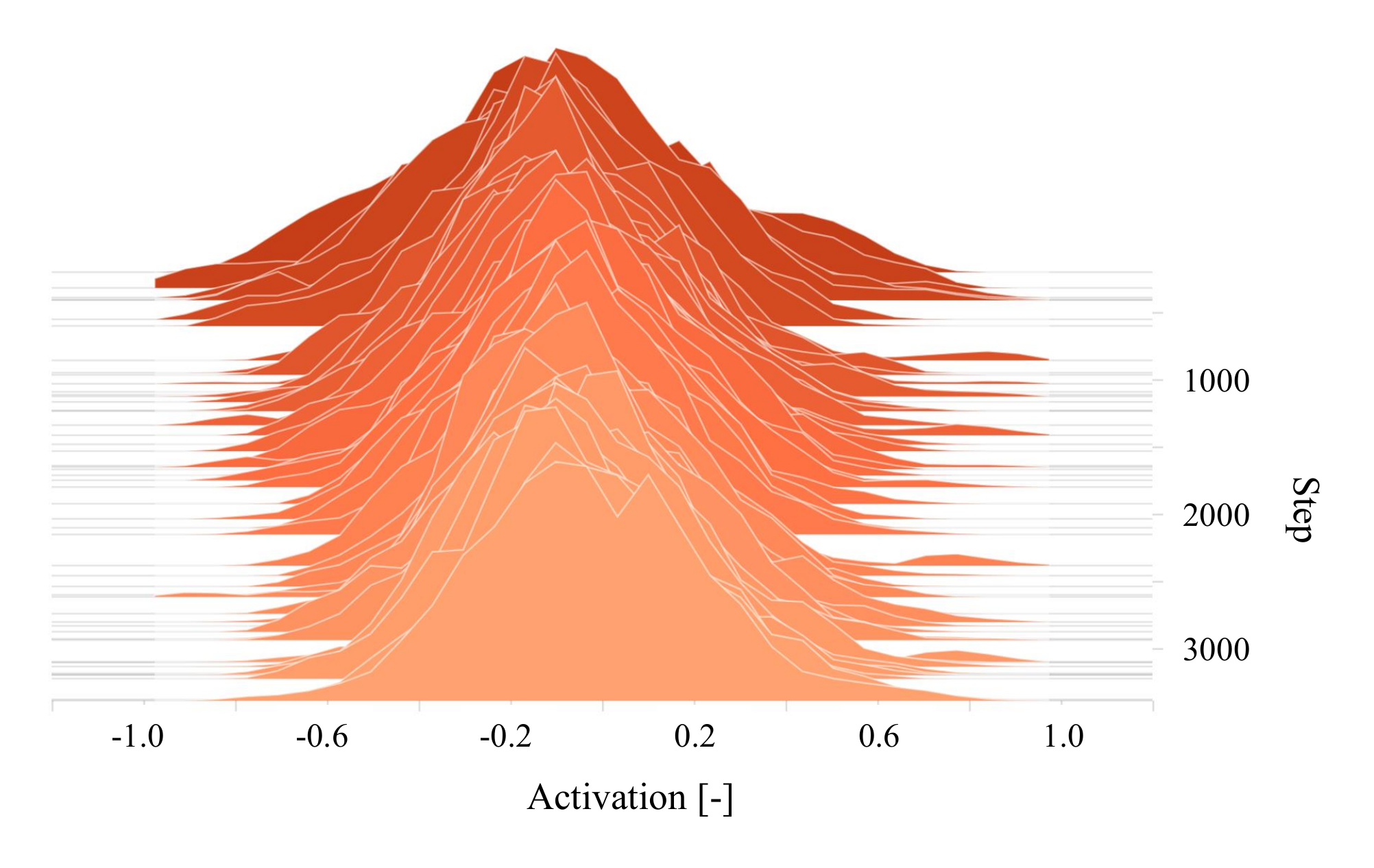}
\caption{3D Histogram of the average activations for the new feed-forward denoising autoencoder with 10~\% corruption rate.}
\label{fig: histo2}
\end{figure}

The use of normalized initialization, ReLU activation function and batch-normalization as proposed in the literature confirmed to reach higher performance and to overcome saturation problems. Figure \ref{fig: histo2} shows how average network activations change during training, not being stuck in any saturating point. Note that ReLU activation function was applied on the encoder layer while batch-normalization only in the decoder layer. The authors realized that higher performance could be reached using a $tanh$ activation function in the decoder layer, in association with batch-normalization. Conclusively, the SAT results confirmed the superiority of the chosen models' configurations with respect to the original denoising autoencoder architectures with classical uniform weights
initialization, sigmoid activation functions and no batch-normalization. 

The saturation performance results of the proposed models for different CR are summarized in Table \ref{tab:SAT}. Here, network saturation is defined as the SAT lower than 0.1 \cite{RoM1}. Every autoencoder was well above the previous defined limit, meaning that the adopted architecture strategies could efficiently overcome saturation issues. The above identified metric was function of the particular variable and of the applied CR. While the SAT was insensitive to changes in CR for relative humidity, it decreased for both temperature and $CO_2$ data.

\begin{table}[H]
\centering
\caption{SAT for denoising autoencoder neural networks for different CR. "CONV", "FEED", "LSTM" and "Avg" stand for convolutional, feed-forward, LSTM denoising autoencoder and average.}
\label{tab:SAT}
\begin{tabular}{cllllllllll}
 \toprule  
\multicolumn{1}{c}{} & \multicolumn{1}{c}{} & \multicolumn{3}{l}{$T$ [-]}  & \multicolumn{3}{l}{$RH$ [-]} & \multicolumn{3}{l}{$CO_2$ [-]} \\
\cmidrule{3-11}
\multicolumn{1}{l}{} & \multicolumn{1}{r}{\multirow{-2}{*}{CR [-]}} & CONV     & FEED     & LSTM    & CONV     & FEED     & LSTM     & CONV      & FEED     & LSTM     \\
 \toprule 
    \multirow{10}{*}{SAT}& \multicolumn{1}{r}{0.10}& \multicolumn{1}{r}{0.88}& \multicolumn{1}{r}{0.85}& \multicolumn{1}{r}{0.84}& \multicolumn{1}{r}{0.95}& \multicolumn{1}{r}{0.97}& \multicolumn{1}{r}{0.98}& \multicolumn{1}{r}{0.78}& \multicolumn{1}{r}{0.83}& \multicolumn{1}{r}{0.77}\\
                    \multicolumn{1}{r}{}& \multicolumn{1}{r}{0.20}& \multicolumn{1}{r}{0.87}& \multicolumn{1}{r}{0.83}& \multicolumn{1}{r}{0.79}& \multicolumn{1}{r}{0.95}& \multicolumn{1}{r}{0.98}& \multicolumn{1}{r}{0.98}& \multicolumn{1}{r}{0.71}& \multicolumn{1}{r}{0.77}& \multicolumn{1}{r}{0.76}\\
                    \multicolumn{1}{r}{}& \multicolumn{1}{r}{0.30}& \multicolumn{1}{r}{0.83}& \multicolumn{1}{r}{0.82}& \multicolumn{1}{r}{0.77}& \multicolumn{1}{r}{1.00}& \multicolumn{1}{r}{0.98}& \multicolumn{1}{r}{0.98}& \multicolumn{1}{r}{0.65}& \multicolumn{1}{r}{0.73}& \multicolumn{1}{r}{0.69}\\
                    \multicolumn{1}{r}{}& \multicolumn{1}{r}{0.40}& \multicolumn{1}{r}{0.77}& \multicolumn{1}{r}{0.75}& \multicolumn{1}{r}{0.73}& \multicolumn{1}{r}{0.99}& \multicolumn{1}{r}{0.96}& \multicolumn{1}{r}{0.99}& \multicolumn{1}{r}{0.75}& \multicolumn{1}{r}{0.66}& \multicolumn{1}{r}{0.59}\\
                    \multicolumn{1}{r}{}& \multicolumn{1}{r}{0.50}& \multicolumn{1}{r}{0.73}& \multicolumn{1}{r}{0.72}& \multicolumn{1}{r}{0.72}& \multicolumn{1}{r}{0.98}& \multicolumn{1}{r}{0.97}& \multicolumn{1}{r}{0.98}& \multicolumn{1}{r}{0.57}& \multicolumn{1}{r}{0.60}& \multicolumn{1}{r}{0.47}\\
                    \multicolumn{1}{r}{}& \multicolumn{1}{r}{0.60}& \multicolumn{1}{r}{0.73}& \multicolumn{1}{r}{0.68}& \multicolumn{1}{r}{0.65}& \multicolumn{1}{r}{0.97}& \multicolumn{1}{r}{0.95}& \multicolumn{1}{r}{0.95}& \multicolumn{1}{r}{0.52}& \multicolumn{1}{r}{0.46}& \multicolumn{1}{r}{0.35}\\
                    \multicolumn{1}{r}{}& \multicolumn{1}{r}{0.70}& \multicolumn{1}{r}{0.69}& \multicolumn{1}{r}{0.69}& \multicolumn{1}{r}{0.59}& \multicolumn{1}{r}{0.94}& \multicolumn{1}{r}{0.96}& \multicolumn{1}{r}{0.96}& \multicolumn{1}{r}{0.51}& \multicolumn{1}{r}{0.48}& \multicolumn{1}{r}{0.32}\\
                    \multicolumn{1}{r}{}& \multicolumn{1}{r}{0.80}& \multicolumn{1}{r}{0.64}& \multicolumn{1}{r}{0.64}& \multicolumn{1}{r}{0.60}& \multicolumn{1}{r}{0.92}& \multicolumn{1}{r}{0.94}& \multicolumn{1}{r}{0.94}& \multicolumn{1}{r}{0.44}& \multicolumn{1}{r}{0.41}& \multicolumn{1}{r}{0.30}\\
                    \multicolumn{1}{r}{}& \multicolumn{1}{r}{0.90}& \multicolumn{1}{r}{0.62}& \multicolumn{1}{r}{0.63}& \multicolumn{1}{r}{0.57}& \multicolumn{1}{r}{0.93}& \multicolumn{1}{r}{0.96}& \multicolumn{1}{r}{0.93}& \multicolumn{1}{r}{0.39}& \multicolumn{1}{r}{0.34}& \multicolumn{1}{r}{0.31}\\

    \cmidrule{3-11}
     &\multicolumn{1}{r}{Avg} & \multicolumn{1}{r}{0.75}& \multicolumn{1}{r}{0.73}& \multicolumn{1}{r}{0.70}& \multicolumn{1}{r}{0.96}& \multicolumn{1}{r}{0.96}& \multicolumn{1}{r}{0.97}& \multicolumn{1}{r}{0.59}& \multicolumn{1}{r}{0.59}& \multicolumn{1}{r}{0.51}\\
 \toprule  
\end{tabular}
\end{table}

\Needspace{5\baselineskip}
\subsection{Data reconstruction performance evaluation}
\label{sec: sub_daily}
		
Firstly, the ability of autoencoder neural networks for capturing daily patterns of environmental data is assessed. Eventually, the proposed models' performance in reconstructing sub-daily indoor environment data gaps are evaluated and compared to classic polynomial interpolations. \par
\begin{figure}[H]

\centering

\includegraphics[width=.27\textwidth]{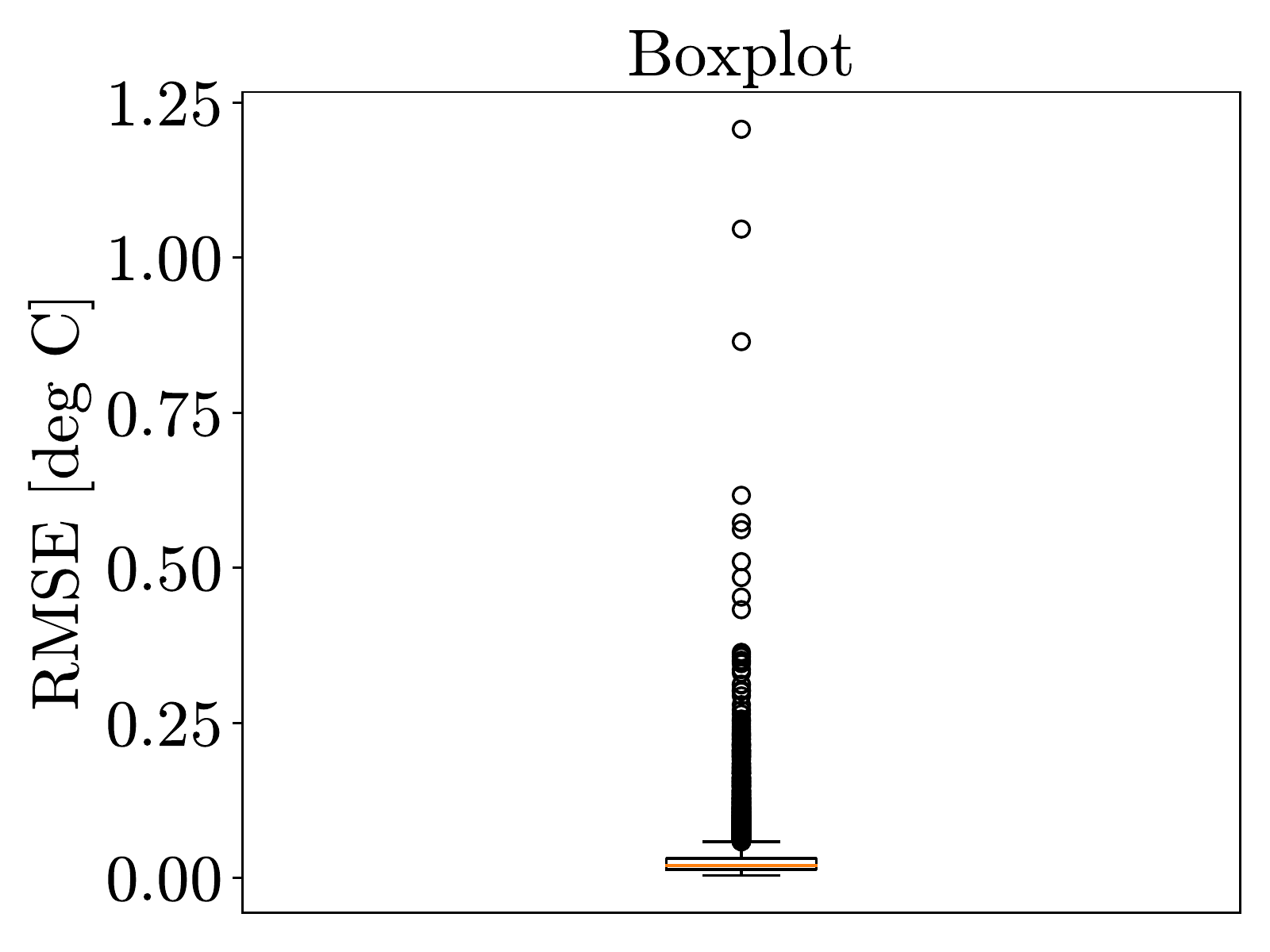}\hfill
\includegraphics[width=.27\textwidth]{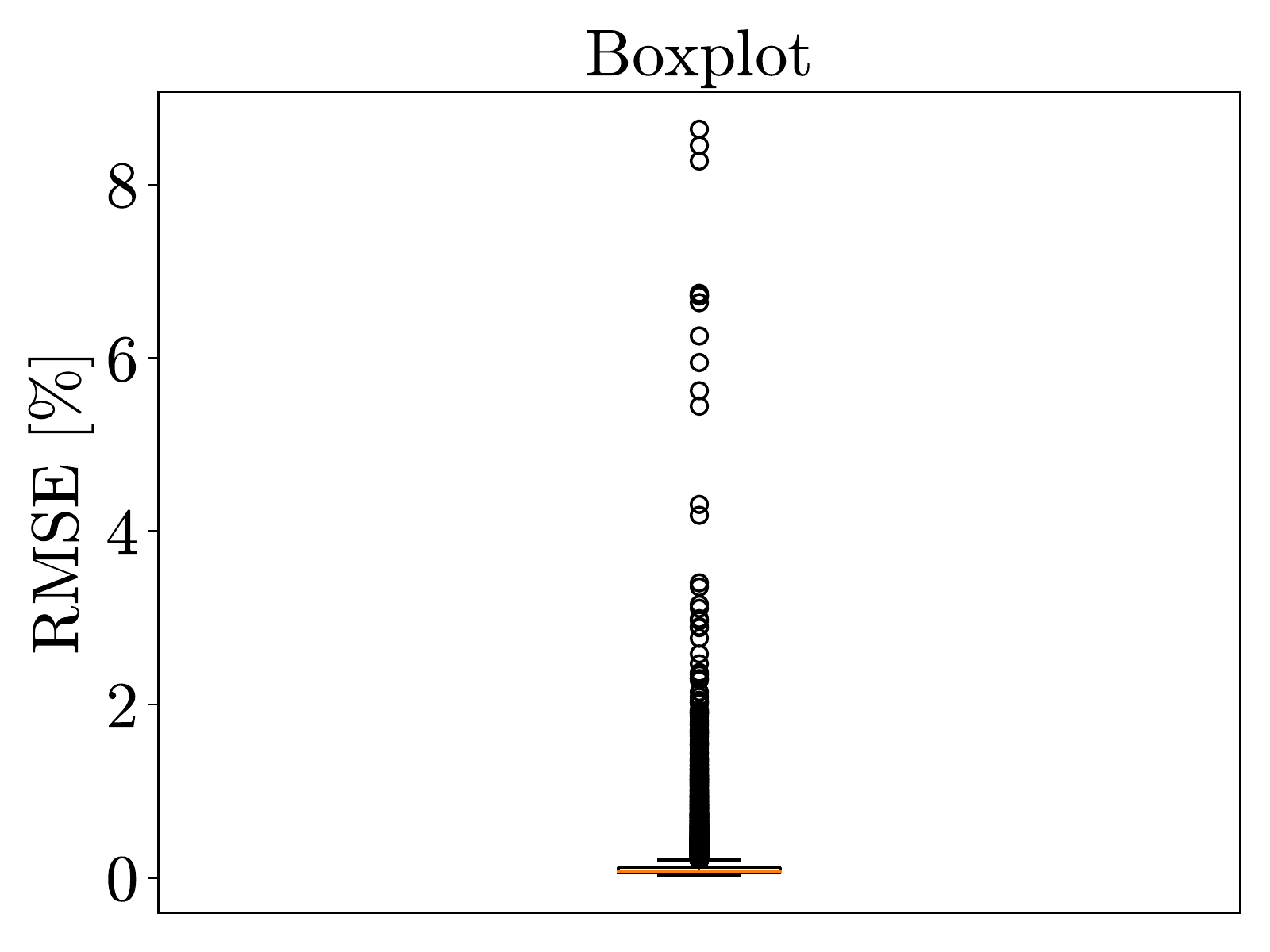}\hfill
\includegraphics[width=.27\textwidth]{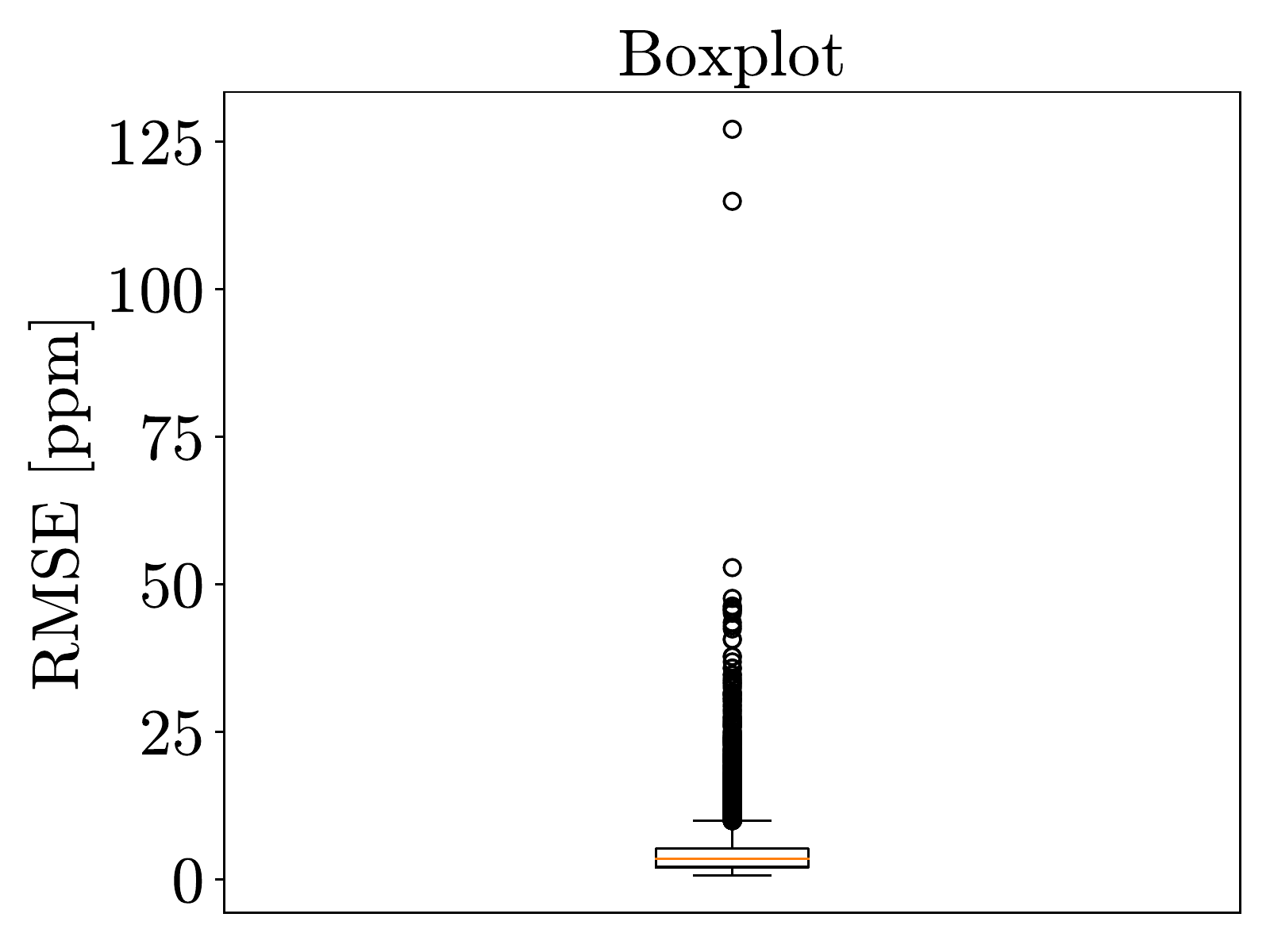}
\includegraphics[width=.27\textwidth]{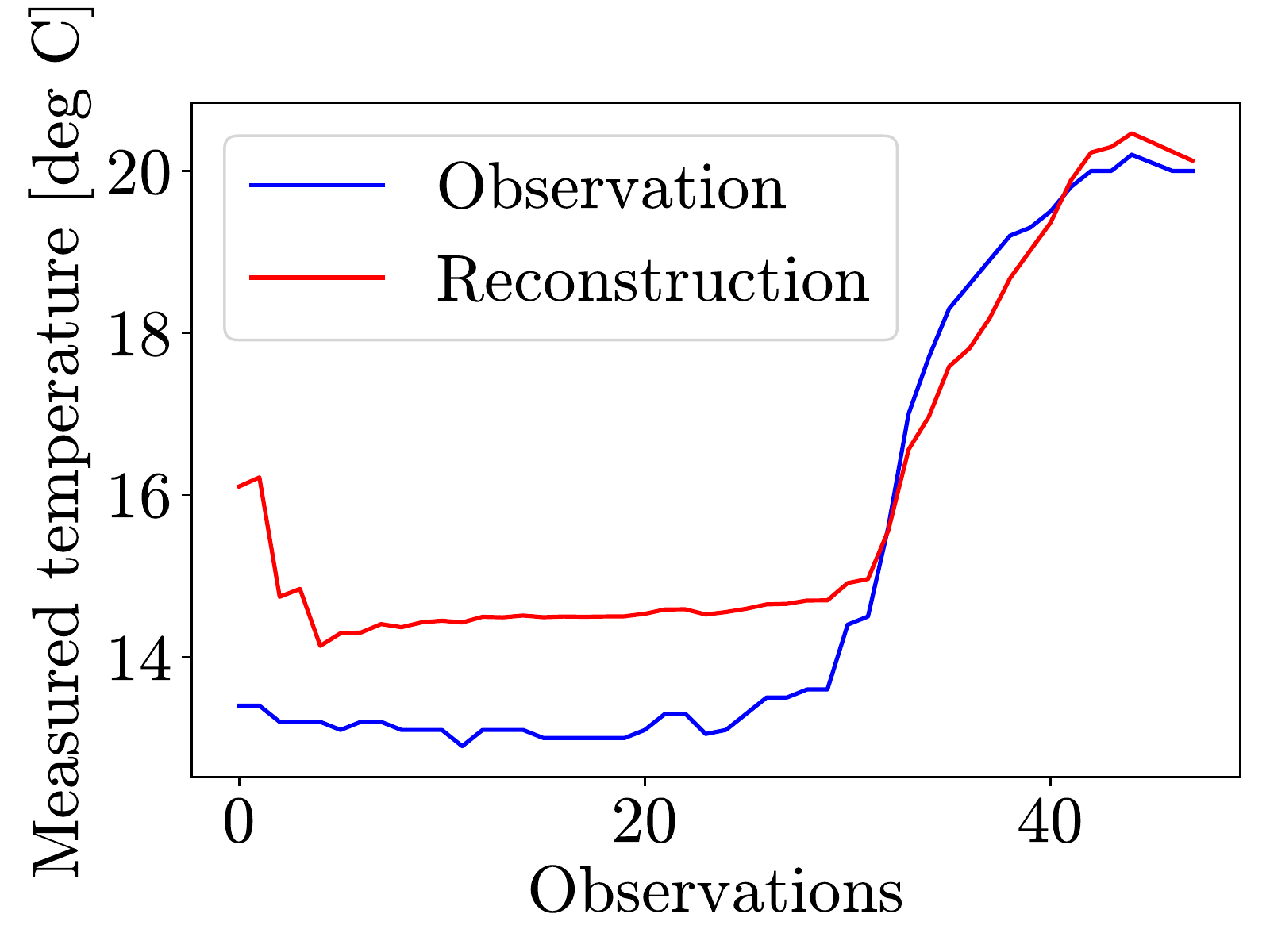}\hfill
\includegraphics[width=.27\textwidth]{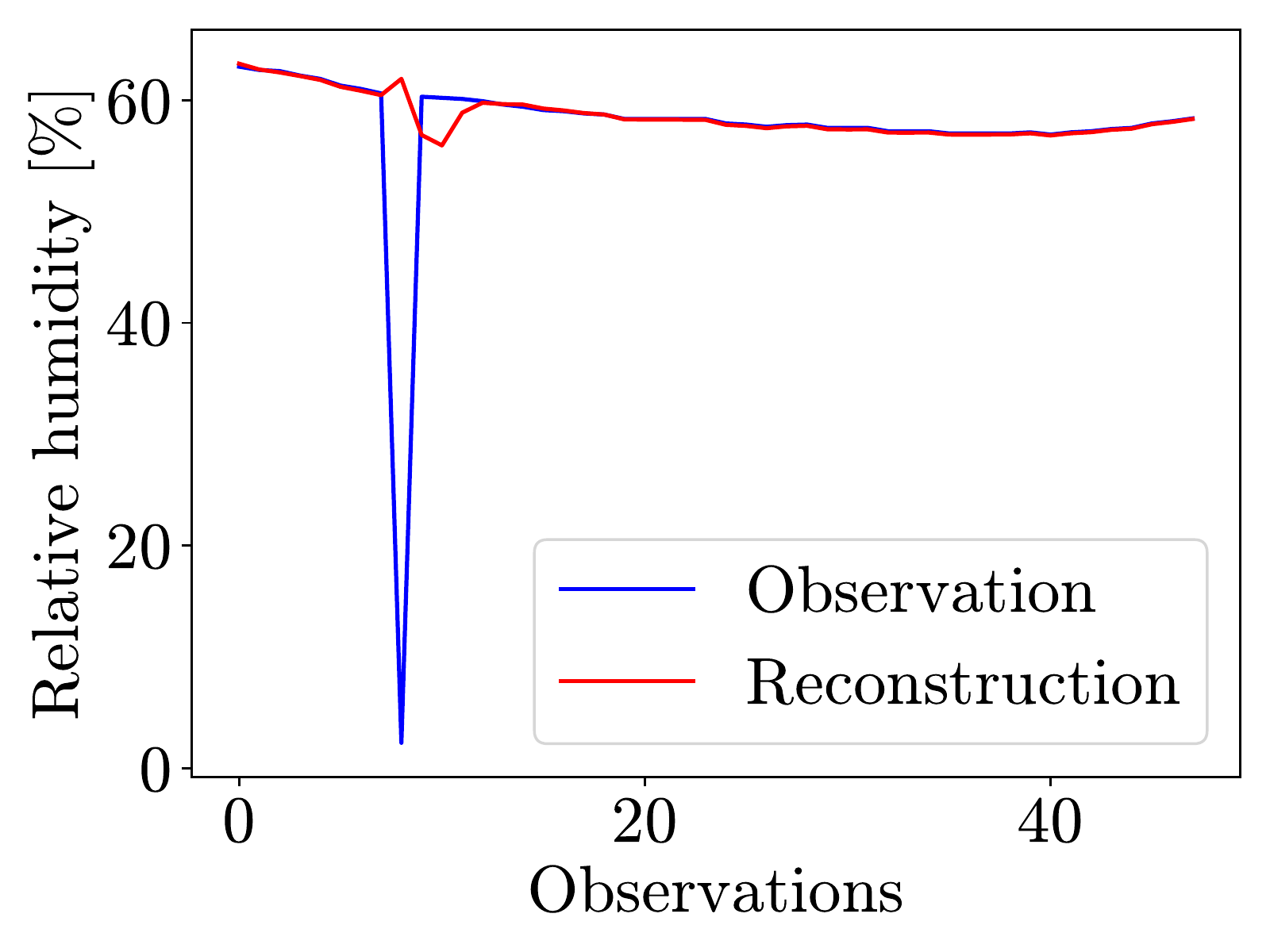}\hfill
\includegraphics[width=.27\textwidth]{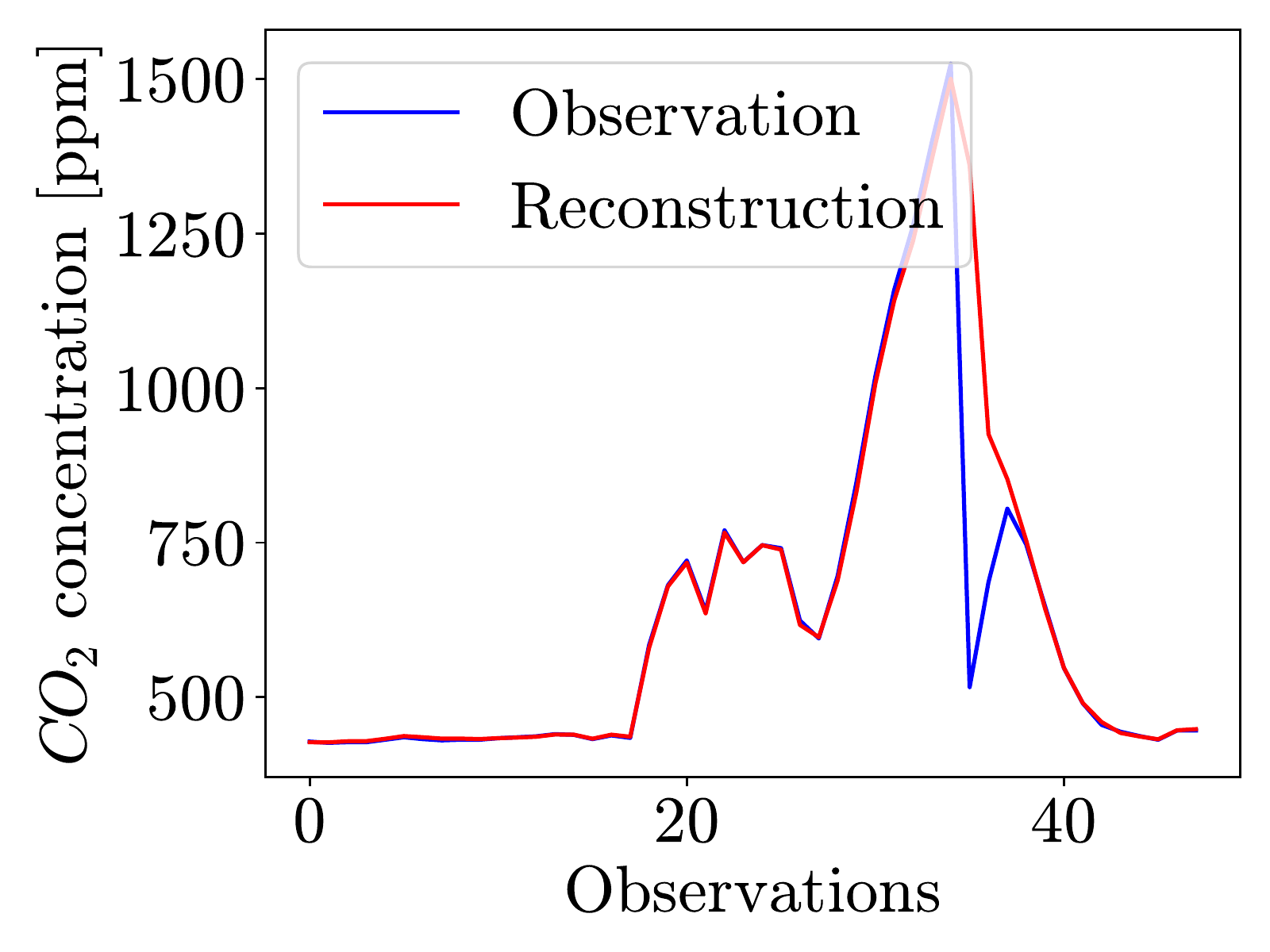}
\caption{Boxplots of of the observed sequences and reconstruction of a day with atypical behavior. }
\label{fig:patterns}

\end{figure}

Figure \ref{fig:patterns} shows the RMSE variance for a LSTM denoising autoencoder. In particular, the model was trained with a masking noise of 0.1 and applied to the non corrupted evaluation sets. Here, the RMSE was computed on the single sequence as seen in Section \ref{sec: metrics}. The average RMSEs over the number of evaluation sequences were 0.027 \textdegree C, 0.12~\% and 4.25 ppm, respectively for $T$, $RH$ and $CO_2$ data. Furthermore, all the reconstruction residuals computed with the LSTM architecture were lower than the convolutional and feed-forward ones. \par
Based on the boxplots presented in Figure \ref{fig:patterns}, the observed days with RMSE out of the measured IQR are represented. These could be considered as sequences with atypical behavior, for which the indoor environment data patterns cannot be detected by the model. Figure \ref{fig:patterns} shows also, for every variable, how a random day with atypical behavior looks like and how it is reconstructed by the LSTM autoencoder. In particular, the represented sequence of relative humidity data presents an outlier in the early morning observation. This anomaly, which could be caused by sensors' malfunctioning, was not detected during the data preprocessing, but it was identified with the proposed model.

Table \ref{tab:subdaily} summarizes the behavior of all the models, at different CR, when applied to the corrupted data. For every variable there was a polynomial degree for which the daily data trends were better fit by the interpolation. While indoor air temperature data were more accurately described by cubic correlations, relative humidity and $CO_2$ concentration data had, respectively, more linear and quadratic trends. However, all autoencoders performed by a large margin better than baseline approaches for all variables. In particular, the performance of the convolutional configuration outperformed, in average, all the alternative models. In this regard, the RMSE was 37~\% lower than cubic interpolation for indoor air temperature, 24~\% lower than linear interpolation for relative humidity and 30~\% lower than quadratic interpolation for $CO_2$ concentration. In terms of NRMSE, missing relative humidity data could be reconstructed with higher accuracy, when compared to other variables. On that place, the worst behavior was obtained with $CO_2$ data. In particular, the NRMSE of the convolutional configuration in case of $RH$ data was 75~\% lower than $T$ and 90~\% lower than $CO_2$. These results were consistent with the SAT trend, being higher for $RH$ and lower for $CO_2$ (Table \ref{tab:SAT}). 

\begin{table}[H]
\centering
\caption{Performance of denoising autoencoder neural networks and polynomial interpolations for reconstructing sub-daily indoor environment data gaps. "CONV", "FEED", "LSTM" stand for convolutional, feed-forward and LSTM denoising autoencoder. "LIN", "QUAD" and "CUB" stand for linear, quadratic and cubic interpolation. "Avg" stands for average.}
\label{tab:subdaily}
\begin{tabular}{cllllllllll}
                                 \\
\toprule
\multicolumn{1}{l}{} &
  \multicolumn{1}{r}{\multirow{2}{*}{CR [-]}} &
  \multicolumn{3}{l}{$T$ [\textdegree C]} & \multicolumn{3}{l}{$RH$ [\%]} & \multicolumn{3}{l}{$CO_2$ [ppm]} \\
    \cmidrule{3-11}
\multicolumn{1}{l}{} &
  \multicolumn{1}{c}{} &
  CONV &
  FEED &
  LSTM &
  CONV &
  FEED &
  LSTM &
  CONV &
  FEED &
  LSTM \\
  \toprule
\multirow{10}{*}{RMSE}& \multicolumn{1}{r}{0.10}& \multicolumn{1}{r}{0.22}& \multicolumn{1}{r}{0.32}& \multicolumn{1}{r}{0.33}& \multicolumn{1}{r}{0.73}& \multicolumn{1}{r}{1.14}& \multicolumn{1}{r}{1.05}& \multicolumn{1}{r}{49.10}& \multicolumn{1}{r}{64.58}& \multicolumn{1}{r}{64.88}\\
                    \multicolumn{1}{r}{}& \multicolumn{1}{r}{0.20}& \multicolumn{1}{r}{0.31}& \multicolumn{1}{r}{0.36}& \multicolumn{1}{r}{0.47}& \multicolumn{1}{r}{0.90}& \multicolumn{1}{r}{1.23}& \multicolumn{1}{r}{1.47}& \multicolumn{1}{r}{61.16}& \multicolumn{1}{r}{69.41}& \multicolumn{1}{r}{82.51}\\
                    \multicolumn{1}{r}{}& \multicolumn{1}{r}{0.30}& \multicolumn{1}{r}{0.36}& \multicolumn{1}{r}{0.42}& \multicolumn{1}{r}{0.53}& \multicolumn{1}{r}{1.08}& \multicolumn{1}{r}{1.35}& \multicolumn{1}{r}{1.78}& \multicolumn{1}{r}{69.11}& \multicolumn{1}{r}{75.04}& \multicolumn{1}{r}{89.00}\\
                    \multicolumn{1}{r}{}& \multicolumn{1}{r}{0.40}& \multicolumn{1}{r}{0.41}& \multicolumn{1}{r}{0.47}& \multicolumn{1}{r}{0.59}& \multicolumn{1}{r}{1.23}& \multicolumn{1}{r}{1.45}& \multicolumn{1}{r}{2.11}& \multicolumn{1}{r}{78.25}& \multicolumn{1}{r}{81.46}& \multicolumn{1}{r}{101.64}\\
                    \multicolumn{1}{r}{}& \multicolumn{1}{r}{0.50}& \multicolumn{1}{r}{0.46}& \multicolumn{1}{r}{0.51}& \multicolumn{1}{r}{0.62}& \multicolumn{1}{r}{1.36}& \multicolumn{1}{r}{1.59}& \multicolumn{1}{r}{2.33}& \multicolumn{1}{r}{84.12}& \multicolumn{1}{r}{89.98}& \multicolumn{1}{r}{107.85}\\
                    \multicolumn{1}{r}{}& \multicolumn{1}{r}{0.60}& \multicolumn{1}{r}{0.50}& \multicolumn{1}{r}{0.53}& \multicolumn{1}{r}{0.64}& \multicolumn{1}{r}{1.48}& \multicolumn{1}{r}{1.66}& \multicolumn{1}{r}{2.54}& \multicolumn{1}{r}{91.59}& \multicolumn{1}{r}{95.38}& \multicolumn{1}{r}{110.28}\\
                    \multicolumn{1}{r}{}& \multicolumn{1}{r}{0.70}& \multicolumn{1}{r}{0.51}& \multicolumn{1}{r}{0.52}& \multicolumn{1}{r}{0.63}& \multicolumn{1}{r}{1.54}& \multicolumn{1}{r}{1.66}& \multicolumn{1}{r}{2.72}& \multicolumn{1}{r}{91.43}& \multicolumn{1}{r}{94.86}& \multicolumn{1}{r}{106.16}\\
                    \multicolumn{1}{r}{}& \multicolumn{1}{r}{0.80}& \multicolumn{1}{r}{0.50}& \multicolumn{1}{r}{0.51}& \multicolumn{1}{r}{0.61}& \multicolumn{1}{r}{1.66}& \multicolumn{1}{r}{1.74}& \multicolumn{1}{r}{2.80}& \multicolumn{1}{r}{91.26}& \multicolumn{1}{r}{94.32}& \multicolumn{1}{r}{102.18}\\
                    \multicolumn{1}{r}{}& \multicolumn{1}{r}{0.90}& \multicolumn{1}{r}{0.50}& \multicolumn{1}{r}{0.49}& \multicolumn{1}{r}{0.60}& \multicolumn{1}{r}{1.73}& \multicolumn{1}{r}{1.78}& \multicolumn{1}{r}{3.00}& \multicolumn{1}{r}{89.63}& \multicolumn{1}{r}{90.52}& \multicolumn{1}{r}{96.75}\\
                    \cmidrule{3-11}
                                & \multicolumn{1}{r}{Avg} & \multicolumn{1}{r}{0.42}& \multicolumn{1}{r}{0.46}& \multicolumn{1}{r}{0.56}& \multicolumn{1}{r}{1.30}& \multicolumn{1}{r}{1.51}& \multicolumn{1}{r}{2.20}& \multicolumn{1}{r}{78.41}& \multicolumn{1}{r}{83.95}& \multicolumn{1}{r}{95.69}\\
                                \toprule
\multirow{10}{*}{NRMSE [-]}& \multicolumn{1}{r}{0.10}& \multicolumn{1}{r}{0.15}& \multicolumn{1}{r}{0.21}& \multicolumn{1}{r}{0.22}& \multicolumn{1}{r}{0.04}& \multicolumn{1}{r}{0.07}& \multicolumn{1}{r}{0.06}& \multicolumn{1}{r}{0.43}& \multicolumn{1}{r}{0.56}& \multicolumn{1}{r}{0.56}\\
                    \multicolumn{1}{r}{}& \multicolumn{1}{r}{0.20}& \multicolumn{1}{r}{0.21}& \multicolumn{1}{r}{0.24}& \multicolumn{1}{r}{0.31}& \multicolumn{1}{r}{0.05}& \multicolumn{1}{r}{0.07}& \multicolumn{1}{r}{0.08}& \multicolumn{1}{r}{0.53}& \multicolumn{1}{r}{0.61}& \multicolumn{1}{r}{0.72}\\
                    \multicolumn{1}{r}{}& \multicolumn{1}{r}{0.30}& \multicolumn{1}{r}{0.24}& \multicolumn{1}{r}{0.28}& \multicolumn{1}{r}{0.35}& \multicolumn{1}{r}{0.06}& \multicolumn{1}{r}{0.08}& \multicolumn{1}{r}{0.10}& \multicolumn{1}{r}{0.60}& \multicolumn{1}{r}{0.66}& \multicolumn{1}{r}{0.78}\\
                    \multicolumn{1}{r}{}& \multicolumn{1}{r}{0.40}& \multicolumn{1}{r}{0.27}& \multicolumn{1}{r}{0.31}& \multicolumn{1}{r}{0.39}& \multicolumn{1}{r}{0.07}& \multicolumn{1}{r}{0.08}& \multicolumn{1}{r}{0.12}& \multicolumn{1}{r}{0.68}& \multicolumn{1}{r}{0.71}& \multicolumn{1}{r}{0.88}\\
                    \multicolumn{1}{r}{}& \multicolumn{1}{r}{0.50}& \multicolumn{1}{r}{0.31}& \multicolumn{1}{r}{0.34}& \multicolumn{1}{r}{0.41}& \multicolumn{1}{r}{0.08}& \multicolumn{1}{r}{0.09}& \multicolumn{1}{r}{0.13}& \multicolumn{1}{r}{0.73}& \multicolumn{1}{r}{0.79}& \multicolumn{1}{r}{0.94}\\
                    \multicolumn{1}{r}{}& \multicolumn{1}{r}{0.60}& \multicolumn{1}{r}{0.33}& \multicolumn{1}{r}{0.35}& \multicolumn{1}{r}{0.42}& \multicolumn{1}{r}{0.09}& \multicolumn{1}{r}{0.10}& \multicolumn{1}{r}{0.14}& \multicolumn{1}{r}{0.80}& \multicolumn{1}{r}{0.83}& \multicolumn{1}{r}{0.96}\\
                    \multicolumn{1}{r}{}& \multicolumn{1}{r}{0.70}& \multicolumn{1}{r}{0.34}& \multicolumn{1}{r}{0.35}& \multicolumn{1}{r}{0.42}& \multicolumn{1}{r}{0.09}& \multicolumn{1}{r}{0.10}& \multicolumn{1}{r}{0.15}& \multicolumn{1}{r}{0.80}& \multicolumn{1}{r}{0.83}& \multicolumn{1}{r}{0.92}\\
                    \multicolumn{1}{r}{}& \multicolumn{1}{r}{0.80}& \multicolumn{1}{r}{0.33}& \multicolumn{1}{r}{0.34}& \multicolumn{1}{r}{0.41}& \multicolumn{1}{r}{0.10}& \multicolumn{1}{r}{0.10}& \multicolumn{1}{r}{0.16}& \multicolumn{1}{r}{0.80}& \multicolumn{1}{r}{0.82}& \multicolumn{1}{r}{0.89}\\
                    \multicolumn{1}{r}{}& \multicolumn{1}{r}{0.90}& \multicolumn{1}{r}{0.33}& \multicolumn{1}{r}{0.33}& \multicolumn{1}{r}{0.40}& \multicolumn{1}{r}{0.10}& \multicolumn{1}{r}{0.10}& \multicolumn{1}{r}{0.17}& \multicolumn{1}{r}{0.78}& \multicolumn{1}{r}{0.79}& \multicolumn{1}{r}{0.84}\\
                                   \cmidrule{3-11}
                                & \multicolumn{1}{r}{Avg} & \multicolumn{1}{r}{0.28}& \multicolumn{1}{r}{0.31}& \multicolumn{1}{r}{0.37}& \multicolumn{1}{r}{0.07}& \multicolumn{1}{r}{0.09}& \multicolumn{1}{r}{0.12}& \multicolumn{1}{r}{0.68}& \multicolumn{1}{r}{0.73}& \multicolumn{1}{r}{0.83}\\
                                \toprule
\multicolumn{1}{l}{}            & \multicolumn{1}{c}{}        & LIN  & QUAD & CUB  & LIN  & QUAD & CUB  & LIN    & QUAD   & CUB    \\
\toprule
\multirow{10}{*}{RMSE}& \multicolumn{1}{r}{0.10}& \multicolumn{1}{r}{0.58}& \multicolumn{1}{r}{0.52}& \multicolumn{1}{r}{0.39}& \multicolumn{1}{r}{1.31}& \multicolumn{1}{r}{1.10}& \multicolumn{1}{r}{0.94}& \multicolumn{1}{r}{94.89}& \multicolumn{1}{r}{79.97}& \multicolumn{1}{r}{72.00}\\
                    \multicolumn{1}{r}{}& \multicolumn{1}{r}{0.20}& \multicolumn{1}{r}{0.67}& \multicolumn{1}{r}{0.64}& \multicolumn{1}{r}{0.49}& \multicolumn{1}{r}{1.48}& \multicolumn{1}{r}{1.28}& \multicolumn{1}{r}{1.13}& \multicolumn{1}{r}{106.67}& \multicolumn{1}{r}{92.86}& \multicolumn{1}{r}{85.32}\\
                    \multicolumn{1}{r}{}& \multicolumn{1}{r}{0.30}& \multicolumn{1}{r}{0.73}& \multicolumn{1}{r}{0.73}& \multicolumn{1}{r}{0.57}& \multicolumn{1}{r}{1.59}& \multicolumn{1}{r}{1.45}& \multicolumn{1}{r}{1.29}& \multicolumn{1}{r}{114.41}& \multicolumn{1}{r}{102.74}& \multicolumn{1}{r}{96.76}\\
                    \multicolumn{1}{r}{}& \multicolumn{1}{r}{0.40}& \multicolumn{1}{r}{0.78}& \multicolumn{1}{r}{0.82}& \multicolumn{1}{r}{0.66}& \multicolumn{1}{r}{1.70}& \multicolumn{1}{r}{1.59}& \multicolumn{1}{r}{1.49}& \multicolumn{1}{r}{118.96}& \multicolumn{1}{r}{113.61}& \multicolumn{1}{r}{113.68}\\
                    \multicolumn{1}{r}{}& \multicolumn{1}{r}{0.50}& \multicolumn{1}{r}{0.80}& \multicolumn{1}{r}{0.91}& \multicolumn{1}{r}{0.74}& \multicolumn{1}{r}{1.77}& \multicolumn{1}{r}{1.76}& \multicolumn{1}{r}{1.69}& \multicolumn{1}{r}{121.94}& \multicolumn{1}{r}{122.14}& \multicolumn{1}{r}{125.89}\\
                    \multicolumn{1}{r}{}& \multicolumn{1}{r}{0.60}& \multicolumn{1}{r}{0.79}& \multicolumn{1}{r}{0.95}& \multicolumn{1}{r}{0.82}& \multicolumn{1}{r}{1.82}& \multicolumn{1}{r}{1.89}& \multicolumn{1}{r}{1.89}& \multicolumn{1}{r}{121.25}& \multicolumn{1}{r}{124.12}& \multicolumn{1}{r}{148.58}\\
                    \multicolumn{1}{r}{}& \multicolumn{1}{r}{0.70}& \multicolumn{1}{r}{0.77}& \multicolumn{1}{r}{0.95}& \multicolumn{1}{r}{0.82}& \multicolumn{1}{r}{1.86}& \multicolumn{1}{r}{2.01}& \multicolumn{1}{r}{2.03}& \multicolumn{1}{r}{120.32}& \multicolumn{1}{r}{122.30}& \multicolumn{1}{r}{138.57}\\
                    \multicolumn{1}{r}{}& \multicolumn{1}{r}{0.80}& \multicolumn{1}{r}{0.75}& \multicolumn{1}{r}{0.92}& \multicolumn{1}{r}{0.70}& \multicolumn{1}{r}{1.88}& \multicolumn{1}{r}{2.21}& \multicolumn{1}{r}{2.31}& \multicolumn{1}{r}{117.48}& \multicolumn{1}{r}{122.40}& \multicolumn{1}{r}{120.25}\\
                    \multicolumn{1}{r}{}& \multicolumn{1}{r}{0.90}& \multicolumn{1}{r}{0.74}& \multicolumn{1}{r}{0.86}& \multicolumn{1}{r}{0.77}& \multicolumn{1}{r}{1.91}& \multicolumn{1}{r}{3.15}& \multicolumn{1}{r}{3.24}& \multicolumn{1}{r}{112.40}& \multicolumn{1}{r}{126.33}& \multicolumn{1}{r}{125.01}\\
                          \cmidrule{3-11}
                                & \multicolumn{1}{r}{Avg} & \multicolumn{1}{r}{0.73}& \multicolumn{1}{r}{0.81}& \multicolumn{1}{r}{0.66}& \multicolumn{1}{r}{1.70}& \multicolumn{1}{r}{1.83}& \multicolumn{1}{r}{1.78}& \multicolumn{1}{r}{114.26}& \multicolumn{1}{r}{111.83}& \multicolumn{1}{r}{114.01}\\
                                \toprule
\multirow{10}{*}{NRMSE [-]}& \multicolumn{1}{r}{0.10}& \multicolumn{1}{r}{0.39}& \multicolumn{1}{r}{0.34}& \multicolumn{1}{r}{0.27}& \multicolumn{1}{r}{0.07}& \multicolumn{1}{r}{0.06}& \multicolumn{1}{r}{0.05}& \multicolumn{1}{r}{0.83}& \multicolumn{1}{r}{0.69}& \multicolumn{1}{r}{0.63}\\
                    \multicolumn{1}{r}{}& \multicolumn{1}{r}{0.20}& \multicolumn{1}{r}{0.45}& \multicolumn{1}{r}{0.42}& \multicolumn{1}{r}{0.33}& \multicolumn{1}{r}{0.08}& \multicolumn{1}{r}{0.07}& \multicolumn{1}{r}{0.06}& \multicolumn{1}{r}{0.93}& \multicolumn{1}{r}{0.81}& \multicolumn{1}{r}{0.74}\\
                    \multicolumn{1}{r}{}& \multicolumn{1}{r}{0.30}& \multicolumn{1}{r}{0.49}& \multicolumn{1}{r}{0.48}& \multicolumn{1}{r}{0.38}& \multicolumn{1}{r}{0.09}& \multicolumn{1}{r}{0.08}& \multicolumn{1}{r}{0.07}& \multicolumn{1}{r}{1.00}& \multicolumn{1}{r}{0.89}& \multicolumn{1}{r}{0.84}\\
                    \multicolumn{1}{r}{}& \multicolumn{1}{r}{0.40}& \multicolumn{1}{r}{0.52}& \multicolumn{1}{r}{0.55}& \multicolumn{1}{r}{0.44}& \multicolumn{1}{r}{0.09}& \multicolumn{1}{r}{0.09}& \multicolumn{1}{r}{0.08}& \multicolumn{1}{r}{1.03}& \multicolumn{1}{r}{0.99}& \multicolumn{1}{r}{0.99}\\
                    \multicolumn{1}{r}{}& \multicolumn{1}{r}{0.50}& \multicolumn{1}{r}{0.53}& \multicolumn{1}{r}{0.60}& \multicolumn{1}{r}{0.49}& \multicolumn{1}{r}{0.10}& \multicolumn{1}{r}{0.10}& \multicolumn{1}{r}{0.09}& \multicolumn{1}{r}{1.06}& \multicolumn{1}{r}{1.06}& \multicolumn{1}{r}{1.10}\\
                    \multicolumn{1}{r}{}& \multicolumn{1}{r}{0.60}& \multicolumn{1}{r}{0.53}& \multicolumn{1}{r}{0.63}& \multicolumn{1}{r}{0.54}& \multicolumn{1}{r}{0.10}& \multicolumn{1}{r}{0.11}& \multicolumn{1}{r}{0.11}& \multicolumn{1}{r}{1.05}& \multicolumn{1}{r}{1.08}& \multicolumn{1}{r}{1.29}\\
                    \multicolumn{1}{r}{}& \multicolumn{1}{r}{0.70}& \multicolumn{1}{r}{0.51}& \multicolumn{1}{r}{0.63}& \multicolumn{1}{r}{0.54}& \multicolumn{1}{r}{0.10}& \multicolumn{1}{r}{0.11}& \multicolumn{1}{r}{0.11}& \multicolumn{1}{r}{1.05}& \multicolumn{1}{r}{1.07}& \multicolumn{1}{r}{1.21}\\
                    \multicolumn{1}{r}{}& \multicolumn{1}{r}{0.80}& \multicolumn{1}{r}{0.51}& \multicolumn{1}{r}{0.61}& \multicolumn{1}{r}{0.46}& \multicolumn{1}{r}{0.11}& \multicolumn{1}{r}{0.12}& \multicolumn{1}{r}{0.13}& \multicolumn{1}{r}{1.02}& \multicolumn{1}{r}{1.07}& \multicolumn{1}{r}{1.05}\\
                    \multicolumn{1}{r}{}& \multicolumn{1}{r}{0.90}& \multicolumn{1}{r}{0.49}& \multicolumn{1}{r}{0.57}& \multicolumn{1}{r}{0.51}& \multicolumn{1}{r}{0.11}& \multicolumn{1}{r}{0.18}& \multicolumn{1}{r}{0.18}& \multicolumn{1}{r}{0.98}& \multicolumn{1}{r}{1.10}& \multicolumn{1}{r}{1.09}\\
                             \cmidrule{3-11}
                                & \multicolumn{1}{r}{Avg} & \multicolumn{1}{r}{0.49}& \multicolumn{1}{r}{0.54}& \multicolumn{1}{r}{0.44}& \multicolumn{1}{r}{0.09}& \multicolumn{1}{r}{0.10}& \multicolumn{1}{r}{0.10}& \multicolumn{1}{r}{0.99}& \multicolumn{1}{r}{0.97}& \multicolumn{1}{r}{0.99}\\
                                \toprule
\end{tabular}
\end{table}

\begin{figure}[H]

\centering

\includegraphics[width=.27\textwidth]{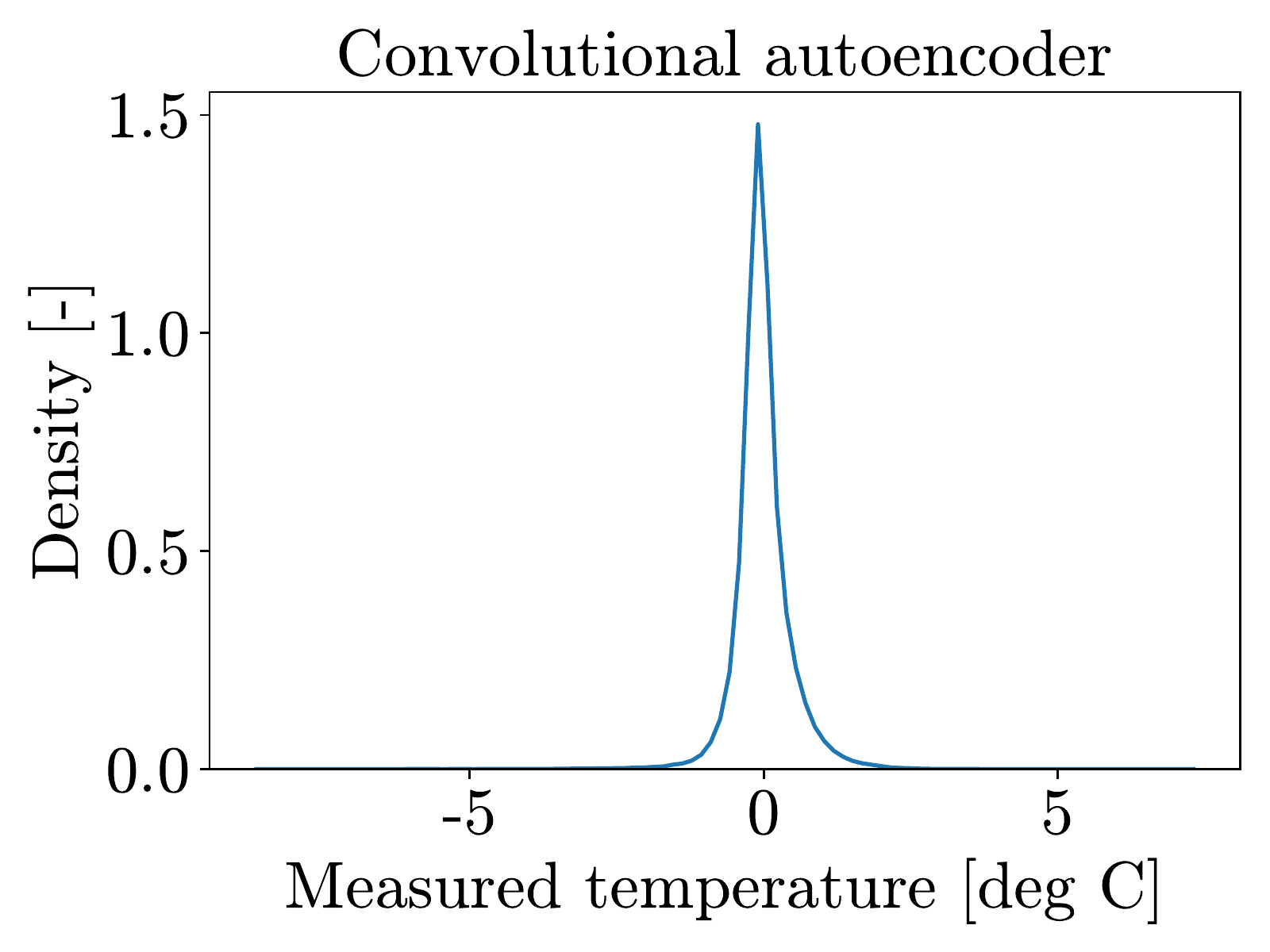}\hfill
\includegraphics[width=.27\textwidth]{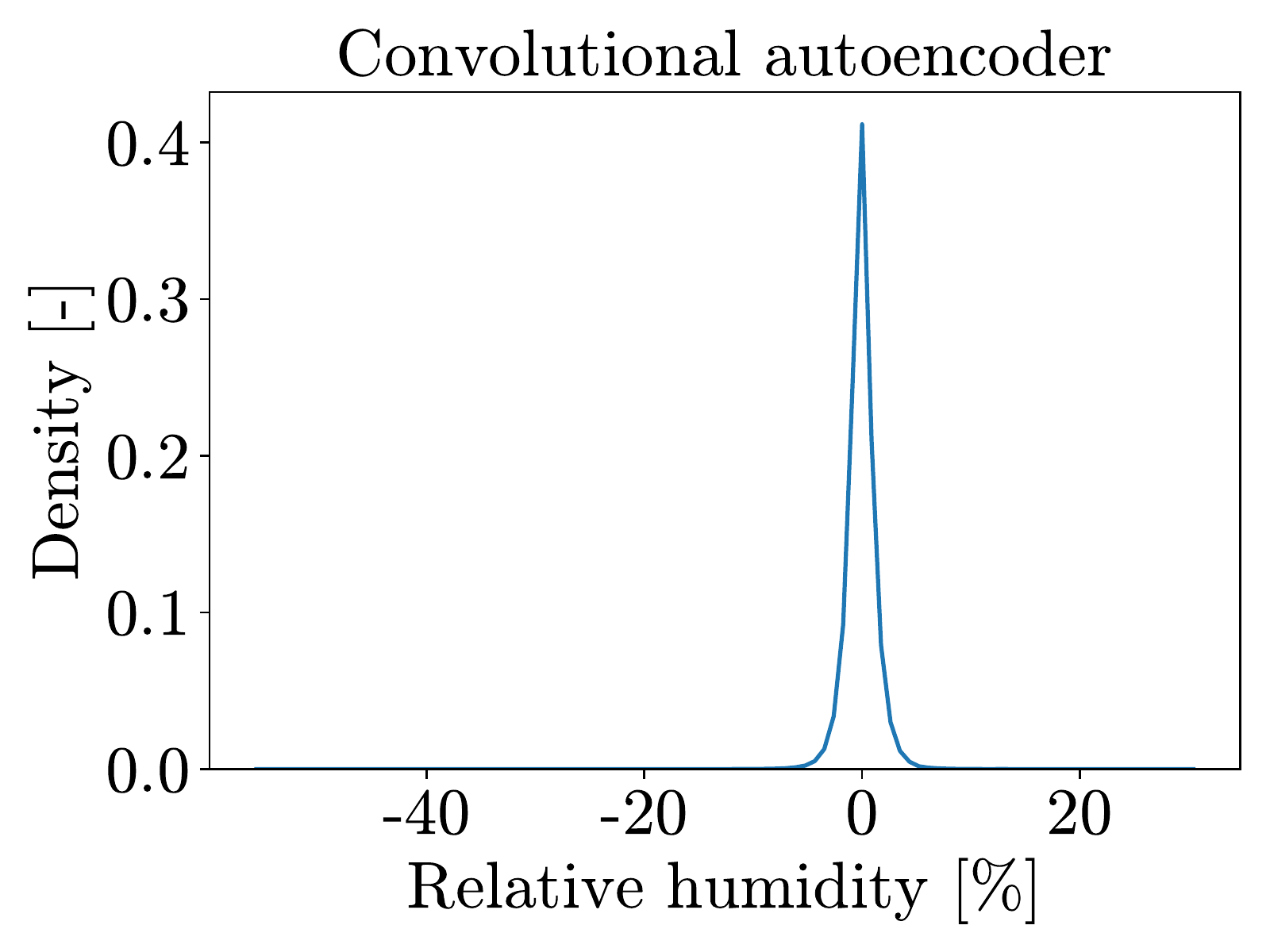}\hfill
\includegraphics[width=.27\textwidth]{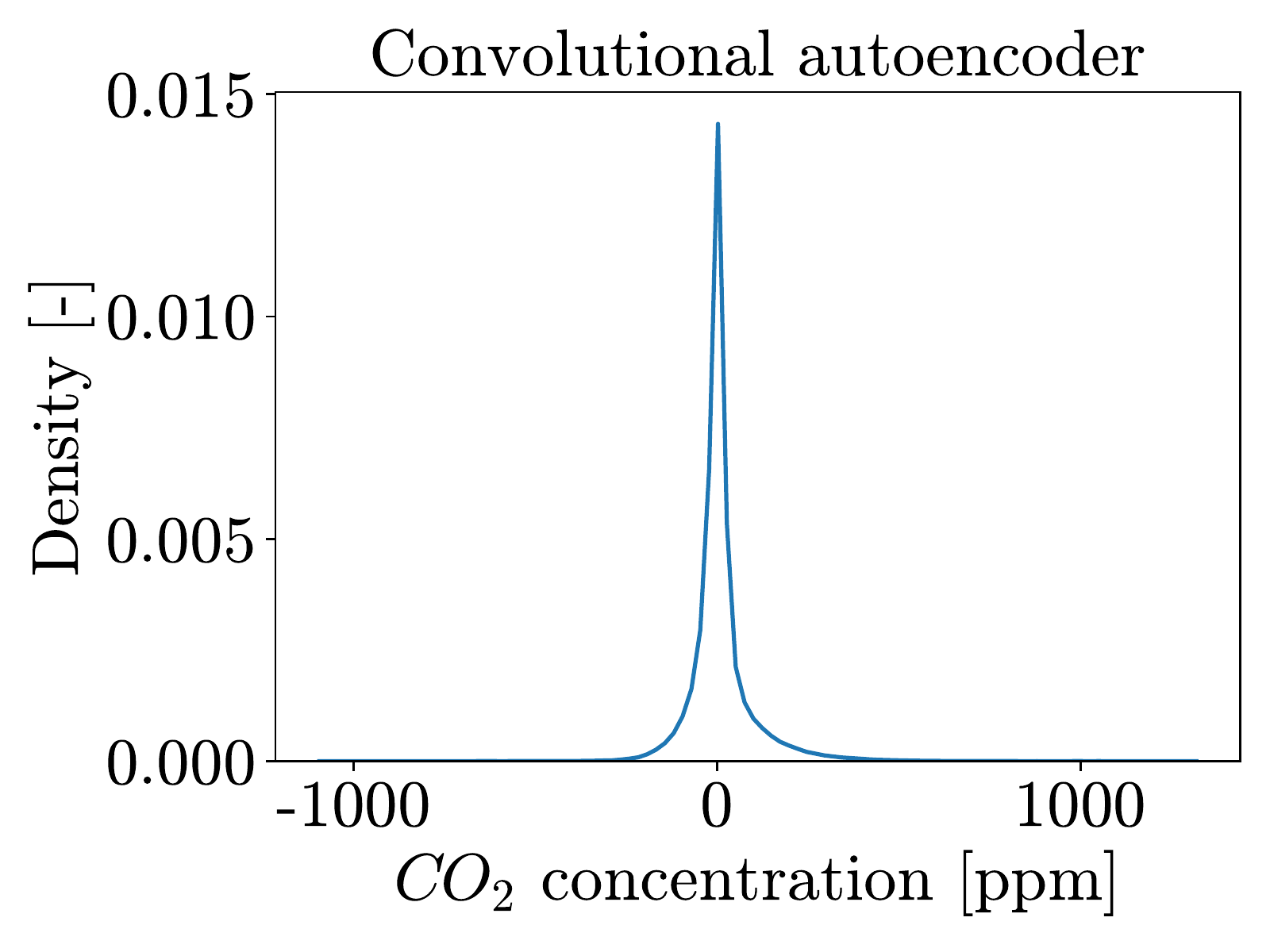}
\caption{Density distribution of the reconstruction residuals for a corruption rate of 0.5.} 
\label{fig:err_rec}

\end{figure}

Figure \ref{fig:err_rec} shows the density distribution of the reconstruction residuals produced by the best models identified by the previous table (i.e. convolutional autoencoder), for a corruption rate of 0.5. Here, the reconstruction residuals were computed as the differences between the observed and inserted corrupted values on the evaluation set. For each variable, it was possible to note a normal distribution with a mean of zero, hence confirming that the autoencoders were well developed \cite{jiayuan}. The presence of extreme values in these pictures could be traced back to the undetected outliers, as explained at the beginning of this subsection.

In order to support the previous results, some descriptive statistics of the complete and filled evaluation set are shown in Table \ref{tab:stat_rec}. Here, the effects of the missing values inserting with the best model (i.e. convolutional autoencoder) are presented with a corruption rate ranging from 0 (i.e. original data set) to 0.9. In average, for each variable, the descriptive statistics remained almost unchanged. The worst effect on the final statistics of the evaluation set was registered on the $CO_2$ concentration data, hence confirming the findings in Table \ref{tab:subdaily}.

\begin{table}[H]
\centering
\caption{Descriptive statistics for the evaluation set before (CR = 0.00) and after (CR $>$ 0.00) missing data reconstruction with the convolutional autoencoder. "Std" and "Avg" stand for standard deviation and average.}
\label{tab:stat_rec}
\begin{tabular}{cllllllllllll}
\\
\toprule
\multirow{2}{*}{CR [-]} & \multicolumn{4}{l}{$T$ [\textdegree C]} & \multicolumn{4}{l}{$RH$ [\%]} & \multicolumn{4}{l}{$CO_2$ [ppm]} \\
\cmidrule{2-13}
         & Min   & Max   & Mean  & Std  & Min  & Max   & Mean  & Std   & Min    & Max     & Mean   & Std    \\
\toprule
0.00 & 12.90 & 31.70 & 22.68 & 1.12 & 0.90 & 81.30 & 39.08 & 10.98 & 192.50 & 2000.00 & 510.78 & 122.81 \\
\toprule
0.10     & 12.90 & 31.70 & 22.67 & 1.11 & 0.90 & 81.30 & 39.06 & 10.96 & 195.00 & 2179.84 & 510.90 & 121.83 \\
0.20     & 12.90 & 31.70 & 22.67 & 1.11 & 0.90 & 81.30 & 39.06 & 10.94 & 195.00 & 2044.92 & 509.18 & 120.04 \\
0.30     & 12.90 & 30.10 & 22.67 & 1.11 & 0.90 & 78.45 & 39.02 & 11.02 & 195.00 & 2062.05 & 507.51 & 116.02 \\
0.40     & 12.90 & 30.10 & 22.70 & 1.09 & 0.90 & 74.80 & 39.04 & 11.04 & 192.50 & 2000.00 & 509.30 & 116.51 \\
0.50     & 13.10 & 30.10 & 22.68 & 1.05 & 0.90 & 81.30 & 39.12 & 10.96 & 195.00 & 2140.17 & 505.35 & 107.22 \\
0.60     & 13.10 & 30.10 & 22.68 & 1.05 & 0.90 & 72.20 & 39.11 & 10.87 & 192.50 & 2042.48 & 498.78 & 99.12  \\
0.70     & 13.10 & 28.50 & 22.70 & 1.04 & 0.90 & 72.20 & 39.04 & 10.74 & 195.00 & 2000.00 & 512.10 & 105.56 \\
0.80     & 13.20 & 28.50 & 22.69 & 1.00 & 0.90 & 72.20 & 39.23 & 10.63 & 195.00 & 2000.00 & 504.87 & 96.45  \\
0.90     & 13.20 & 28.40 & 22.65 & 0.99 & 0.90 & 72.20 & 39.13 & 10.73 & 195.00 & 1376.00 & 508.78 & 91.04  \\
\cmidrule{2-13}
Avg  & 13.03 & 29.91 & 22.68 & 1.06 & 0.90 & 76.22 & 39.09 & 10.88 & 194.44 & 1982.83 & 507.42 & 108.20 \\
\toprule

\end{tabular}
\end{table}

\begin{figure}[H]

\centering

\includegraphics[width=.27\textwidth]{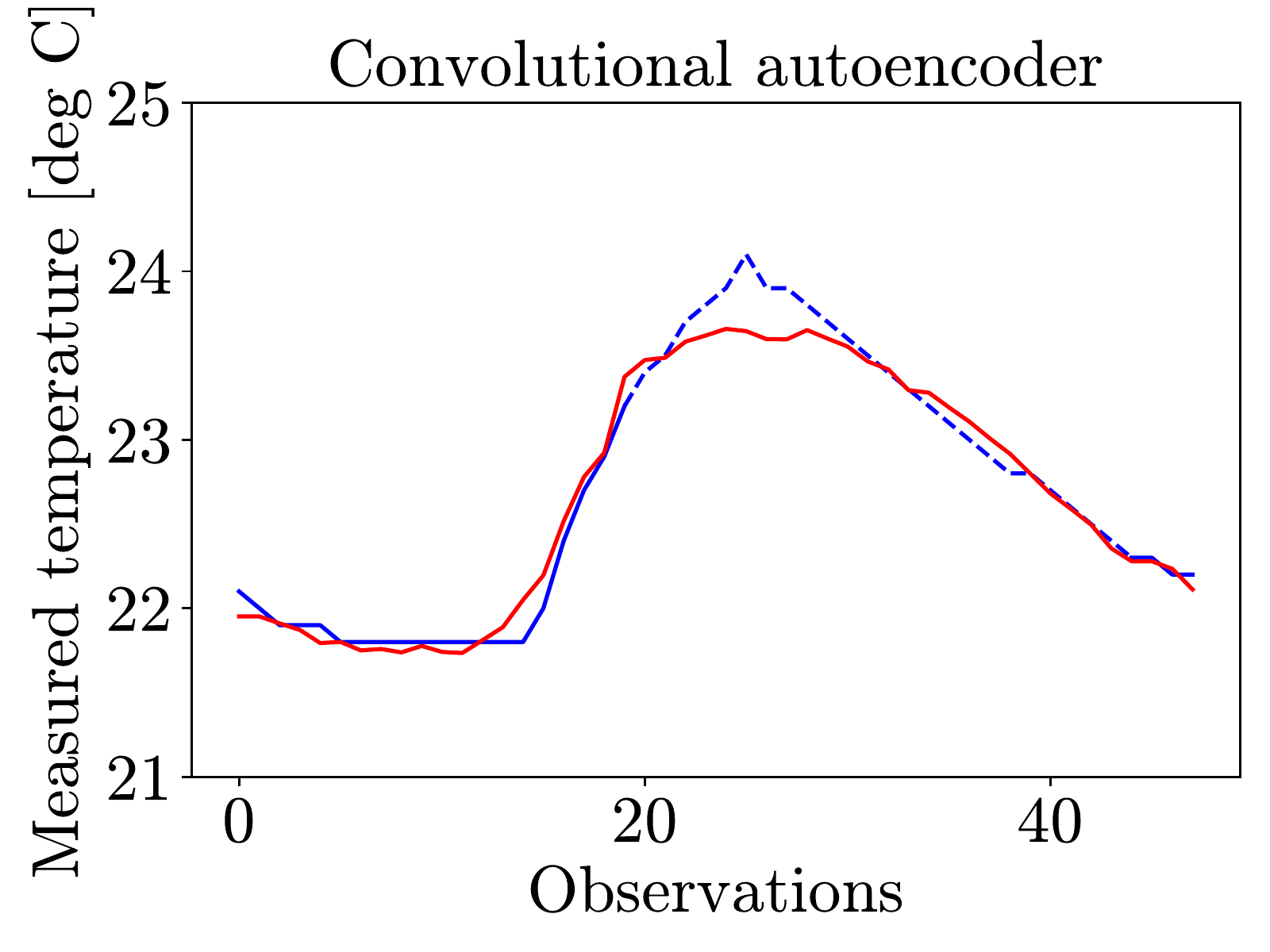}\hfill
\includegraphics[width=.27\textwidth]{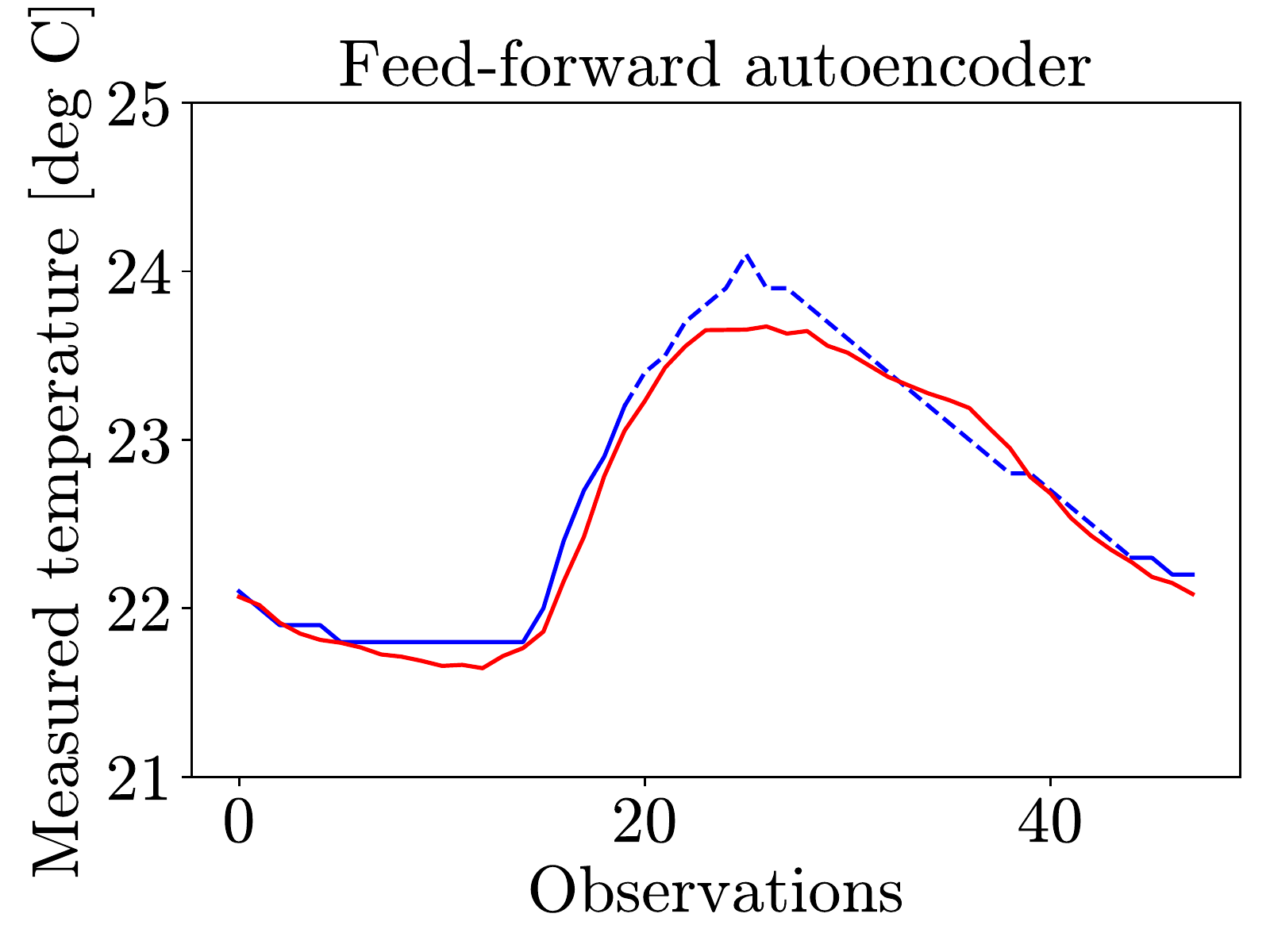}\hfill
\includegraphics[width=.27\textwidth]{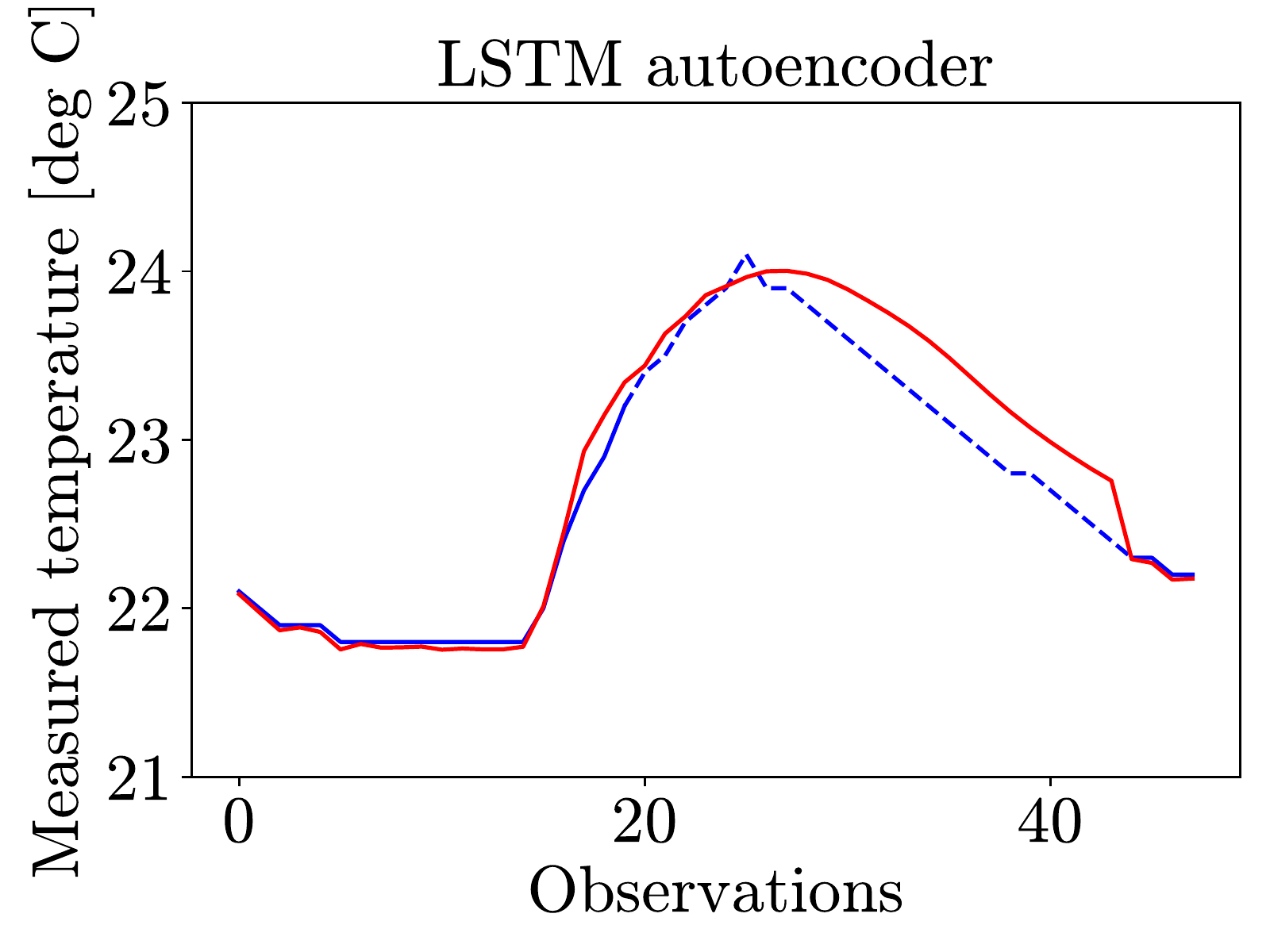}
\includegraphics[width=.27\textwidth]{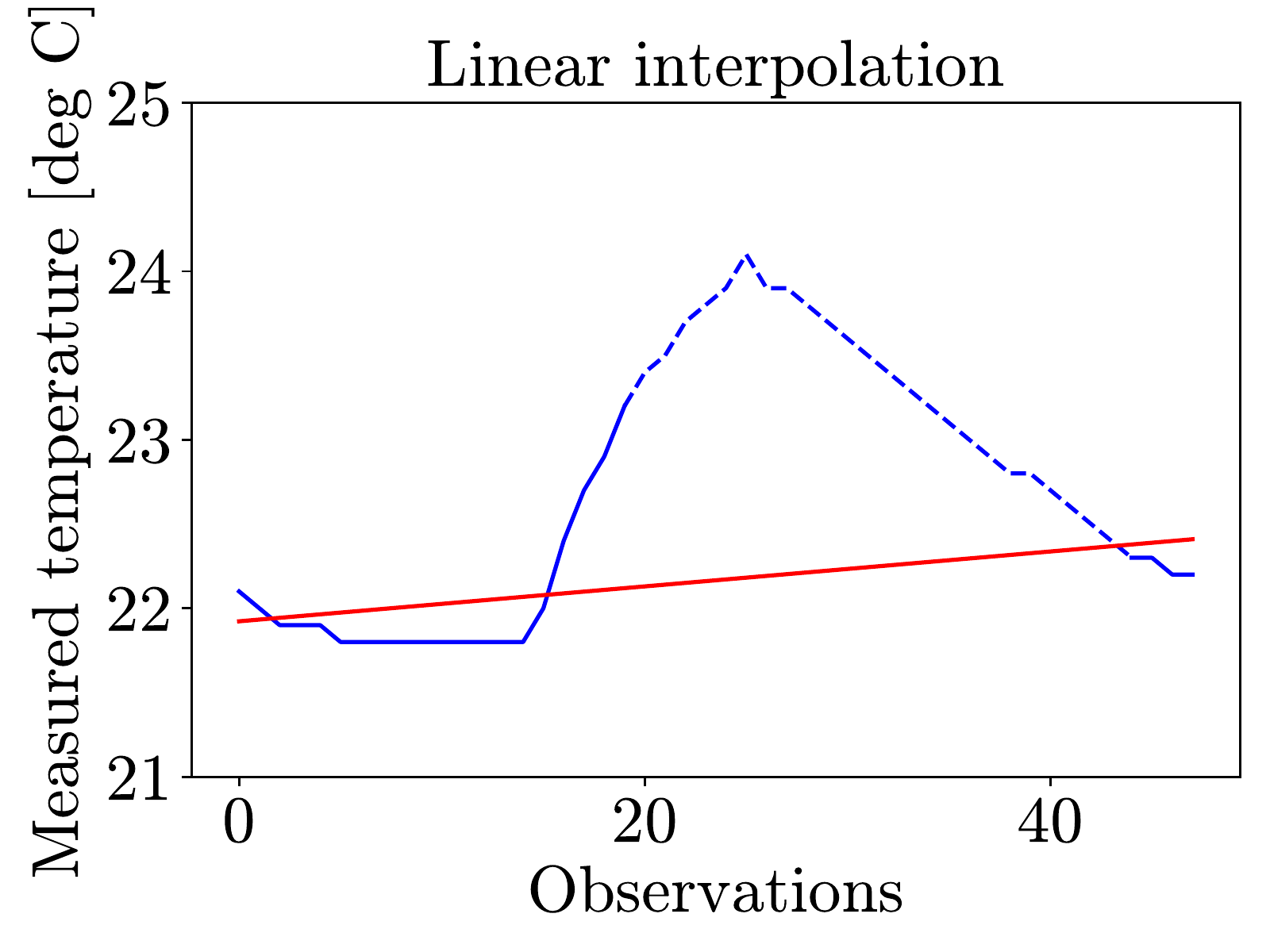}\hfill
\includegraphics[width=.27\textwidth]{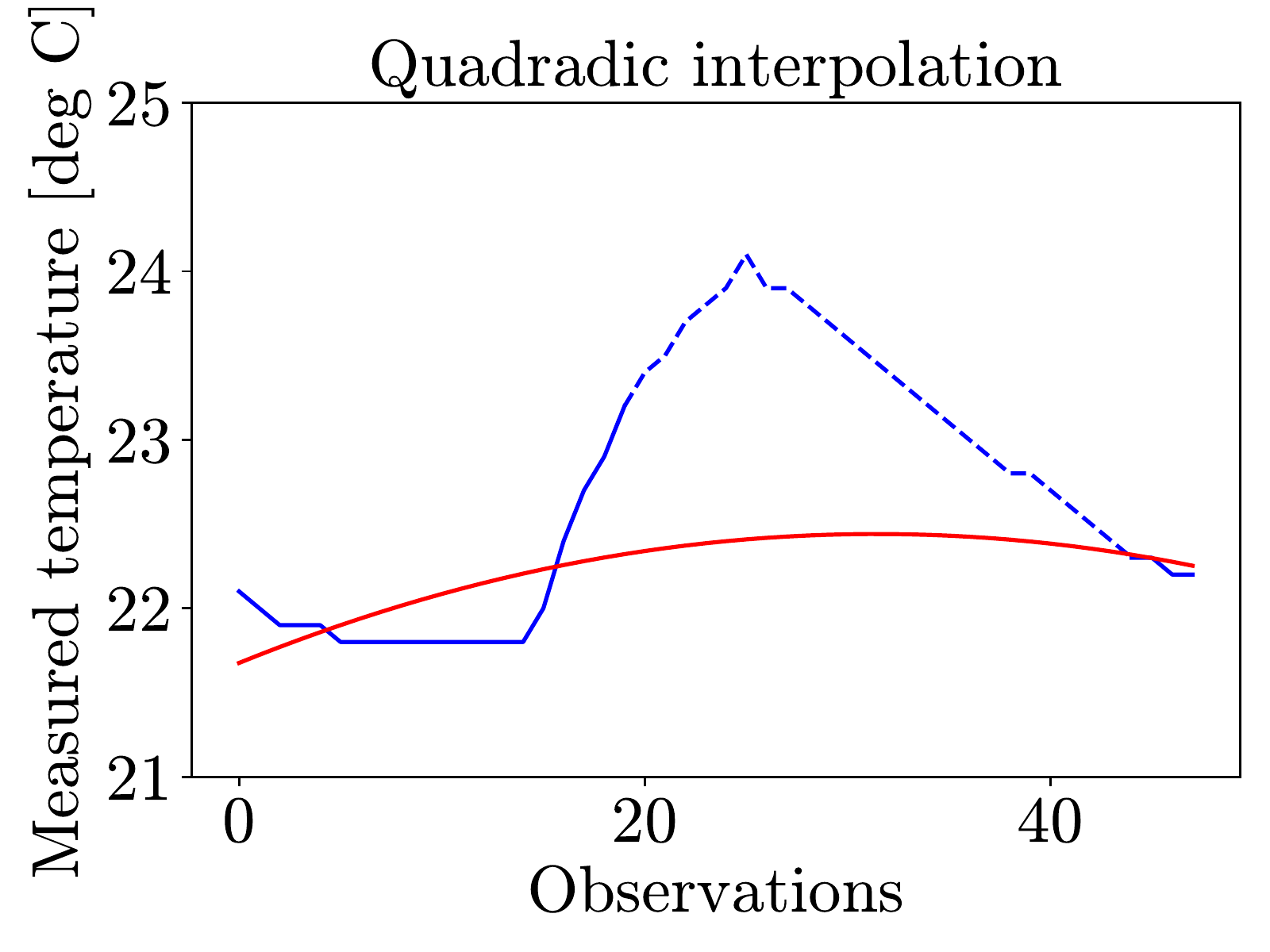}\hfill
\includegraphics[width=.27\textwidth]{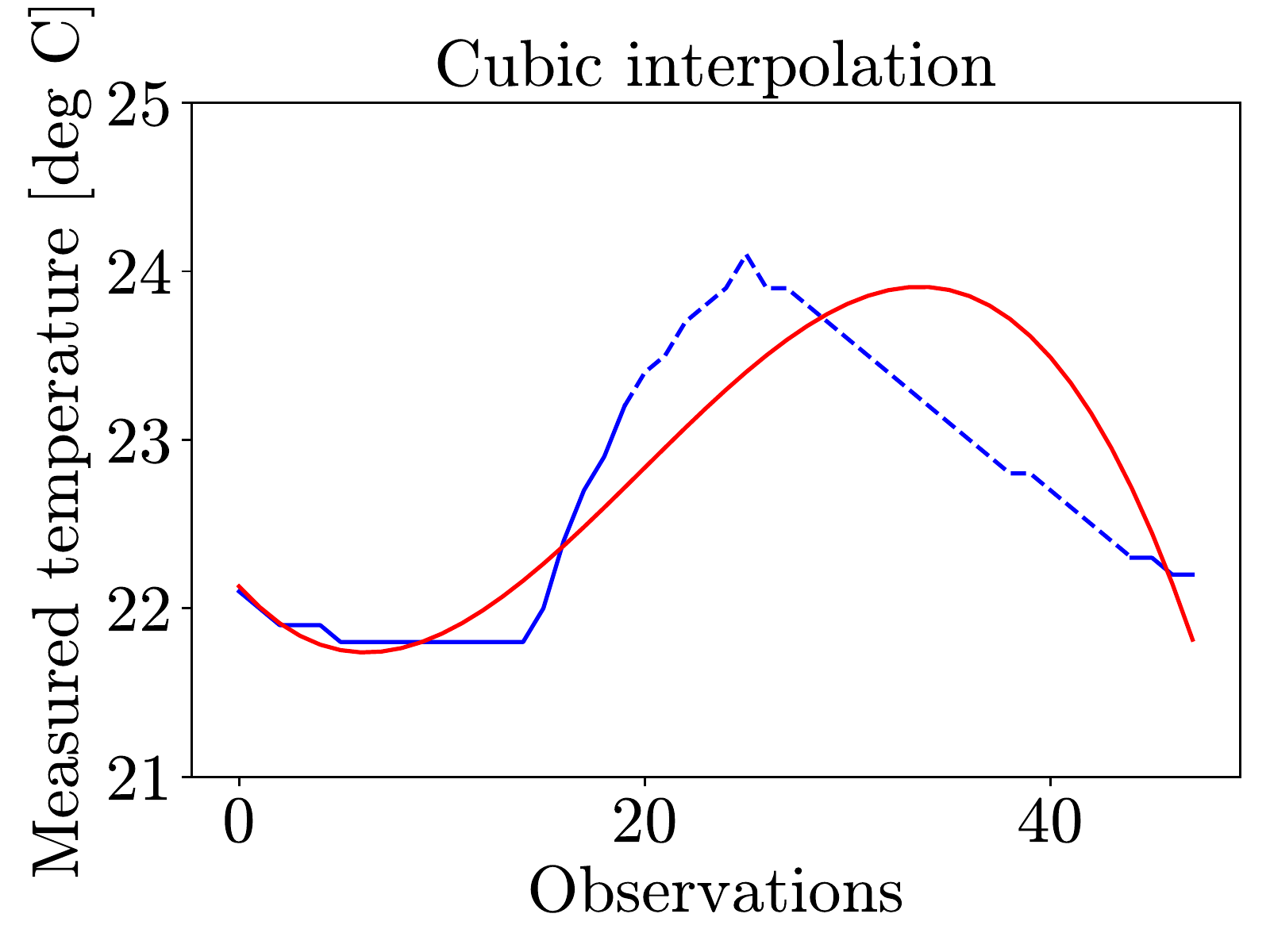}
\includegraphics[width=.27\textwidth]{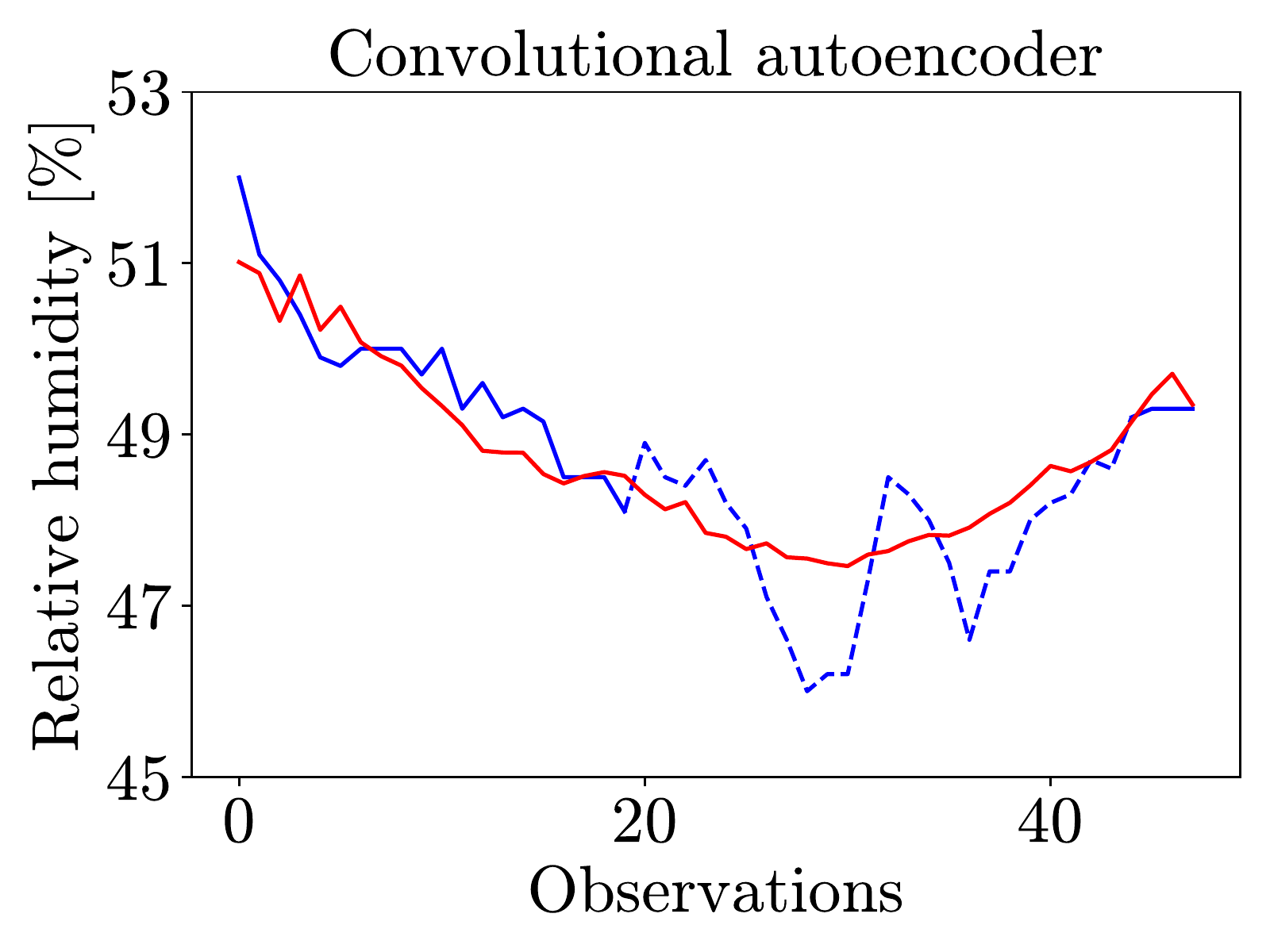}\hfill
\includegraphics[width=.27\textwidth]{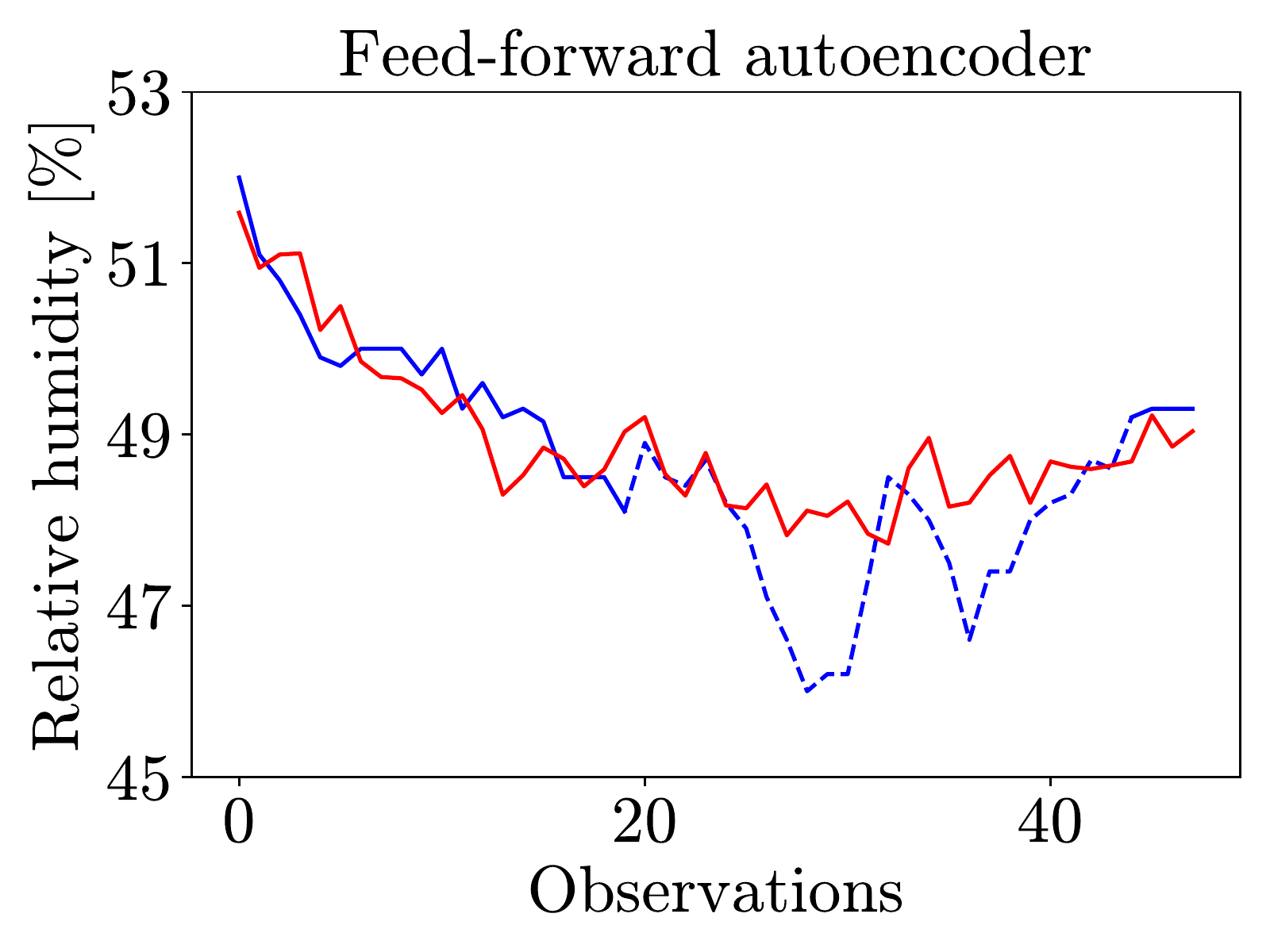}\hfill
\includegraphics[width=.27\textwidth]{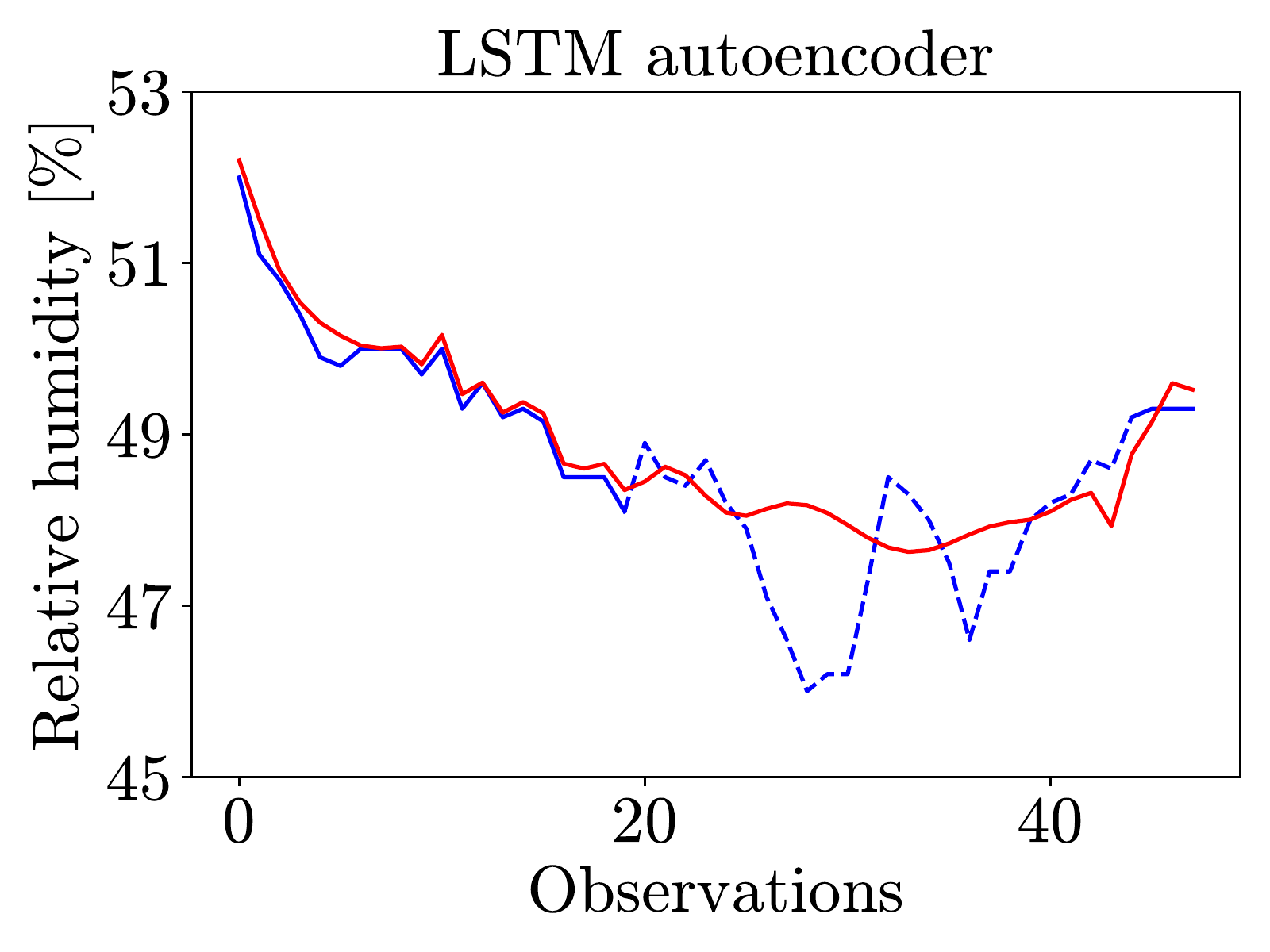}
\includegraphics[width=.27\textwidth]{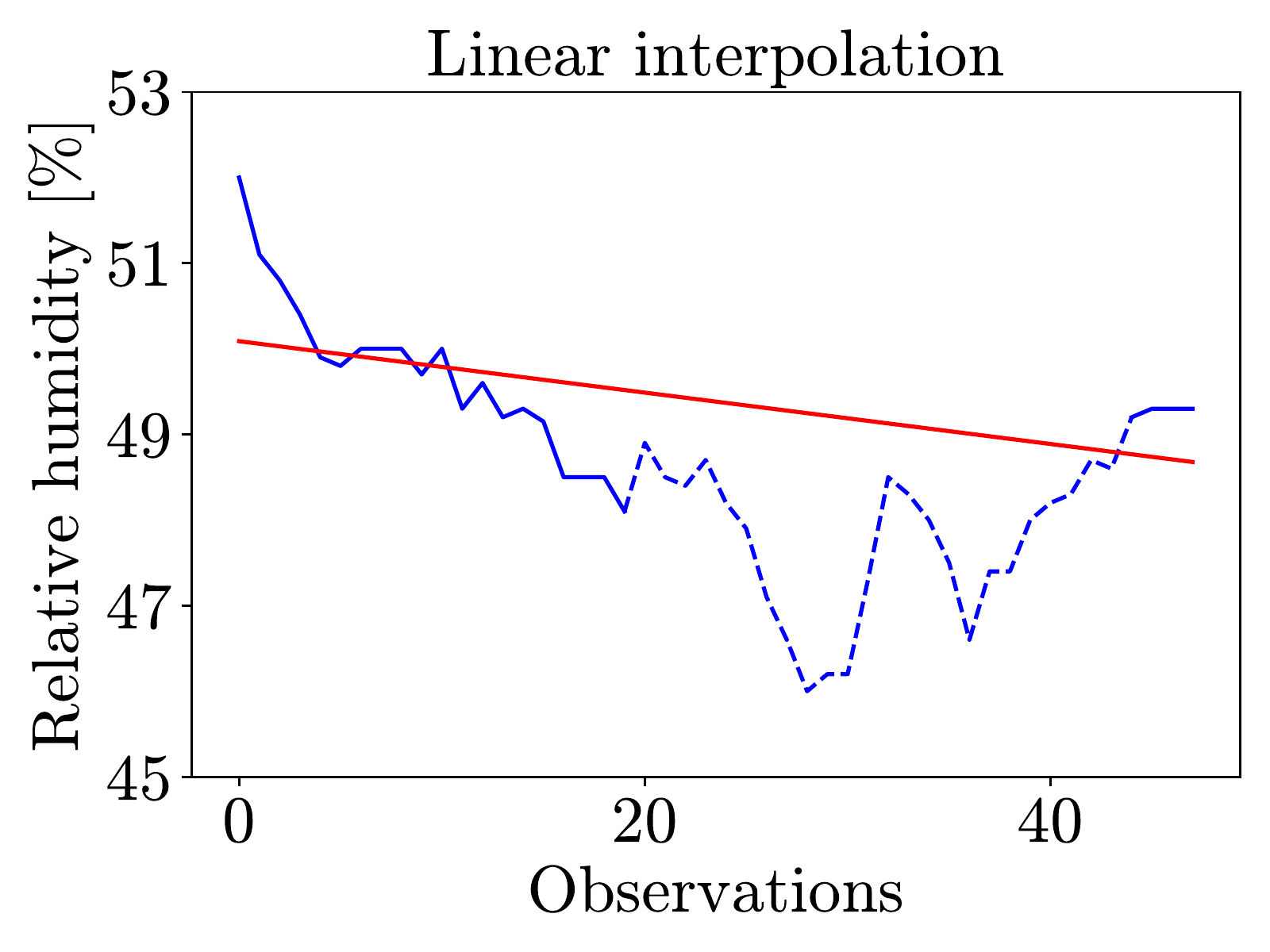}\hfill
\includegraphics[width=.27\textwidth]{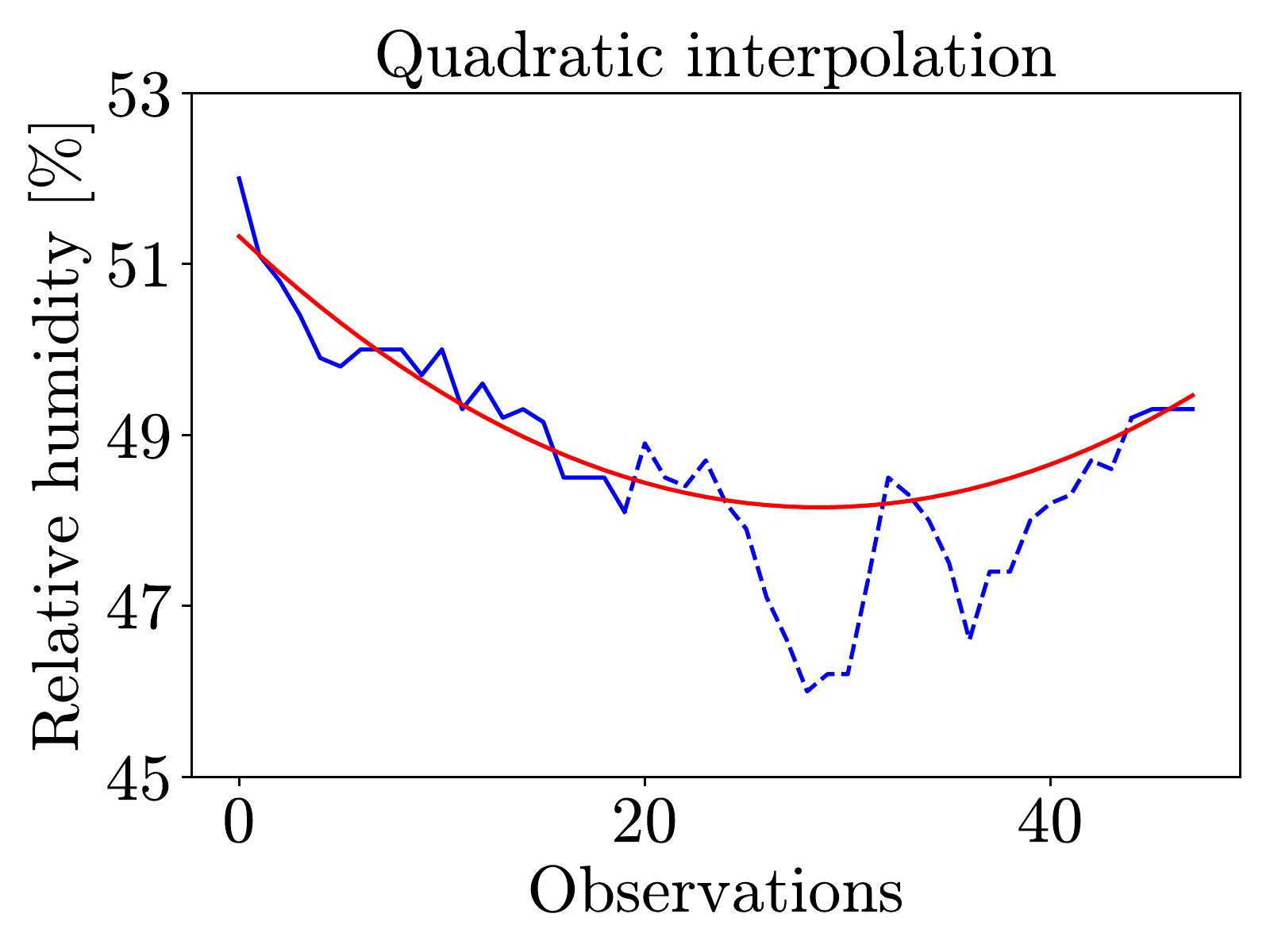}\hfill
\includegraphics[width=.27\textwidth]{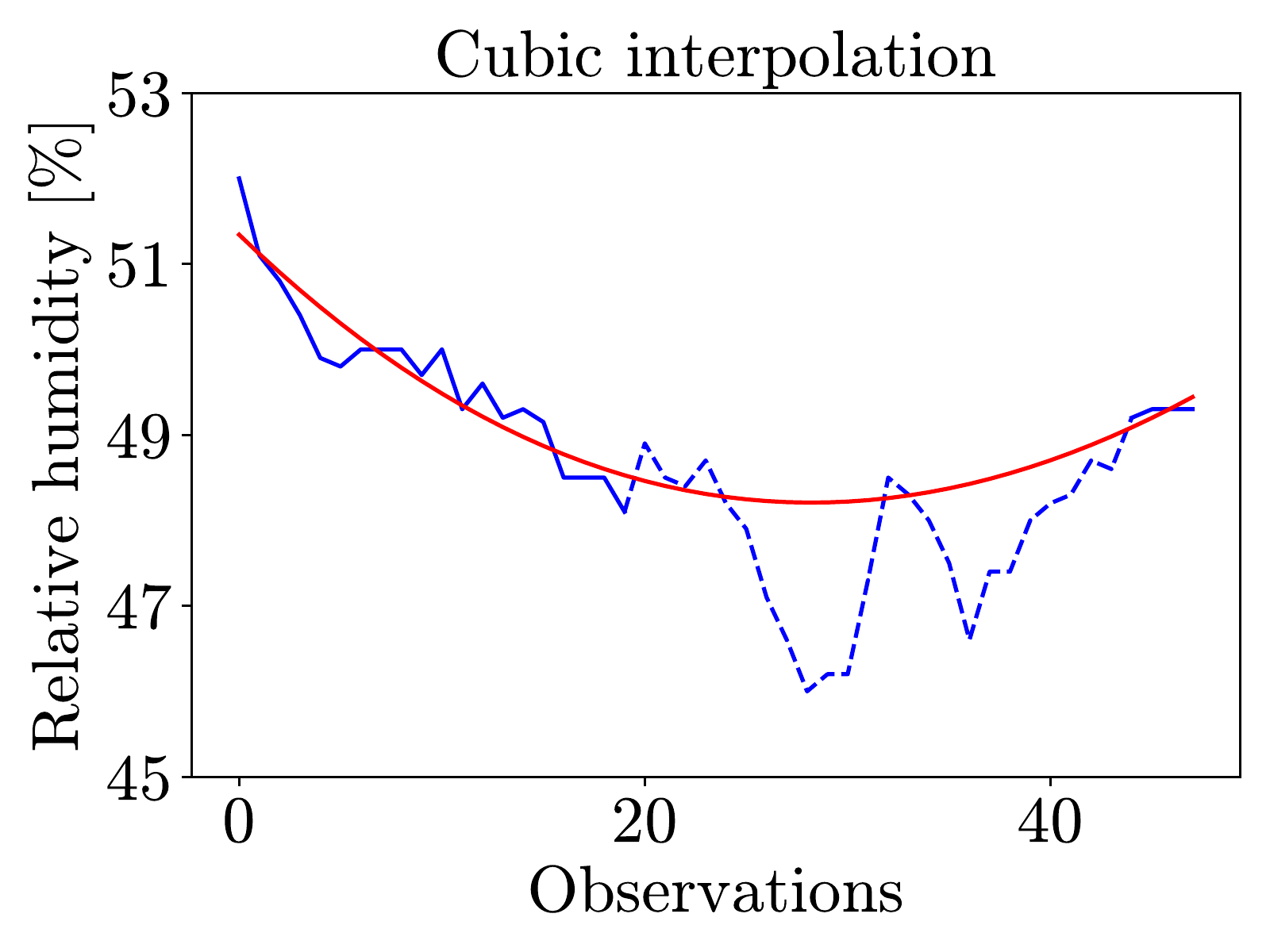}
\includegraphics[width=.27\textwidth]{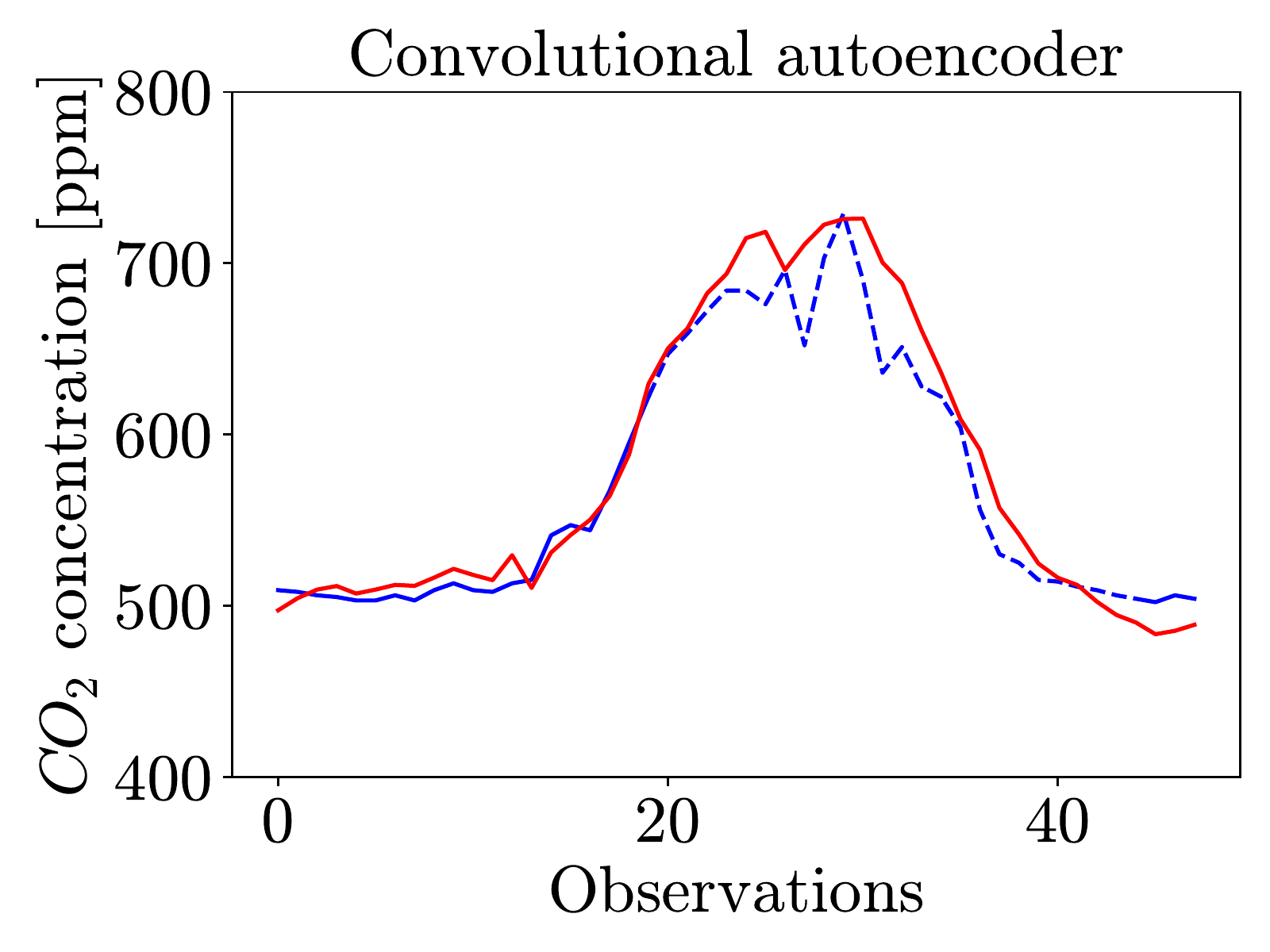}\hfill
\includegraphics[width=.27\textwidth]{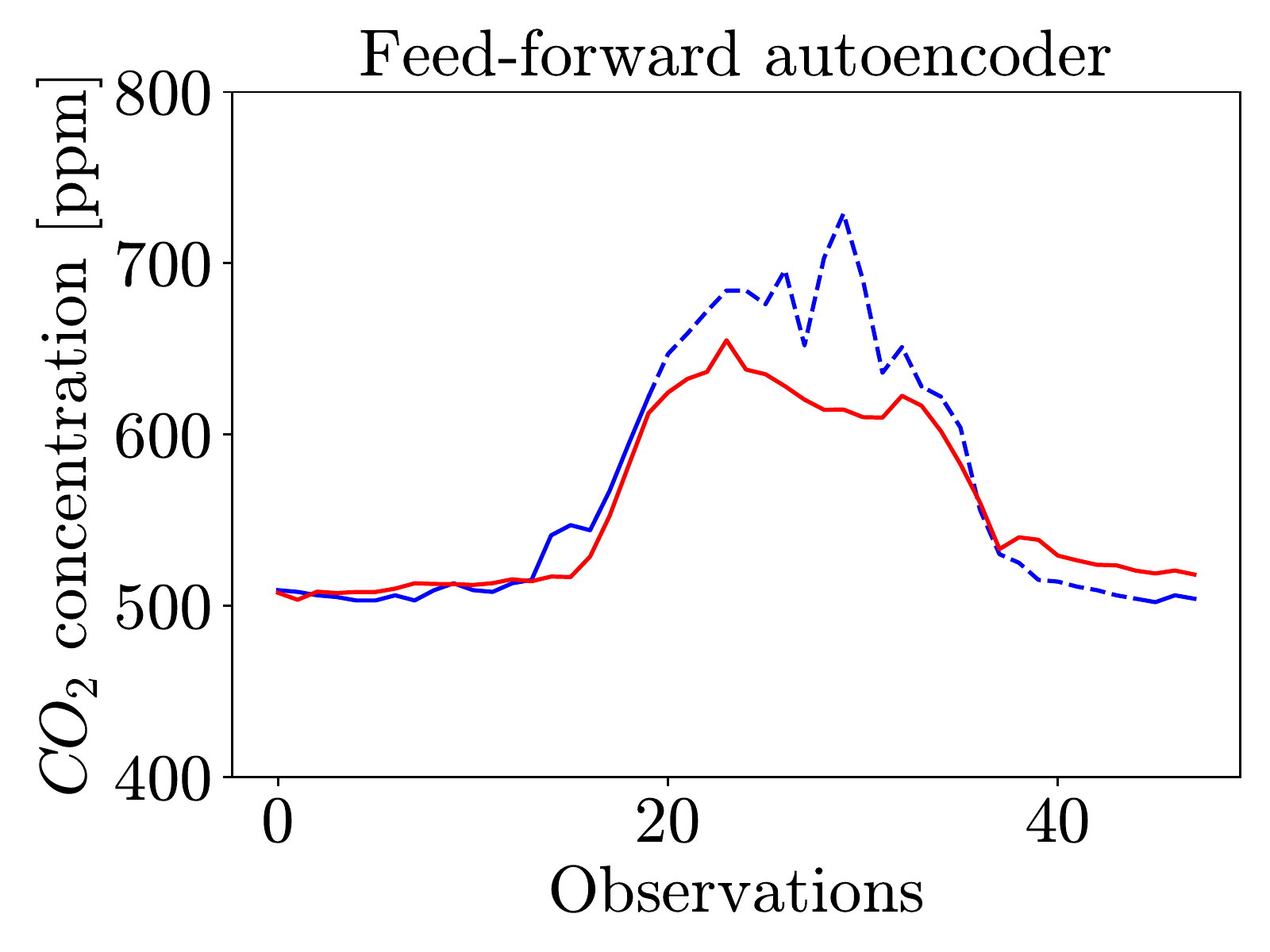}\hfill
\includegraphics[width=.27\textwidth]{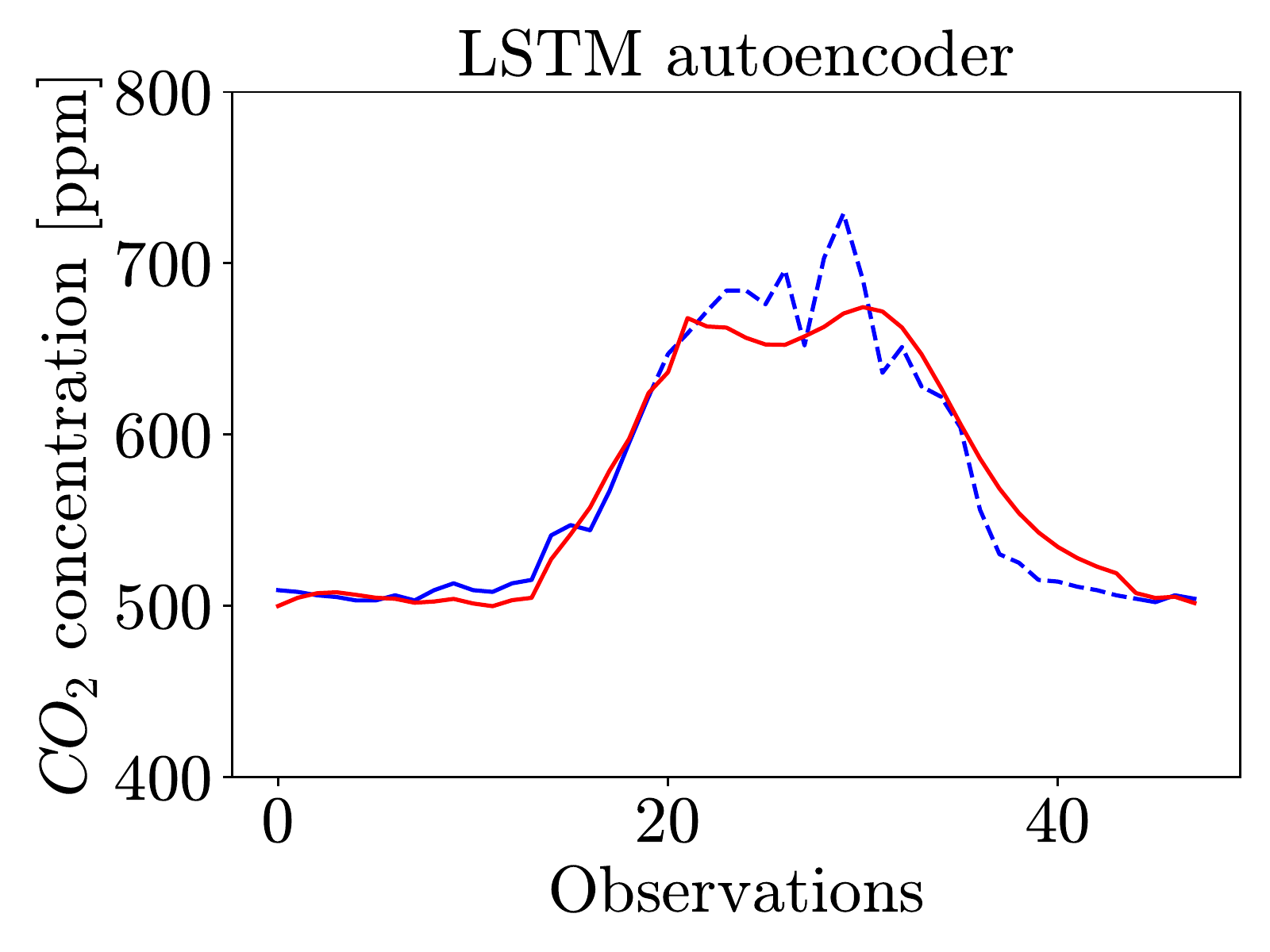}
\includegraphics[width=.27\textwidth]{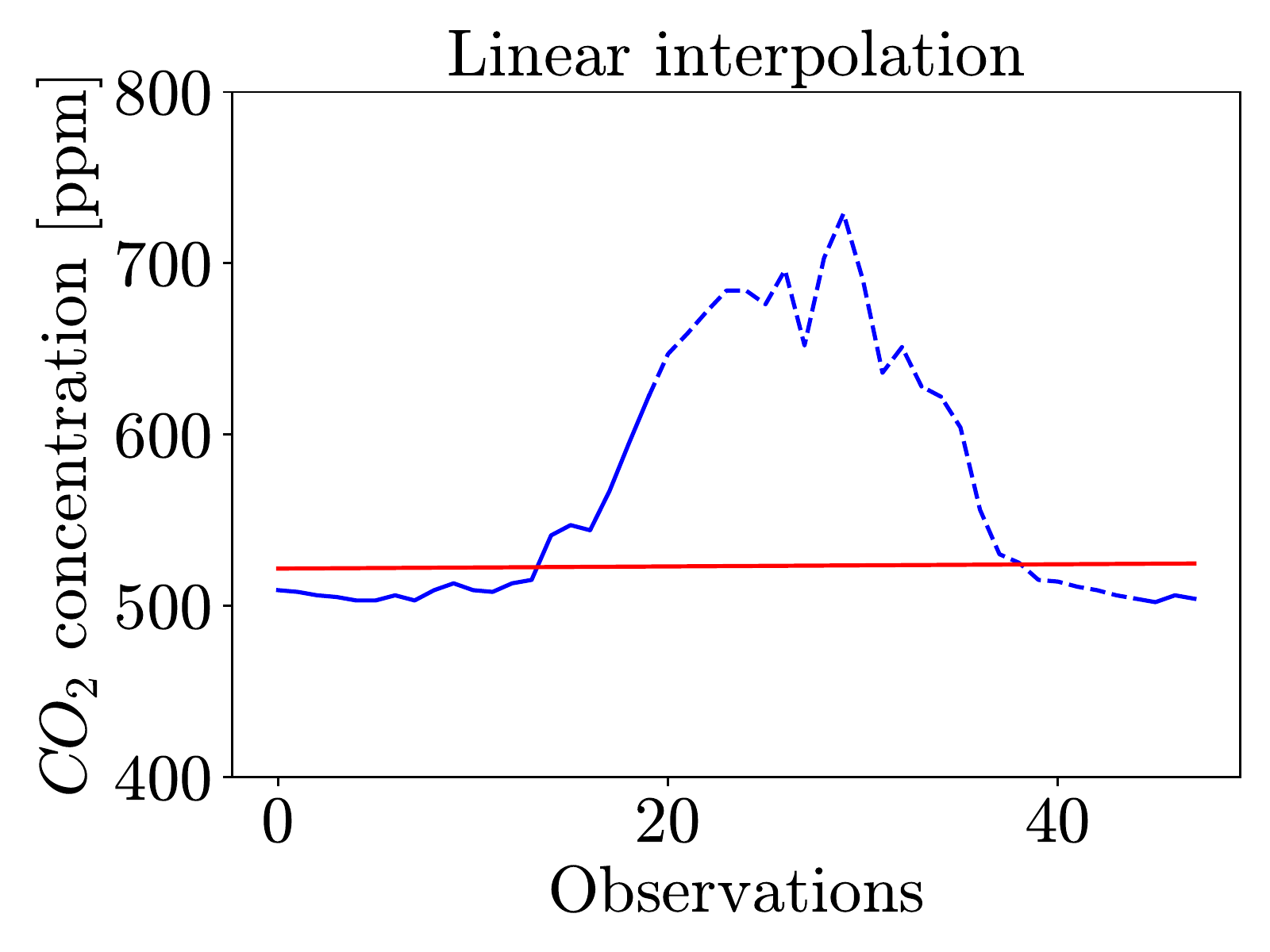}\hfill
\includegraphics[width=.27\textwidth]{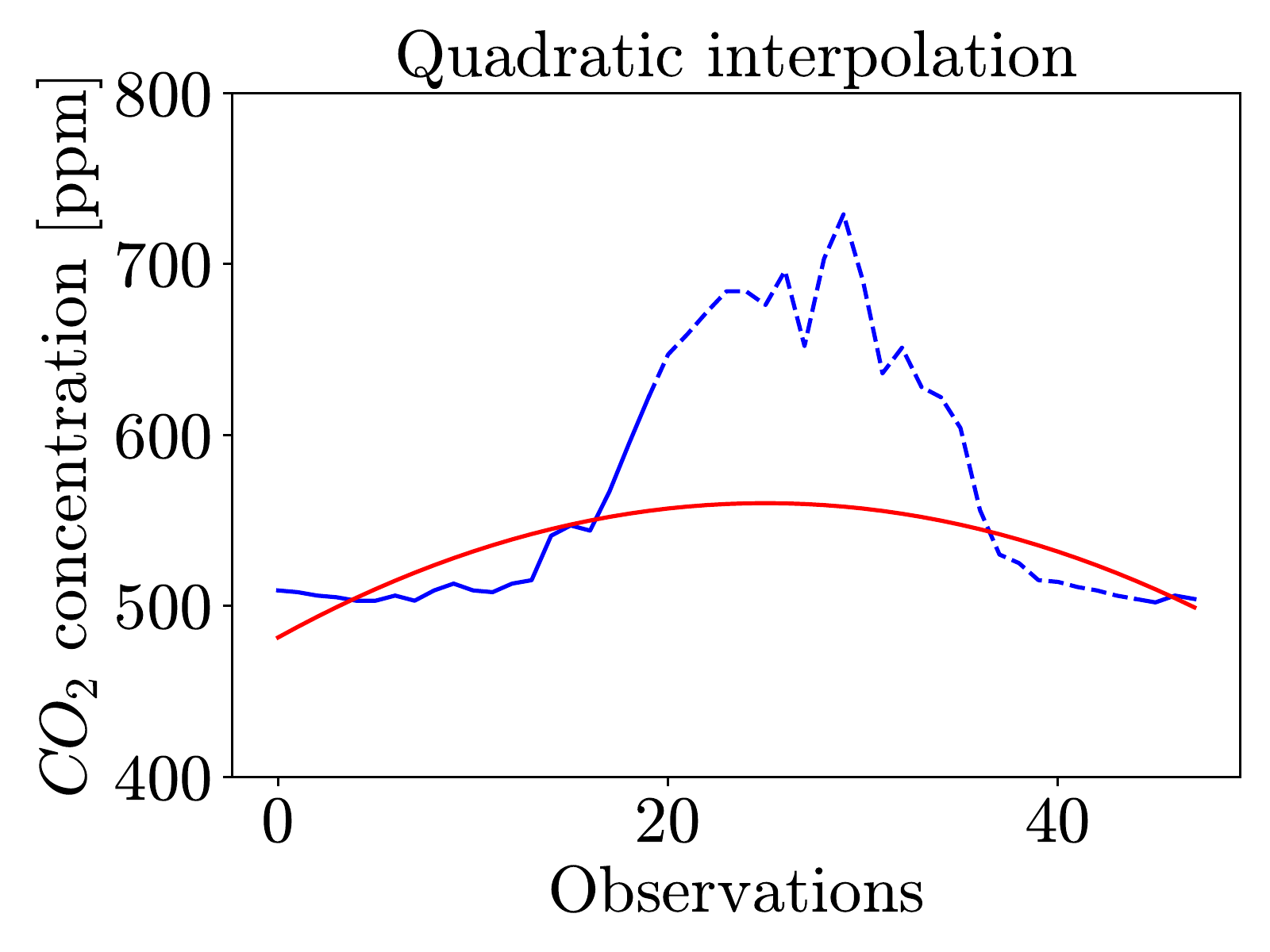}\hfill
\includegraphics[width=.27\textwidth]{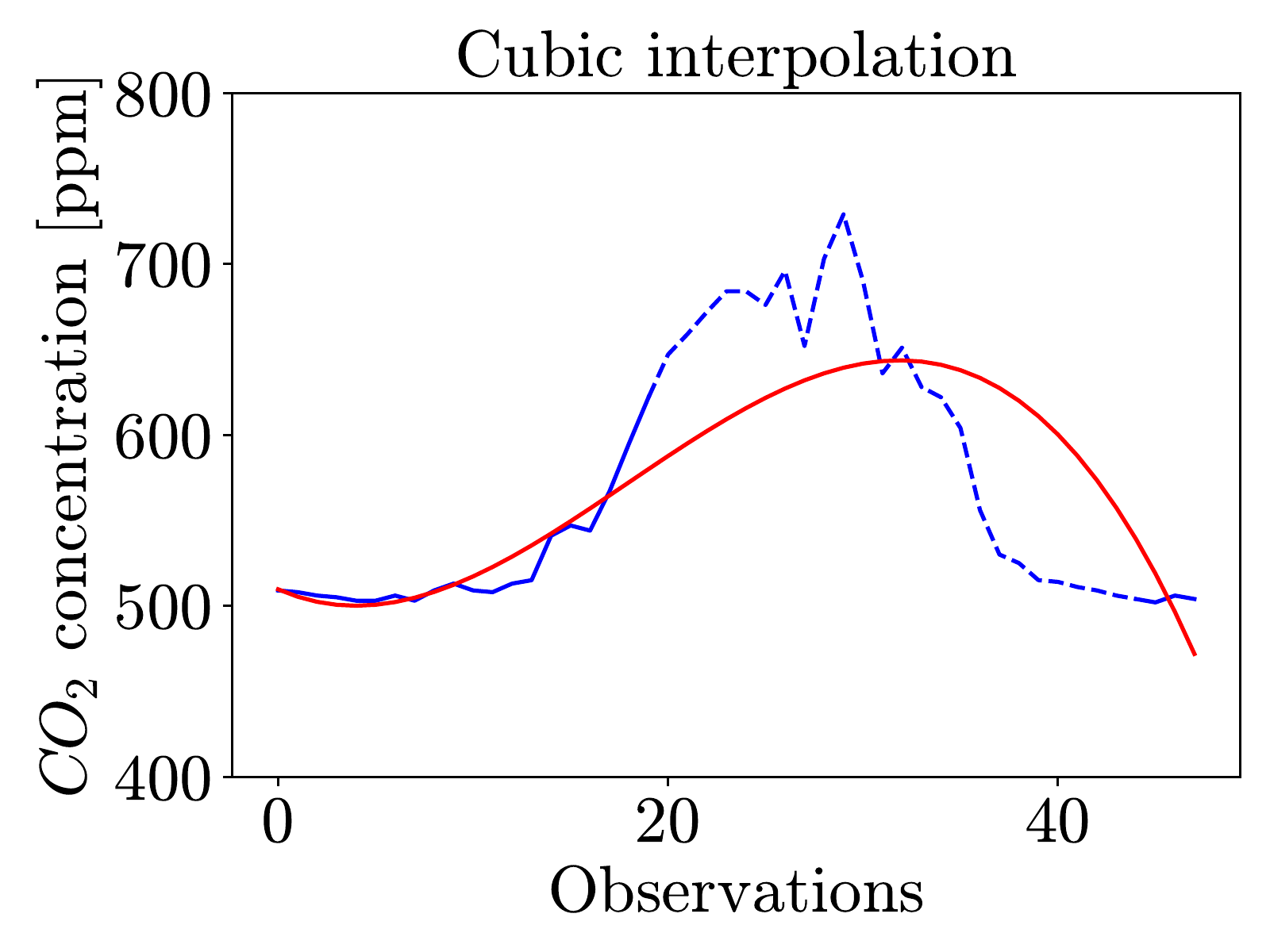}

\caption{One day-long indoor environment data reconstruction. Blue colored line represents the real data. Hashed blue colored line represents the missing data. Red colored line represents the reconstruction of the whole day with the adopted model. Observations were sampled to 30 minutes steps.}
\label{fig:data_reconstruction}

\end{figure}

Figure \ref{fig:data_reconstruction} shows exemplary indoor environment data reconstruction over one random day from the evaluation sets. All presented data were corrupted with a masking noise of 0.5. The presented data confirmed once again the results presented in Table \ref{tab:subdaily}, namely, that the proposed denoising autoencoder architectures were suitable for filling the missing indoor environment data sequences.

\subsection{Data forecasting performance evaluation}
\label{sec: forecasting}

As presented in earlier sections, the proposed autoencoder neural networks were implemented to reconstruct short-term indoor environment data gaps, since building data sets often contain missing values that could hinder further energy analysis. Nonetheless, the same models could be also used for short-term indoor environment data forecasting. \par 
Table \ref{tab:forecasting} summarizes the performance of the implemented autoencoders for different predictive horizons. In summary, all the models performed similarly well. However, there was a clear improvement, with respect to the data reconstruction case, in the LSTM configuration. In particular, the RMSE of the LSTM model was 14~\% lower for indoor air temperature and 20~\% lower for $CO_2$ concentration. In terms of NRMSE, even in this case, the proposed autoencoder neural networks could forecast relative humidity data better than other variables. In particular, the NRMSE of the LSTM configuration in case of $RH$ data was 59~\% lower than $T$ and 80~\% lower than $CO_2$ data (Table \ref{tab:forecasting}).

\begin{table}[H]
\centering
\caption{Performance of denoising autoencoder neural networks for forecasting indoor environment data. "CONV", "FEED" and "LSTM" stand for convolutional, feed-forward and LSTM denoising autoencoder. "Avg" stands for average. PH is the predictive horizon.}
\label{tab:forecasting}
\begin{tabular}{cllllllllll}
\\
\toprule
\multicolumn{1}{l}{} &
  \multicolumn{1}{r}{\multirow{2}{*}{PH [h]}} &
  \multicolumn{3}{l}{$T$ [\textdegree C]} &
  \multicolumn{3}{l}{$RH$ [\%]} &
  \multicolumn{3}{l}{$CO_2$ [ppm]} \\
    \cmidrule{3-11}
\multicolumn{1}{l}{} &
  \multicolumn{1}{c}{} &
  CONV &
  FEED &
  LSTM &
  CONV &
  FEED &
  LSTM &
  CONV &
  FEED &
  LSTM \\
  \toprule
\multirow{10}{*}{RMSE}& \multicolumn{1}{r}{2.50}& \multicolumn{1}{r}{0.19}& \multicolumn{1}{r}{0.18}& \multicolumn{1}{r}{0.17}& \multicolumn{1}{r}{0.89}& \multicolumn{1}{r}{0.97}& \multicolumn{1}{r}{0.89}& \multicolumn{1}{r}{25.62}& \multicolumn{1}{r}{23.86}& \multicolumn{1}{r}{25.34}\\
                    \multicolumn{1}{r}{}& \multicolumn{1}{r}{5.00}& \multicolumn{1}{r}{0.31}& \multicolumn{1}{r}{0.30}& \multicolumn{1}{r}{0.29}& \multicolumn{1}{r}{1.47}& \multicolumn{1}{r}{1.44}& \multicolumn{1}{r}{1.37}& \multicolumn{1}{r}{43.00}& \multicolumn{1}{r}{42.38}& \multicolumn{1}{r}{39.33}\\
                    \multicolumn{1}{r}{}& \multicolumn{1}{r}{7.00}& \multicolumn{1}{r}{0.42}& \multicolumn{1}{r}{0.41}& \multicolumn{1}{r}{0.40}& \multicolumn{1}{r}{1.72}& \multicolumn{1}{r}{1.83}& \multicolumn{1}{r}{1.76}& \multicolumn{1}{r}{55.62}& \multicolumn{1}{r}{55.29}& \multicolumn{1}{r}{56.11}\\
                    \multicolumn{1}{r}{}& \multicolumn{1}{r}{9.50}& \multicolumn{1}{r}{0.50}& \multicolumn{1}{r}{0.48}& \multicolumn{1}{r}{0.46}& \multicolumn{1}{r}{2.15}& \multicolumn{1}{r}{2.21}& \multicolumn{1}{r}{2.18}& \multicolumn{1}{r}{72.37}& \multicolumn{1}{r}{72.84}& \multicolumn{1}{r}{74.21}\\
                    \multicolumn{1}{r}{}& \multicolumn{1}{r}{12.00}& \multicolumn{1}{r}{0.55}& \multicolumn{1}{r}{0.52}& \multicolumn{1}{r}{0.52}& \multicolumn{1}{r}{2.47}& \multicolumn{1}{r}{2.45}& \multicolumn{1}{r}{2.41}& \multicolumn{1}{r}{81.27}& \multicolumn{1}{r}{82.55}& \multicolumn{1}{r}{82.11}\\
                    \multicolumn{1}{r}{}& \multicolumn{1}{r}{14.50}& \multicolumn{1}{r}{0.62}& \multicolumn{1}{r}{0.60}& \multicolumn{1}{r}{0.59}& \multicolumn{1}{r}{2.88}& \multicolumn{1}{r}{2.85}& \multicolumn{1}{r}{2.67}& \multicolumn{1}{r}{102.71}& \multicolumn{1}{r}{102.40}& \multicolumn{1}{r}{104.51}\\
                    \multicolumn{1}{r}{}& \multicolumn{1}{r}{17.00}& \multicolumn{1}{r}{0.73}& \multicolumn{1}{r}{0.66}& \multicolumn{1}{r}{0.66}& \multicolumn{1}{r}{2.84}& \multicolumn{1}{r}{2.88}& \multicolumn{1}{r}{2.81}& \multicolumn{1}{r}{108.13}& \multicolumn{1}{r}{107.22}& \multicolumn{1}{r}{107.79}\\
                    \multicolumn{1}{r}{}& \multicolumn{1}{r}{19.00}& \multicolumn{1}{r}{0.74}& \multicolumn{1}{r}{0.64}& \multicolumn{1}{r}{0.63}& \multicolumn{1}{r}{2.93}& \multicolumn{1}{r}{2.95}& \multicolumn{1}{r}{2.84}& \multicolumn{1}{r}{102.40}& \multicolumn{1}{r}{101.58}& \multicolumn{1}{r}{102.02}\\
                    \multicolumn{1}{r}{}& \multicolumn{1}{r}{21.50}& \multicolumn{1}{r}{0.75}& \multicolumn{1}{r}{0.60}& \multicolumn{1}{r}{0.60}& \multicolumn{1}{r}{3.01}& \multicolumn{1}{r}{3.01}& \multicolumn{1}{r}{3.18}& \multicolumn{1}{r}{96.89}& \multicolumn{1}{r}{96.14}& \multicolumn{1}{r}{96.88}\\
                                  \cmidrule{3-11}
                                & \multicolumn{1}{r}{Avg} & \multicolumn{1}{r}{0.53}& \multicolumn{1}{r}{0.49}& \multicolumn{1}{r}{0.48}& \multicolumn{1}{r}{2.26}& \multicolumn{1}{r}{2.29}& \multicolumn{1}{r}{2.23}& \multicolumn{1}{r}{76.45}& \multicolumn{1}{r}{76.03}& \multicolumn{1}{r}{76.48}\\
                                \toprule
\multirow{10}{*}{NRMSE}& \multicolumn{1}{r}{2.50}& \multicolumn{1}{r}{0.13}& \multicolumn{1}{r}{0.12}& \multicolumn{1}{r}{0.11}& \multicolumn{1}{r}{0.05}& \multicolumn{1}{r}{0.05}& \multicolumn{1}{r}{0.05}& \multicolumn{1}{r}{0.22}& \multicolumn{1}{r}{0.21}& \multicolumn{1}{r}{0.22}\\
                    \multicolumn{1}{r}{}& \multicolumn{1}{r}{5.00}& \multicolumn{1}{r}{0.20}& \multicolumn{1}{r}{0.20}& \multicolumn{1}{r}{0.19}& \multicolumn{1}{r}{0.08}& \multicolumn{1}{r}{0.08}& \multicolumn{1}{r}{0.08}& \multicolumn{1}{r}{0.37}& \multicolumn{1}{r}{0.37}& \multicolumn{1}{r}{0.34}\\
                    \multicolumn{1}{r}{}& \multicolumn{1}{r}{7.00}& \multicolumn{1}{r}{0.28}& \multicolumn{1}{r}{0.27}& \multicolumn{1}{r}{0.27}& \multicolumn{1}{r}{0.10}& \multicolumn{1}{r}{0.10}& \multicolumn{1}{r}{0.10}& \multicolumn{1}{r}{0.48}& \multicolumn{1}{r}{0.48}& \multicolumn{1}{r}{0.49}\\
                    \multicolumn{1}{r}{}& \multicolumn{1}{r}{9.50}& \multicolumn{1}{r}{0.33}& \multicolumn{1}{r}{0.32}& \multicolumn{1}{r}{0.31}& \multicolumn{1}{r}{0.12}& \multicolumn{1}{r}{0.12}& \multicolumn{1}{r}{0.12}& \multicolumn{1}{r}{0.63}& \multicolumn{1}{r}{0.63}& \multicolumn{1}{r}{0.65}\\
                    \multicolumn{1}{r}{}& \multicolumn{1}{r}{12.00}& \multicolumn{1}{r}{0.36}& \multicolumn{1}{r}{0.35}& \multicolumn{1}{r}{0.35}& \multicolumn{1}{r}{0.14}& \multicolumn{1}{r}{0.14}& \multicolumn{1}{r}{0.14}& \multicolumn{1}{r}{0.71}& \multicolumn{1}{r}{0.72}& \multicolumn{1}{r}{0.71}\\
                    \multicolumn{1}{r}{}& \multicolumn{1}{r}{14.50}& \multicolumn{1}{r}{0.41}& \multicolumn{1}{r}{0.40}& \multicolumn{1}{r}{0.40}& \multicolumn{1}{r}{0.16}& \multicolumn{1}{r}{0.16}& \multicolumn{1}{r}{0.15}& \multicolumn{1}{r}{0.89}& \multicolumn{1}{r}{0.89}& \multicolumn{1}{r}{0.91}\\
                    \multicolumn{1}{r}{}& \multicolumn{1}{r}{17.00}& \multicolumn{1}{r}{0.48}& \multicolumn{1}{r}{0.44}& \multicolumn{1}{r}{0.44}& \multicolumn{1}{r}{0.16}& \multicolumn{1}{r}{0.17}& \multicolumn{1}{r}{0.16}& \multicolumn{1}{r}{0.94}& \multicolumn{1}{r}{0.93}& \multicolumn{1}{r}{0.94}\\
                    \multicolumn{1}{r}{}& \multicolumn{1}{r}{19.00}& \multicolumn{1}{r}{0.49}& \multicolumn{1}{r}{0.42}& \multicolumn{1}{r}{0.42}& \multicolumn{1}{r}{0.17}& \multicolumn{1}{r}{0.17}& \multicolumn{1}{r}{0.16}& \multicolumn{1}{r}{0.89}& \multicolumn{1}{r}{0.88}& \multicolumn{1}{r}{0.89}\\
                    \multicolumn{1}{r}{}& \multicolumn{1}{r}{21.50}& \multicolumn{1}{r}{0.50}& \multicolumn{1}{r}{0.40}& \multicolumn{1}{r}{0.40}& \multicolumn{1}{r}{0.17}& \multicolumn{1}{r}{0.17}& \multicolumn{1}{r}{0.18}& \multicolumn{1}{r}{0.84}& \multicolumn{1}{r}{0.84}& \multicolumn{1}{r}{0.84}\\
                                \cmidrule{3-11}
                                & \multicolumn{1}{r}{Avg} & \multicolumn{1}{r}{0.35}& \multicolumn{1}{r}{0.32}& \multicolumn{1}{r}{0.32}& \multicolumn{1}{r}{0.13}& \multicolumn{1}{r}{0.13}& \multicolumn{1}{r}{0.13}& \multicolumn{1}{r}{0.66}& \multicolumn{1}{r}{0.66}& \multicolumn{1}{r}{0.67}\\
                                \toprule
\end{tabular}
\end{table}

\begin{figure}[H]

\centering

\includegraphics[width=.27\textwidth]{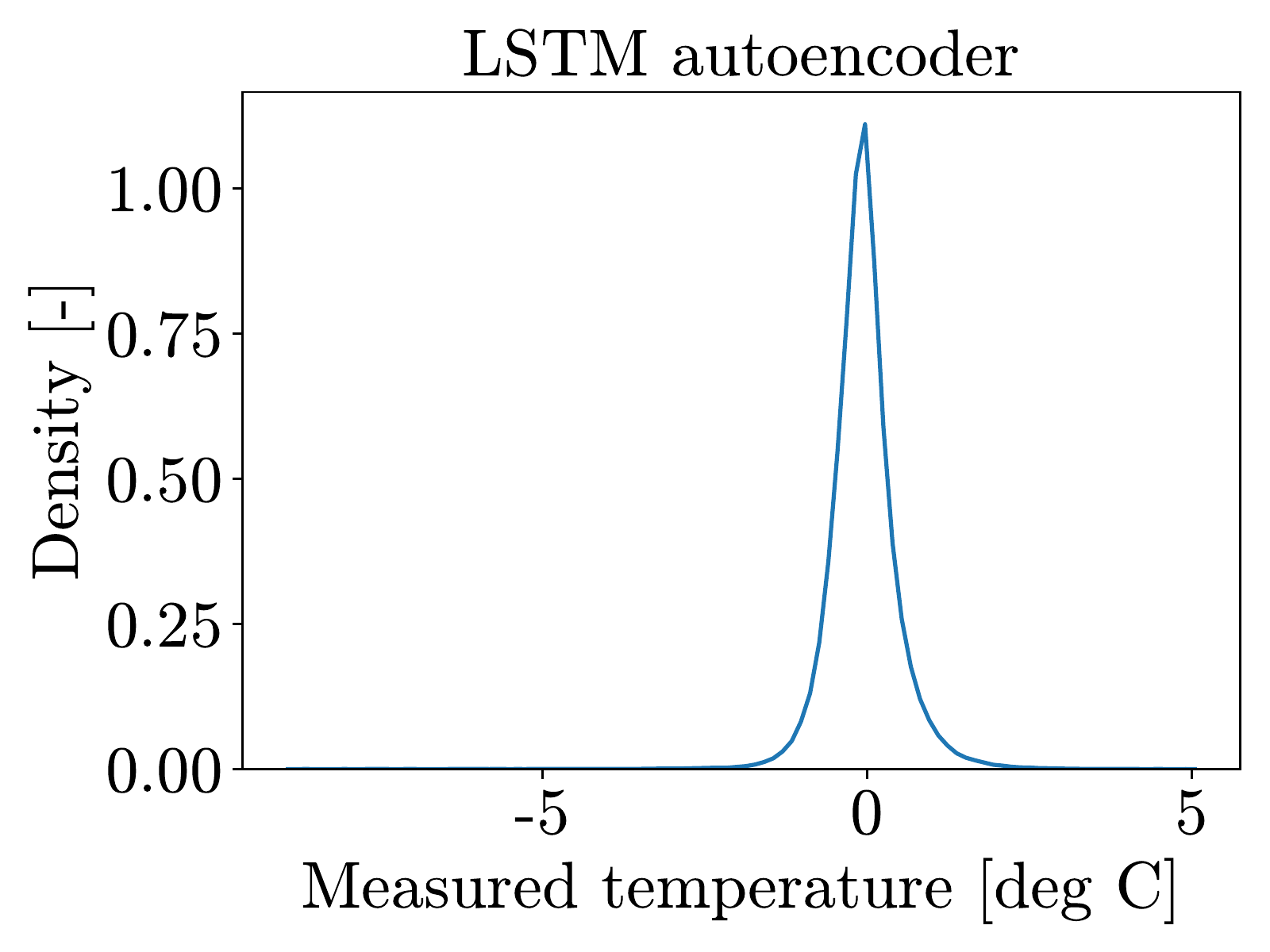}\hfill
\includegraphics[width=.27\textwidth]{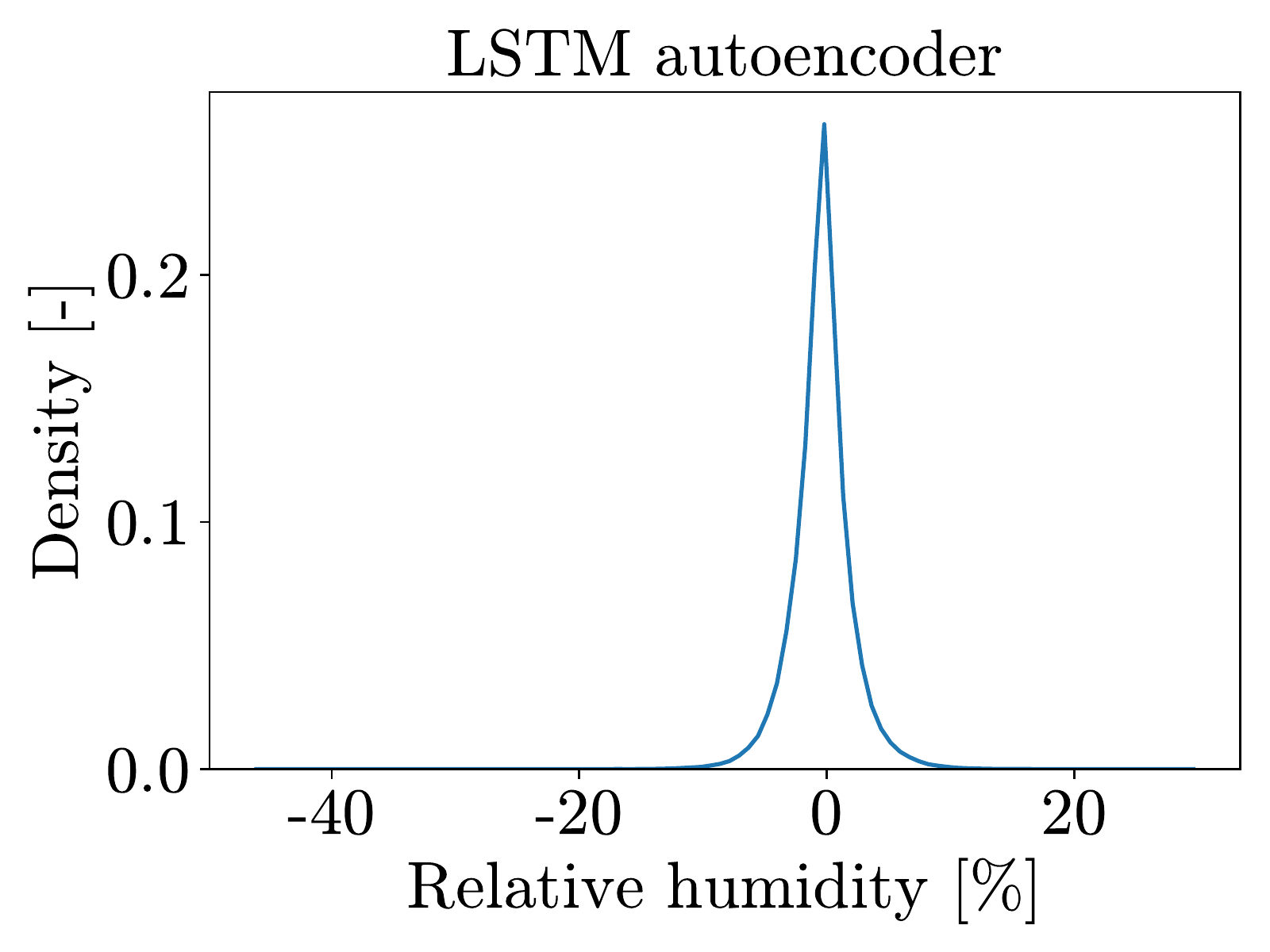}\hfill
\includegraphics[width=.27\textwidth]{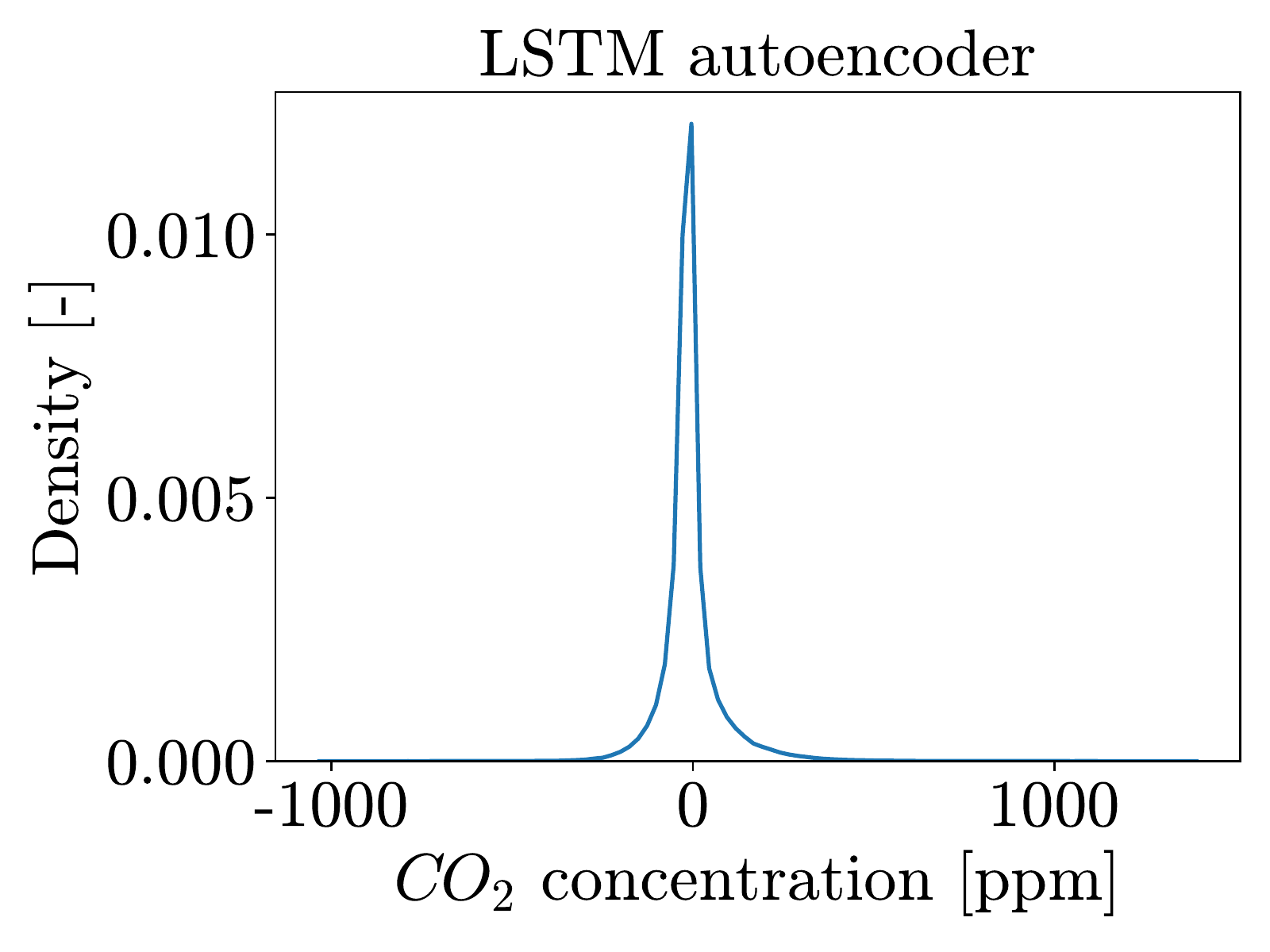}
\caption{Density distribution of the forecasting residuals for a corruption rate of 0.5.} 
\label{fig:err_for}

\end{figure}

Figure \ref{fig:err_for} shows the density distribution of the forecasting residuals produced by the best models identified by the previous table (i.e. LSTM autoencoder), for a predictive horizon of 12 hours. Here, the forecasting residuals were computed as the differences between the observed and predicted values on the evaluation set. For each variable, it was possible to note a normal distribution with a mean of zero, hence confirming that the autoencoders were well developed \cite{jiayuan}. The presence of extreme values in these pictures could be traced back to undetected outliers, as explained in the previous subsection.

\begin{figure}[H]

\centering

\includegraphics[width=.27\textwidth]{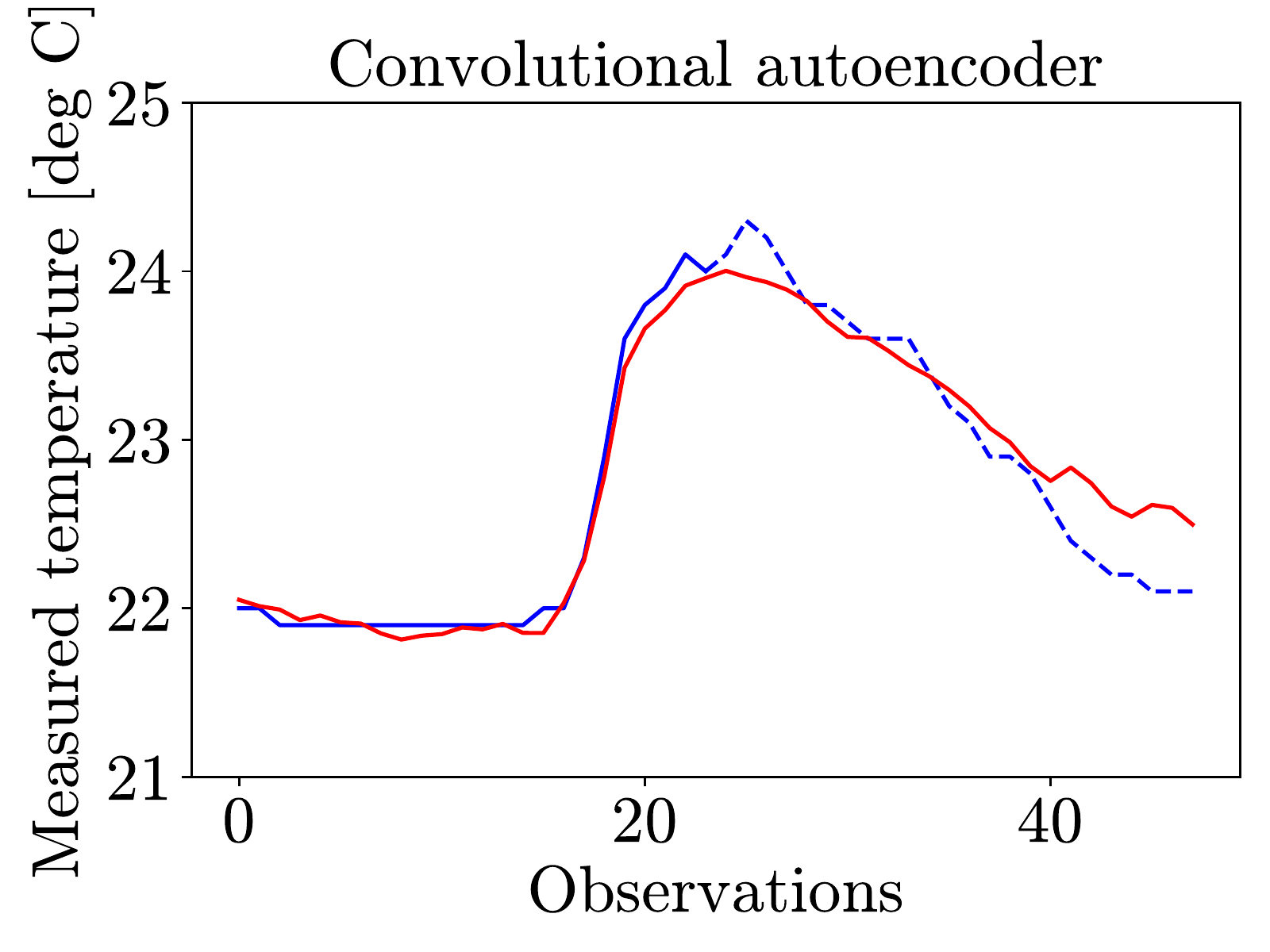}\hfill
\includegraphics[width=.27\textwidth]{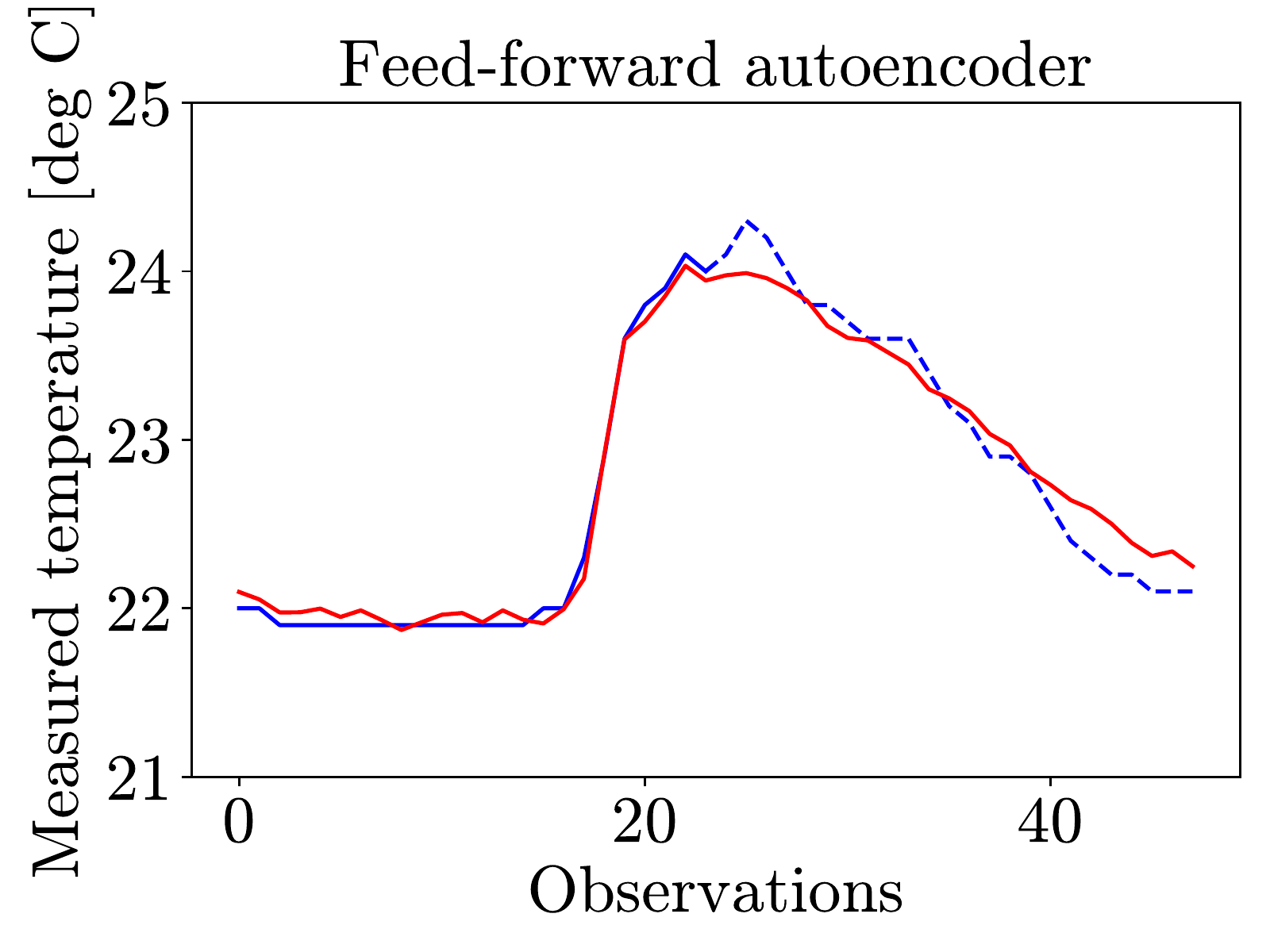}\hfill
\includegraphics[width=.27\textwidth]{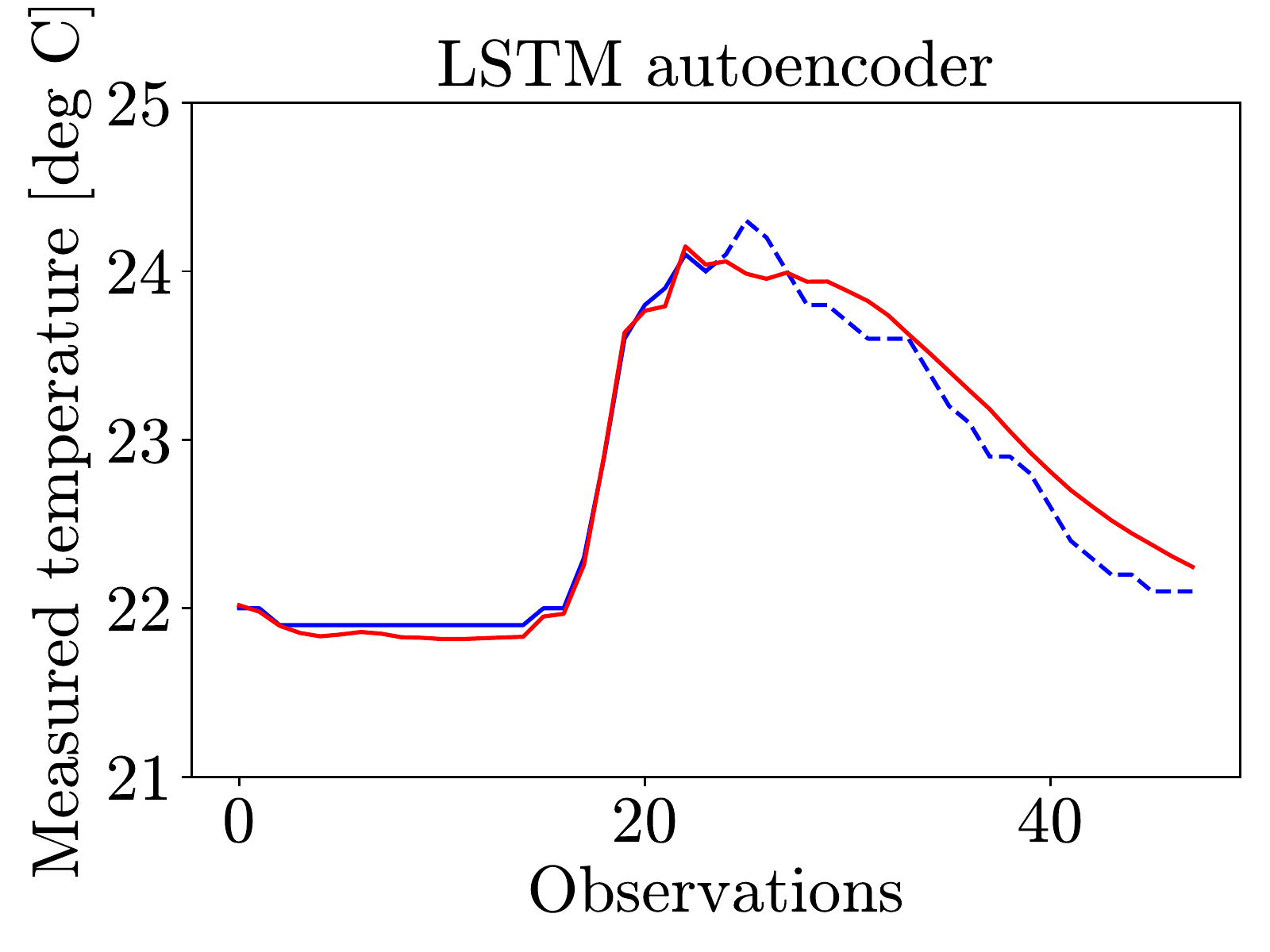}
\includegraphics[width=.27\textwidth]{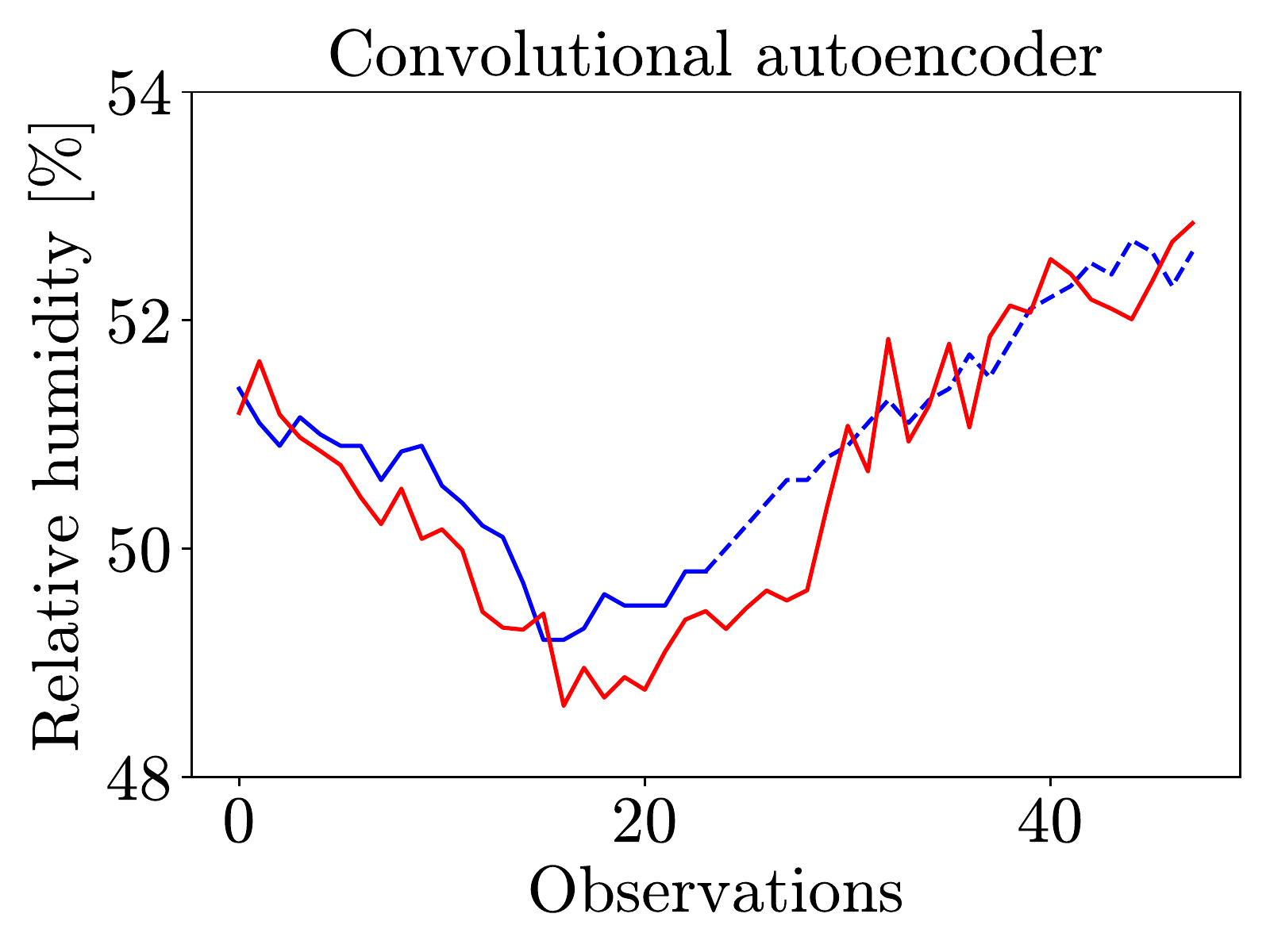}\hfill
\includegraphics[width=.27\textwidth]{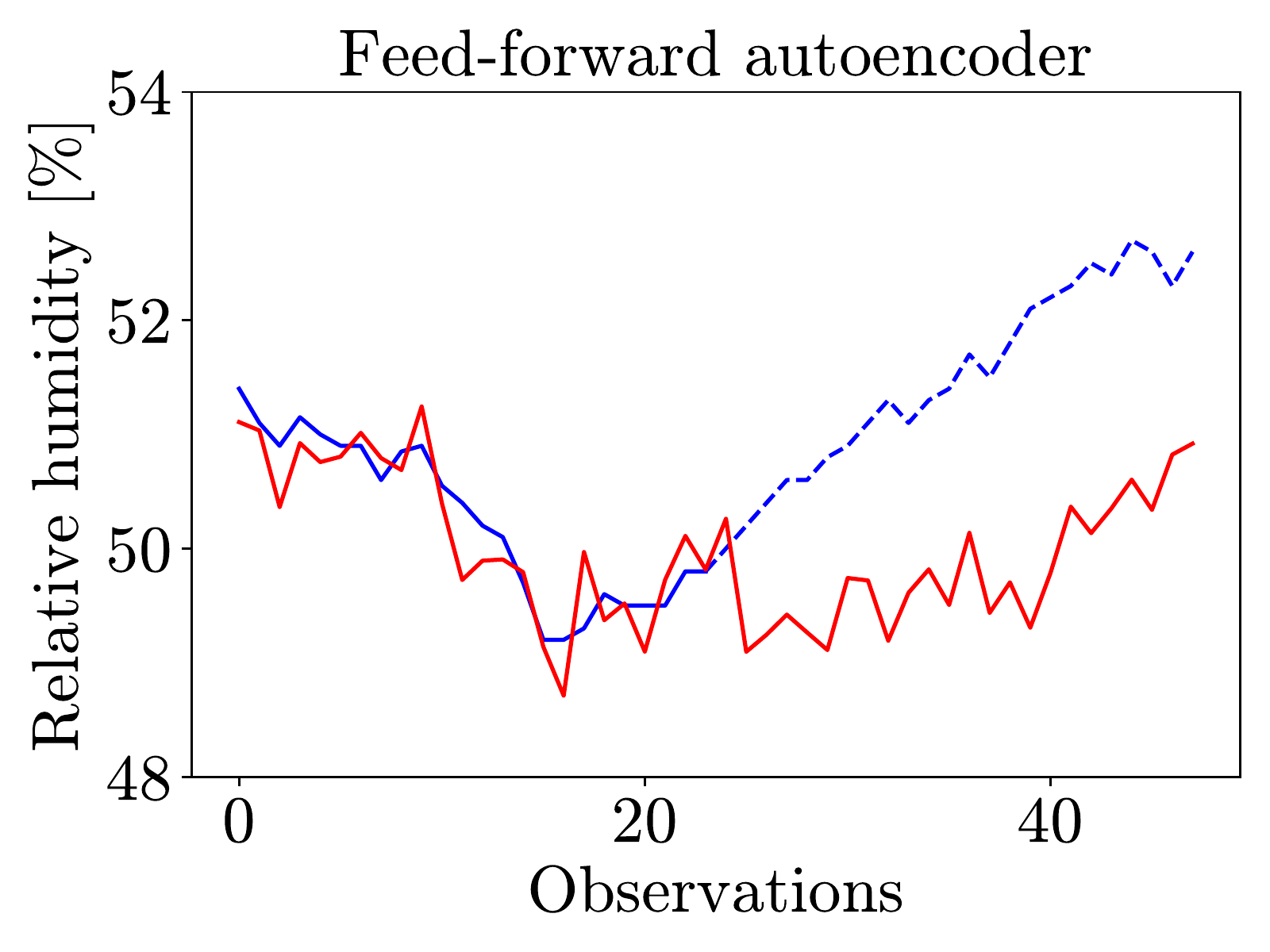}\hfill
\includegraphics[width=.27\textwidth]{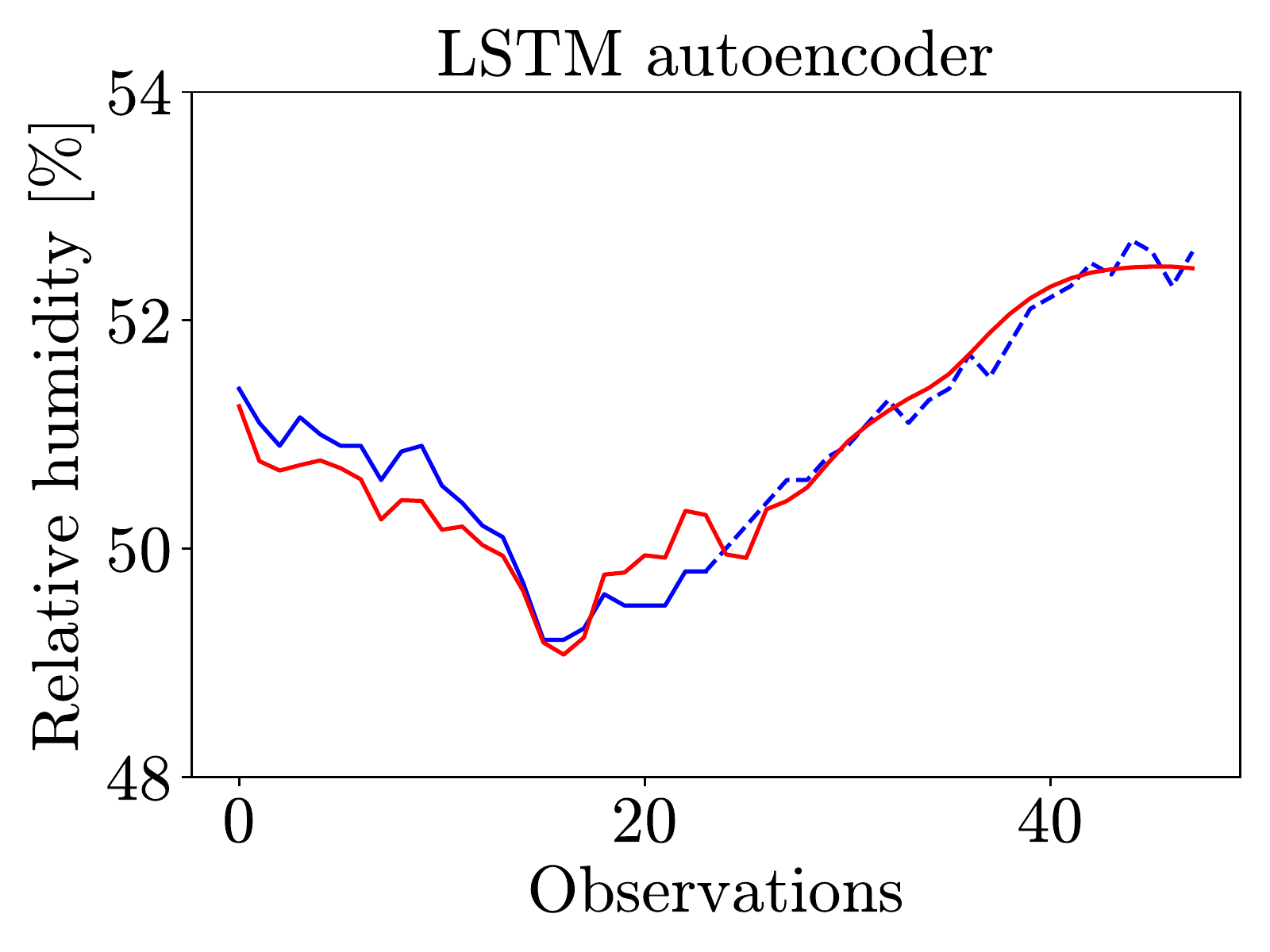}
\includegraphics[width=.27\textwidth]{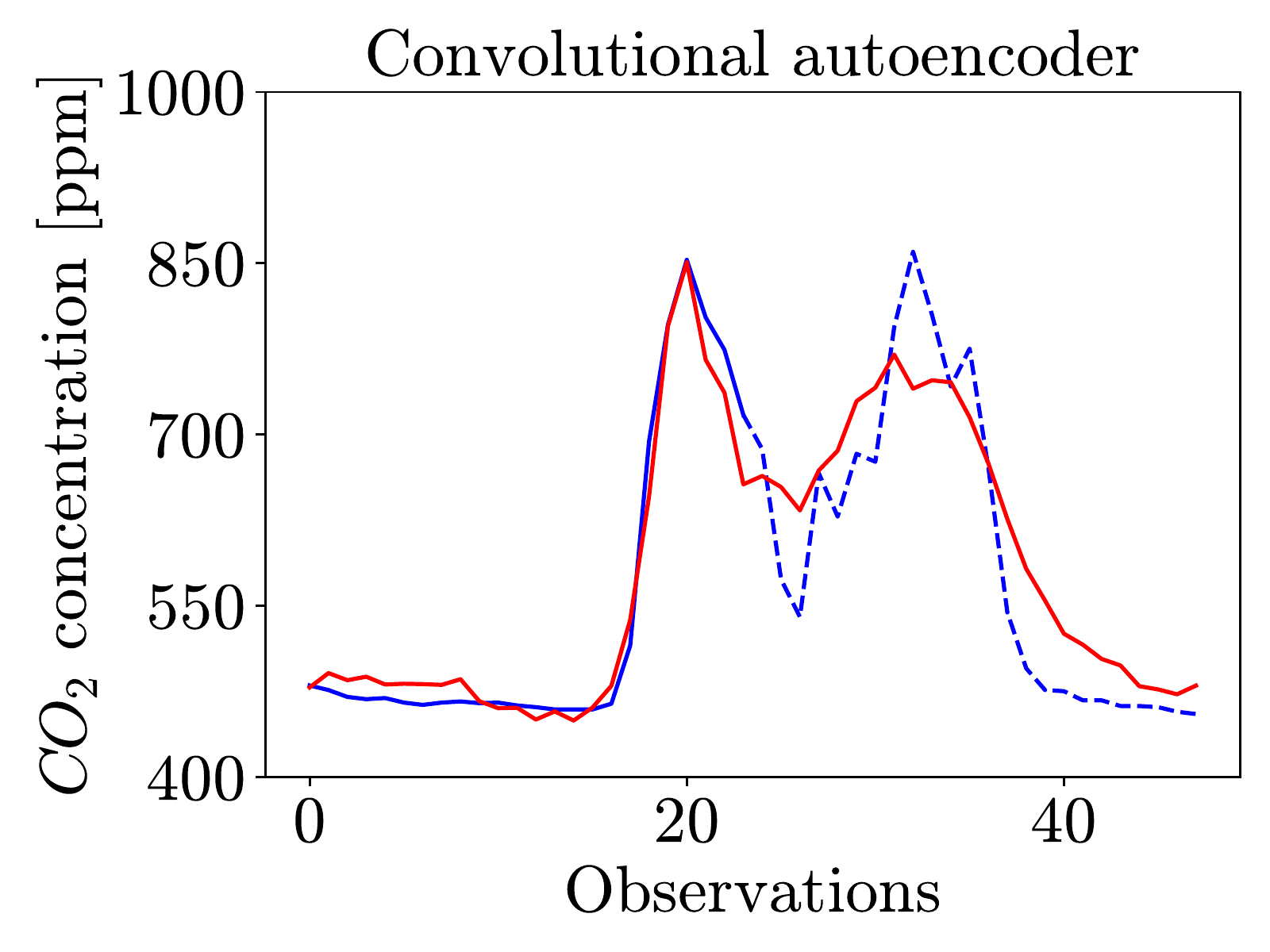}\hfill
\includegraphics[width=.27\textwidth]{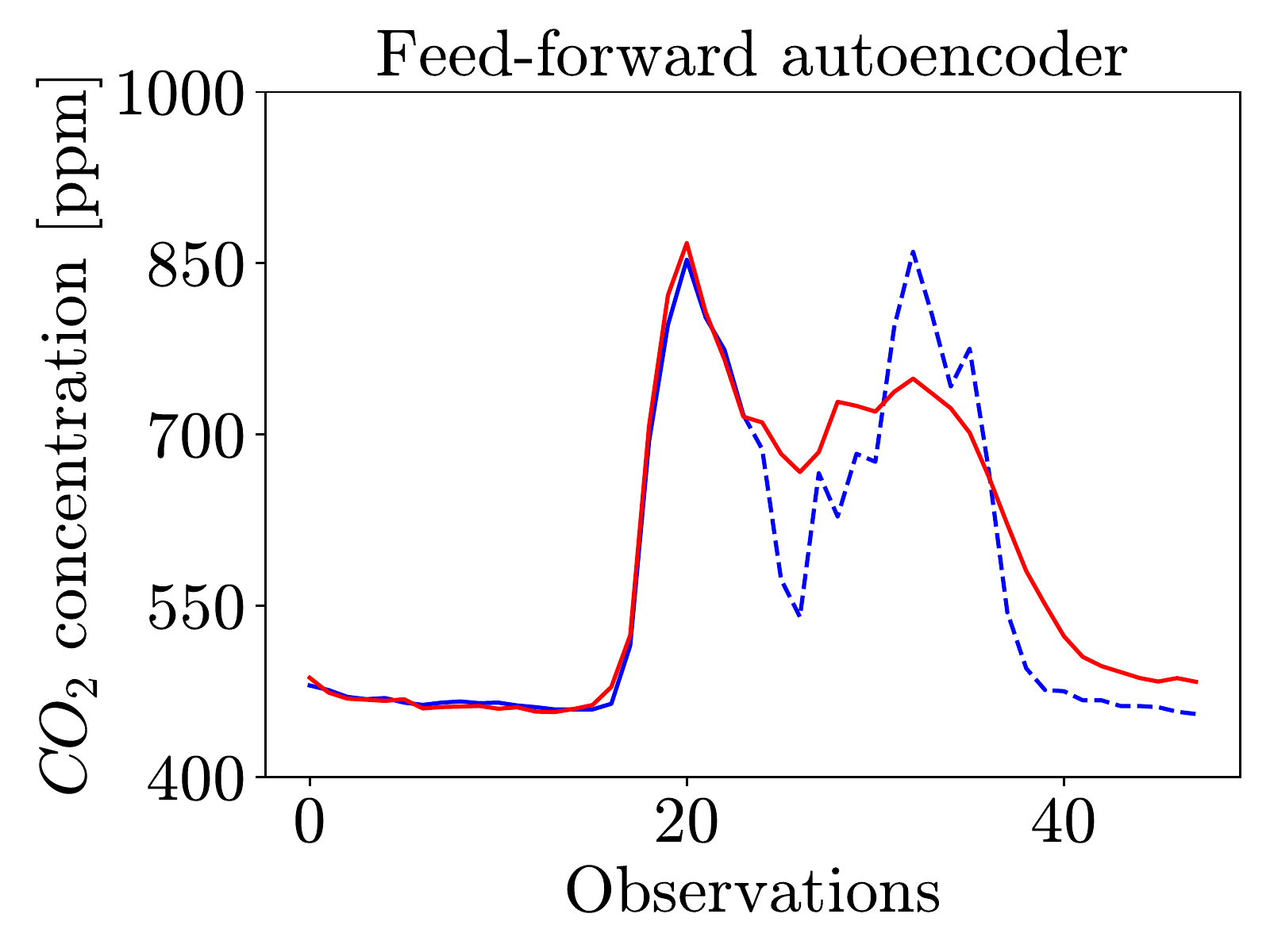}\hfill
\includegraphics[width=.27\textwidth]{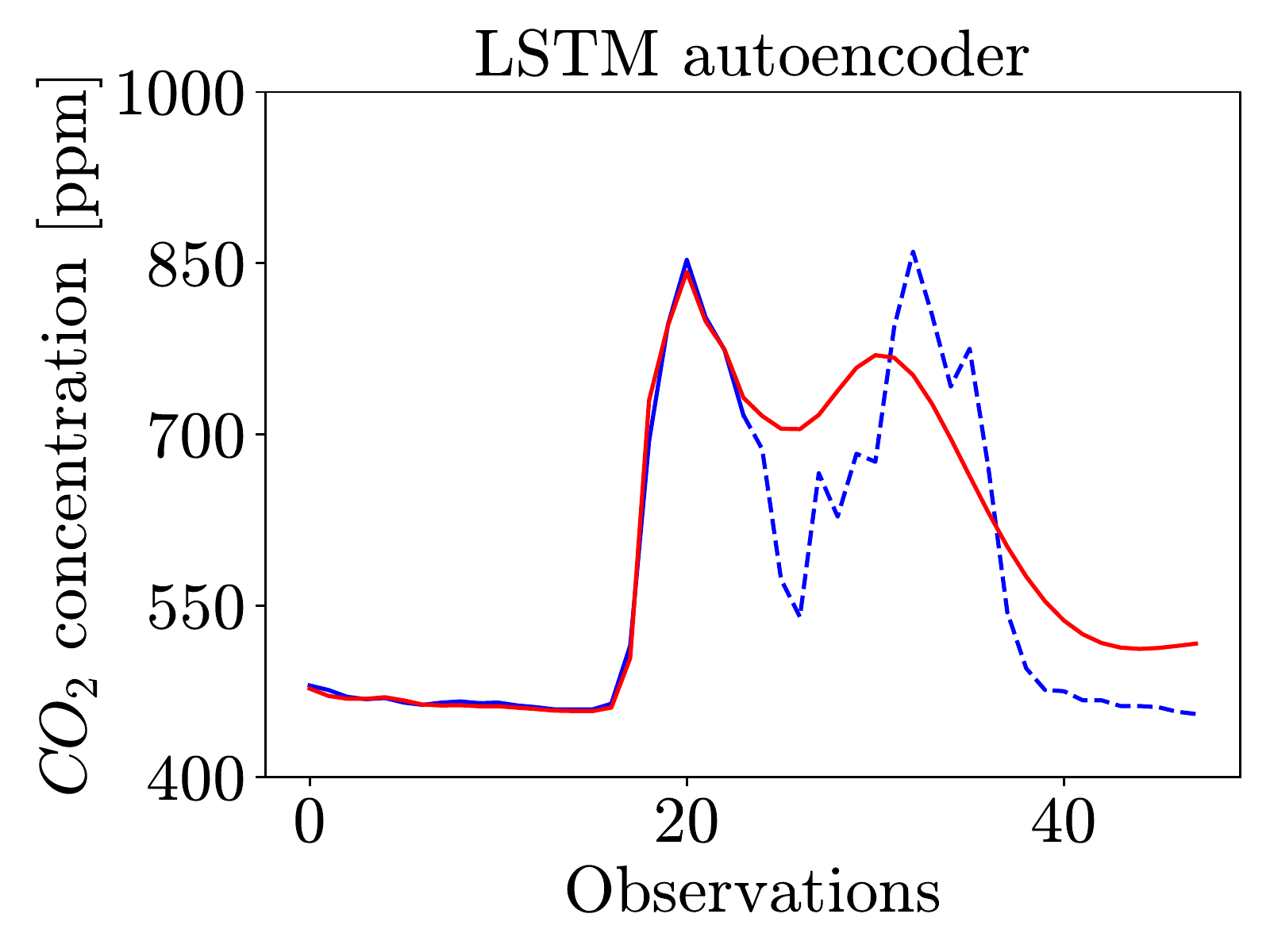}

\caption{One day-long indoor environment data forecasting. Blue colored line represents the real data. Hashed blue colored line represents the missing data. Red colored line represents the reconstruction of the whole day with the adopted model. Observations were sampled to 30 minutes steps.}
\label{fig:data_forecasting}

\end{figure}

Figure \ref{fig:data_forecasting} shows exemplary indoor environment data forecasting over one random day from the evaluation sets. All presented data were corrupted with a masking noise of 0.5 at the end of each time step (predictive horizon of 12 h). The presented data confirmed once again the results presented in Table \ref{tab:forecasting}, namely, that the proposed denoising autoencoder architectures were also suitable for the short-term indoor environment data forecasting.

\section{Discussion and future work}
\label{sec:discussion}

This study presented a method to reconstruct sub-daily indoor environment data time-series, since short-term missing data are often present in building data sets and they could hinder further energy analysis. Considering that, building energy models usually require inputs at hourly resolution \cite{chong} and that, provided a sufficient large training set of data, there are no significant performance drops by modeling occupant behavior data with between minute-wise and hourly time resolutions \cite{RoM2}, it was opted for a 30 minutes frequency to conduct further models' development. An important contribution of this paper is the analysis of autoencoder neural networks for the reconstruction of different types of single indoor environment streams, measured over multiple years. Accordingly, this fills an important research gap in the related literature, since existing studies on missing data inserting either focused on reconstructing a single type of data stream, they required a number of specific information from additional sensors (e.g. weather data) or they were limited by the size of the available data set. \par
As presented in Section \ref{sec:methodology}, data were split in three sets before normalization. Model training was performed using a training set, the optimal model configuration was chosen based on performance on the validation set and the final data reconstruction accuracy was tested using the evaluation set. Approximately 94,085 full days of observations were available, which makes this study -- as far as the authors know -- the largest of its kind for indoor environment data time-series reconstruction purposes. In order to guarantee a significant generalization of the models, it was decided to take an extensive evaluation set. Accordingly, approximately 2.7 million data points were used for models' evaluation. Improved performance of the final models could be achieved by introducing the dimension of each data set as an additional hyperparameter. However, this choice would have led to additional computational costs and, therefore, it was not pursued. \par
In total, 1,890 hyperparameter combinations were explored by applying a grid search. For experiments run in CPU mode on a single laptop with Intel(R) Core(TM) i7-8850H CPU (2.60GHz) processor, the time required for training each model was approximately 0.98, 0.12 and 5.62 s/epoch, respectively for the convolutional, feed-forward and LSTM autoencoders. Additionally, the time required for reconstructing the first eight days of the evaluation set (with the pre-trained models) was approximately 0.066, 0.036 and 0.264 s, respectively. In particular, approximately 7,000 core hours of GPU and CPU computations were exploited using compute sources granted at RWTH Aachen University. However, since the size of the training set was set to 30~\% of the total data set, the time required for simulation would have decreased in case less data had been available. Due to the limited computational resources, the comprehensive model development  that requires extensive computations could not be performed within the building control. Rather, the building control should include the model evaluation which consists of computationally efficient single forward pass, that required around 0.065, 0.036 and 0.256 s for a single day reconstruction, respectively for the convolutional, feed-forward and LSTM autoencoders. The model development, which includes model training and validation, should be conducted as  a single-time offline procedure and the final model should be coupled with room control, as already proposed by  Romana Markovic \cite{markovic2020} and Bauer et al. \cite{bauer}.
\par  
This work provided important insights regarding the occurrence of the neural networks saturation for analytics related to building's performance. The consequence of model saturation was that the weights were not updated due to the vanishing gradient problem (Section \ref{sec: saturation}), which led always to identical predicted values. This problem was observed in case of the convolutional and feed-forward autoencoders, while it was not detected in case of the LSTM autoencoders. The suitable approach to tackle the model saturation was explored in the existing literature on computer vision and general machine learning. Even though all the computed saturation metrics were well above the defined saturation limit (0.1), the SAT decreased with the corruption rate for both indoor air temperature and $CO_2$ concentration and remained almost unchanged for relative humidity. This was inconsistent with the increasing RMSE trend of the latter variable (Table \ref{tab:subdaily}). It can be concluded that the above saturation metric could not be used as an additional performance measure, since it was dependent also on the sequence-variability of the original data. In particular, the worst saturation performance on $CO_2$ data could be explained with the presence of more extreme values and frequent peaks. This could explain the reason why LSTM neural networks suffered saturation issues in the paper proposed by Markovic et al. \cite{RoM1}. In this earlier study, an LSTM-based model was applied to plug-in loads data and saturation occurred in more than 70~\% of the trained configurations. This could be caused by the larger data imbalance of the plug-in loads and the extreme values. Namely, similarly to plug-in loads, the time-series of $CO_{2}$ concentration consist of frequent peaks and extreme values, which showed to be a particular complexity to be considered when using the LSTM for building's energy analysis.
\par
The data reconstruction analysis, applied to the non corrupted data, revealed that the proposed autoencoder neural networks, especially the LSTM configuration, could accurately capture the indoor environment patterns. Accordingly, this represents a significant practical potential for the inclusion of these methods in the real time building control. This could be used for anomaly detection purposes, by identifying data sequences with atypical behavior (e.g. noisy data, sensors' malfunctioning) (Figure \ref{fig:patterns}). On the other hand, the performance of the convolutional configuration spiked out when a certain masking noise was applied to the evaluation sequences. It can be, therefore, stated that the spatial correlations of input data were more important than the temporal ones, when a gap-filling method was investigated. \par
The NRMSE analysis established that relative humidity data patterns were, in general, easier to detect by the proposed models. The lowest performance was observed in case of the $CO_2$ data, due to the presence of more extreme values and frequent peaks compared to the other variables. However, even in the last case, the proposed ANNs could reconstruct the missing sequences with a significant less average error than then adopted baseline approach (approximately 30~\% less in terms of RMSE). Additionally, even the $CO_2$ sequences could be reconstructed with RMSE below 80 $ppm$, which indicases that the reconstruction performance could even be in similar accuracy range as the measuring tolerance of some commercial $CO_2$ sensor devices. 
\par
Similarly to other papers that analyzed the problem of missing data inserting for building monitoring data \cite{jiayuan,luis,peppanen}, ignoring outliers, the reconstruction residuals along the inserted values lay within a normal distribution with a mean of zero. The same model architecture could reproduce a similar distribution for each of the analyzed indoor environment stream, which represents one of the novelty of this work. Additionally, the overall descriptive statistics of the evaluation set were not affected by simulating the missing values with the proposed method. However, the standard deviation of $CO_2$ concentration data was slightly underestimated for high corruption rates (CR $>$ 0.5).
\par
The focus on reconstructing short missing sequences was motivated by several reasons.  Firstly, more than 60~\% of the missing sequences from building control ranged between one and six hours \cite{clar,balt}. Motivated by this common yet rarely addressed problem in room automation, we aimed to explore computational approaches that could tackle this issue. Secondly, one of the starting hypotheses was, that the sub-daily time-series could be the longest sequences from room control that pose stationary time-series properties. As a consequence, the meaningful research boundaries were set and the computational methods to tackle this particular problem were investigated. Nonetheless, the future research activities should include as well  the analysis of the potential and goodness of multi-day time-series reconstruction and forecasting.
\par
A further analysis on the models revealed that the proposed method could forecast the indoor air temperature data even better than calibrated Modelica-based building performance simulation tools applied in other studies \cite{RoM2}. The temporal correlations of input data gained significant importance with respect to the reconstruction case, placing the LSTM configuration on a slightly better performance level than the convolutional one. In this regard, a denoising autoencoder which relies, at the same time, on LSTM and convolutional units could further increase the predictive accuracy of the model. 
\par
One of the possible limitations of this study is a direct consequence of the training process. The proposed autoencoders were, indeed, implemented to capture information related to the daily trends of the observed variables. Accordingly, day-ahead data sequences cannot be reconstructed with the current training scheme. Future work should, therefore,  evaluate the implemented autoencoders for reconstructing and forecasting energy and environmental data time-series over longer time horizons and analyze their performance on different periods. An other limitation of this work is due to the used data set. Despite the extension of the evaluation set, it is not yet clear how the models would behave when applied to monitoring data from other domains. However, it may perhaps be observed, that the quality of the same data set was not affected by climatic and environmental factors. In case of data collected in Germany in and between the years 2014 and 2017, no extreme weather conditions, floods or hurricanes were indeed registered. Having that in mind, the model applicability for the unusual or extreme weather conditions, such as electricity outage, extreme heat or cold was not investigated\footnote{A relevant work of the identification of such events based on extreme weather and environmental conditions was presented by Meier et al. \cite{meier2019}}. In order to increase the generalization capability of the developed methods, the inclusion of monitoring data collected in multiple buildings with significant differences in thermal mass and design should be further researched. In this regard, as a future extension of this work, the proposed pre-trained algorithms might be used to improve generalization to other domains, in combination with a suitable domain adaptation procedure. Finally, in order to facilitate the real-life applications of the models, future studies should focus on the optimization of a single autoencoder for multiple time-series.

\section{Conclusion}
\label{sec:conclusions}

The aim of this paper was to develop an approach for reconstructing short-term indoor environment data time-series. For that purpose, three autoencoder neural networks models were implemented and polynomial interpolation methods were evaluated for baseline comparison. The evaluation of models' performance was conducted using indoor air temperature, relative humidity and $CO_2$ concentration data. The key findings could be summarized as follows:
\begin{itemize}
    \item The proposed method outperformed polynomial interpolation models for filling sub-daily environmental data gaps. 
    \item For the indoor air temperature, relative humidity and $CO_2$ concentration data, the differences between the observed and inserted missing values (i.e. reconstruction residuals) had a normal distribution with a mean of zero.
    \item The standard deviation of the filled data set might be slightly underestimated if the proposed method is used to reconstruct sub-daily $CO_2$ concentration data gaps.  
    \item In average, relative humidity data were easier to reconstruct, while $CO_2$ concentration data were more challenging for the models.
    \item The developed method could be also used for predicting the indoor environment data with high accuracy, over the multi hour time horizon. 
    \item The implementation of normalized initial weights, ReLU activation function and batch-normalization could avoid saturation issues for ANNs. 
\end{itemize}
In summary, the proposed method has a practical potential for the inclusion in the real-time building control, as back-up option in case of sensor failure. However, further evaluations should be carried out, such as the applicability to other domains or to multiple time-series. Furthermore, the same models might be used to complete indoor environment data sets with small gaps (i.e. $<$ 24 h), improving the performance of later applied data-driven models. In this case, further evaluation about the applicability on longer periods should be carried out.

\section{Acknowledgements}
Part of this work was funded by the Deutsche Forschungsgemeinschaft (DFG, German Research Foundation) -- TR 892/4-1 -- and by the German Federal Ministry of Economics and Energy (BMWi) as per resolution of the German Parliament under the funding code 03EN1002A. Simulations were performed with computing resources granted by RWTH Aachen University under project rwth0622. We thank the EBC Institute, E.ON ERC at RWTH Aachen University for providing the monitoring data. This paper benefited greatly from discussions with members of IEA EBC Annex 79.

\biboptions{sort&compress}

\renewcommand{\appendixpagename}{\appendixname}

\renewcommand{\appendixtocname}{\appendixname}

\Needspace{80\baselineskip}
\begin{appendices}
\renewcommand{\thesection}{A.\arabic{section}}
\section{Data cleaning}
\label{sec:cl}
This Appendix provides additional information about the adopted data cleaning process.
\subsection{Outliers}

Outliers were detected favoring model generalization, rather than accuracy. The aim of this paper was, indeed, to provide a tool to reconstruct indoor environment time-series, independently of the type and quality of data. Ma et al. \cite{ma} applied the IQR method for the reconstruction of building electric power data, defining as outlier every value out of the following range:

\begin{equation}
[Q1-1.5\cdot IQR;~ Q3+1.5\cdot IQR]~, 
\label{eq:iqr}
\end{equation}

where $Q_{1}$ is the first quartile of the dataset, $Q_{3}$ is the third quartile, $IQR$ is the difference between the third and first quartile. Data out of the previous interval were replaced with the nearest IQR limit \cite{ma}. However, since the generalization characteristics of ANNs depend on the noise included in the training data \cite{soft}, the authors decided not to follow this approach at the expenses of an overall reduced accuracy \cite{soft}. Accordingly, outliers were detected based on theoretical limits fixed by Markovic et al. \cite{RoM} in a different study, where a subset of the same data set was analyzed. Therefore, temperature was established between -10 and +40 \textdegree C, based on the plausible range for the continental climate in Germany \cite{RoM}. Relative humidity was set between 0 and 100~\% \cite{RoM}, while $CO_2$ concentration was assumed to be between 0 and 2,500 $ppm$ \cite{RoM}. Table \ref{tab:statistics} summarizes descriptive statistics for the data set before and after outliers detection, based on the methods proposed in the literature \cite{RoM,ma}. The IQR method proposed by Ma et al. \cite{ma} seemed to oversimplify the problem, by identifying as outliers a wide range of values (Table \ref{tab:statistics}). It was, therefore, opted for the other approach \cite{RoM}.

\begin{table}[H]
\centering
\caption{Descriptive statistics for the data set before (raw data) and after (\cite{RoM,ma}) outliers detection. Std and L/U IQR stand respectively for standard deviation and lower/upper IQR limit.}
\label{tab:statistics}
\begin{tabular}{lccccccccc}
 \toprule
                   & \multicolumn{3}{l}{T [\textdegree C]} & \multicolumn{3}{l}{RH [\%]} & \multicolumn{3}{l}{$CO_2$ [ppm]} \\
                    \toprule
                   & Raw data & \cite{ma} & \cite{RoM} & Raw data  & \cite{ma} & \cite{RoM} & Raw data  & \cite{ma} & \cite{RoM}  \\
                    \toprule
Min            & 6        & 19.4     & 6        & 0.5       & 1.45     & 0.5      & 0         & 265       & 192       \\
Max            & 2.34E+16 & 25.8     & 37.1     & 2.53E+03  & 72.25    & 99.4     & 1.31E+04  & 737       & 2,000      \\
Mean               & 6.47E+10 & 22.64    & 22.63    & 3.75E+01  & 37.52    & 37.5     & 5.16E+02  & 509.12    & 516.3     \\
Median             & 22.7     & 22.7     & 22.7     & 36.4      & 36.4     & 36.4     & 491       & 491       & 491       \\
Std & 3.60E+13 & 1.22     & 1.4      & 1.18E+01  & 11.01    & 11       & 1.25E+02  & 100.27    & 124.25    \\
L IQR     & 19.4     & 19.4     & 19.4     & 1.45      & 1.45     & 1.45     & 265       & 265       & 265       \\
U IQR    & 25.8     & 25.8     & 25.8     & 72.25     & 72.25    & 72.3     & 737       & 737       & 737       \\
 \toprule
Outliers      & /        & 2.54E+06 & 358      & /         & 1.71E+03 & 313      & /         & 4.18E+06  & 307  \\
 \toprule
\end{tabular}
\end{table}

\subsection{Missing values}

\begin{figure}[H]
    \centering
    \includegraphics[height=0.24\textheight]{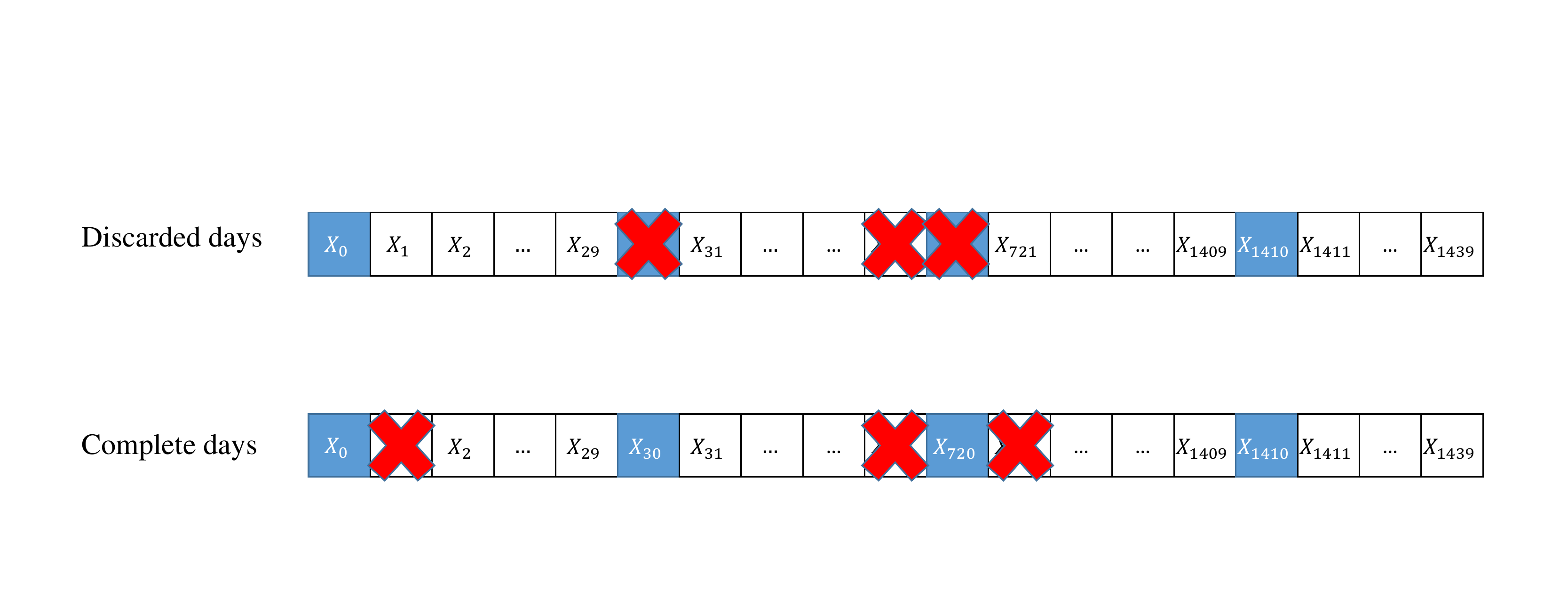}
    \caption{A visual representation of the discarded and complete days. White blocks are minute-wise observations. Blue blocks are observations with 30-minutes frequency resolution.}
    \label{fig:propagation}
\end{figure}

Based on the logging frequency and monitoring duration, it could be expected that around 181 million sets of observations were collected for each variable from 2014 to 2017. Of the latter, only 73 million data points were correctly recorded for $T$ (59.6~\% error rate), while 70 million for both $RH$ and $CO_2$ (61.3~\% error rate). In order to increase the computational efficiency of the models, frequency was reduced from minute-wise to 30 minutes, leading to approximately 2.3 million data points for each variable.
For the missing values handling, a complete case analysis approach was adopted, where only full day of observations with the current resolution were considered. Hence, the number of available monitoring points per variable were reduced to 1.5 million data. Accordingly, from the starting 376,938 daily observations, models were applied only to 94,085 days (75~\% error rate). An overview of the missing values handling strategy is presented in Table \ref{tab:days} and in Figure \ref{fig:propagation}.

\begin{table}[H]
\centering
\caption{Overview of the preprocessed data set.}
\label{tab:days}
\begin{tabular}{|*{4}{p{3cm}|}}
\toprule
 \multicolumn{1}{c}{} & \multicolumn{1}{c}{$T$}& \multicolumn{1}{c}{$RH$}& \multicolumn{1}{c}{$CO_2$} \\
 \toprule
  \multicolumn{1}{l}{Frequency [min]} & 
 \multicolumn{1}{c}{30}& \multicolumn{1}{c}{30}&  \multicolumn{1}{c}{30}\\
 \multicolumn{1}{l}{Expected days} & 
 \multicolumn{1}{c}{125,646}& \multicolumn{1}{c}{125,646}&  \multicolumn{1}{c}{125,646}\\
  \toprule
  \multicolumn{1}{l}{Discarded days} & 
 \multicolumn{1}{c}{94,265}& \multicolumn{1}{c}{94,294}&  \multicolumn{1}{c}{94,294}\\
 \multicolumn{1}{l}{Complete days} & \multicolumn{1}{c}{31,381}&
 \multicolumn{1}{c}{31,352}&  \multicolumn{1}{c}{31,352}\\
 \toprule
\end{tabular}
\end{table}

\section{Overfitting}
\label{sec:ov}

\begin{figure}[H]

\centering

\includegraphics[width=.45\textwidth]{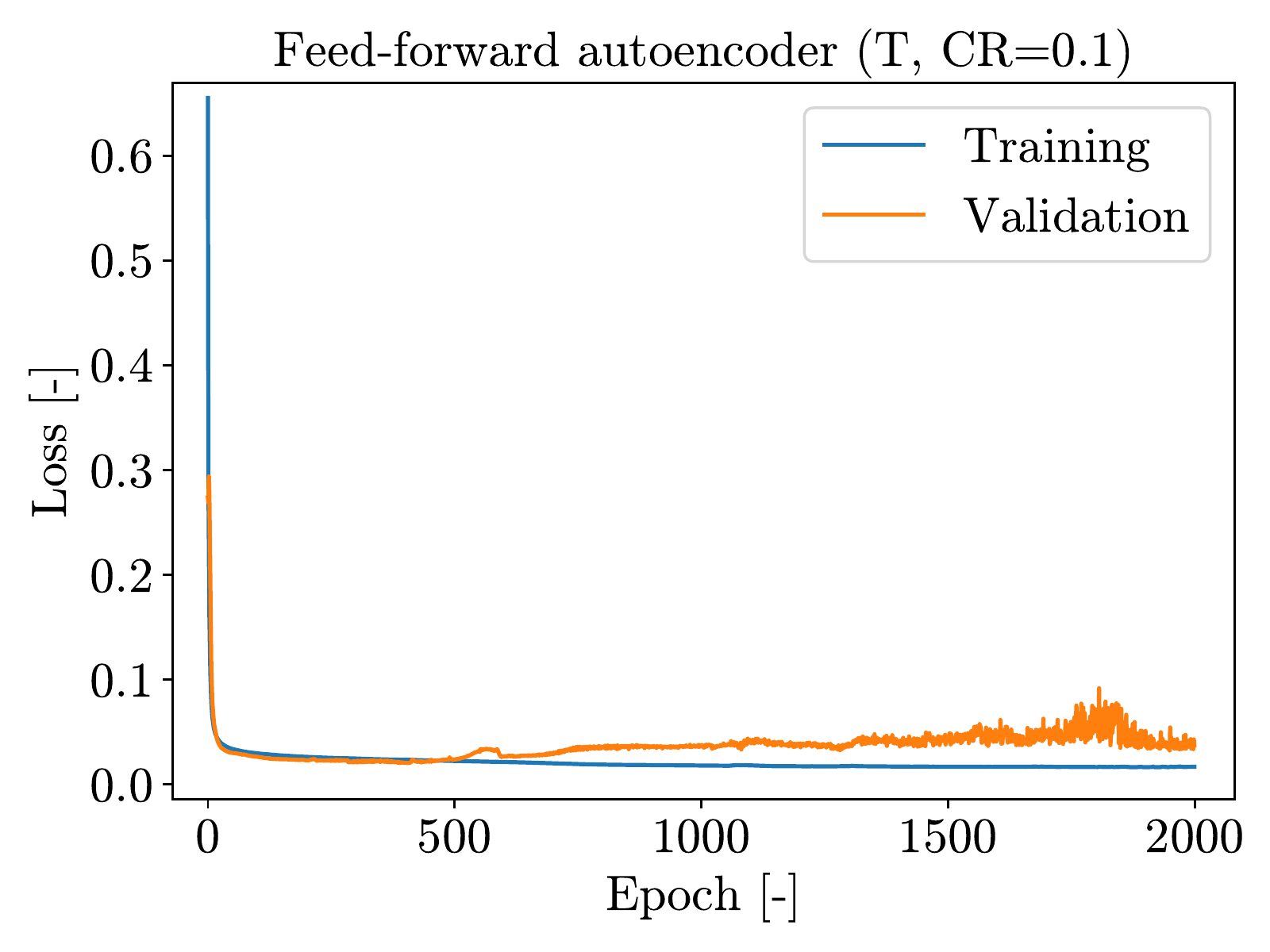}
\caption{Example of a training vs validation curve with overfitting.} 
\label{fig:early}

\end{figure}

Figure \ref{fig:early} shows the training vs validation curve of the feed-forward model applied to the indoor air temperature data (CR = 0.1), in case of overfitting. 

\end{appendices}

\end{document}